%% file: main.tex
\pdfoutput=1
\RequirePackage[T1]{fontenc}
\documentclass[12pt]{article}
\usepackage[height=8.85in,width=6.45in]{geometry}

\usepackage[utf8]{inputenc}
\usepackage{amsmath}
\usepackage{amssymb}
\usepackage{mathtools}
\numberwithin{equation}{section}
\usepackage{slashed}
\usepackage{braket}
\usepackage{enumitem}
\usepackage[svgnames]{xcolor}
\usepackage[colorlinks,citecolor=DarkGreen,linkcolor=FireBrick,urlcolor=FireBrick,linktocpage=true,unicode]{hyperref}

\usepackage{tocloft}

\usepackage[bottom]{footmisc}
\input{macros.tex}

\newcommand*{\email}[1]{\footnote{\href{mailto:#1}{\texttt{#1}}}}

\titlespacing{\section}{0pt}{*1}{*-1}
\titlespacing{\subsection}{0pt}{*1}{*-1}
\titlespacing{\subsubsection}{0pt}{*1}{*-1}
\titlespacing{\paragraph}{0pt}{*1}{1em} %

\setlist[itemize]{
topsep=0em, 
partopsep=0em, 
itemsep=0.08em}

\setlist[enumerate]{
topsep=0em, 
partopsep=0em, 
itemsep=0.08em}

\begin{document}

\begin{titlepage}

\begin{flushright}
Last Update: \today
\end{flushright}

\vskip 2.5em
\begin{center}

{
\LARGE \bfseries %
\begin{spacing}{1.15} %
\input{title} %
\end{spacing}
}

\vskip 1em
Weimin Wu$^{\dagger*}$\email{wwm@u.northwestern.edu}\quad
Maojiang Su$^{\dagger*}$\email{sumaojiang@mail.ustc.edu.cn}\quad
Jerry Yao-Chieh Hu$^{\dagger*}$\email{jhu@u.northwestern.edu}\quad
Zhao Song$^{\ddag}$\footnote{\href{mailto:magic.linuxkde@gmail.com}{\texttt{magic.linuxkde@gmail.com}}}\quad
Han Liu$^{\dagger\S}$\email{hanliu@northwestern.edu}

\def\thefootnote{*}
\footnotetext{Equal contribution. 
Code is available
at \href{https://github.com/MAGICS-LAB/icdl}{GitHub}. 
Latest version is on \href{https://arxiv.org/abs/2411.16549}{arXiv}.}

\vskip 1em

{\small
\begin{tabular}{ll}
 $^\dagger\;$Center for Foundation Models and Generative AI, Northwestern University, Evanston, IL 60208, USA\\
 \hphantom{$^\ddag\;$}Department of Computer Science, Northwestern University, Evanston, IL 60208, USA\\
 $^\ddag\;$Simons Institute for the Theory of Computing, UC Berkeley, Berkeley, CA 94720, USA\\
 $^\S\;$Department of Statistics and Data Science, Northwestern University, Evanston, IL 60208, USA
\end{tabular}}

\end{center}

\noindent
\input{0abstract}

\end{titlepage}

{
\setlength{\parskip}{0em}
\setcounter{tocdepth}{2}
\tableofcontents
}
\setcounter{footnote}{0}

\section{Introduction}
\label{sec:intro}

\input{1intro}

\section{Preliminaries: In-Context Learning and In-Context Gradient Descent}
\label{sec:preliminaries}
\input{2preliminary}

\section{In-Context Gradient Descent on \texorpdfstring{$N$}{}-Layer Neural Networks}
\label{sec:method}

\input{3N_NN_approx}

\section*{Impact Statement}
\input{impact}

\section*{Acknowledgments}
\input{x_acknowledgments}

\clearpage
\newpage
\normalsize

\appendix
\part*{Supplementary Material}
\label{sec:append}

{
\setlength{\parskip}{-0em}
\startcontents[sections]
\printcontents[sections]{ }{1}{}
}

\input{appendix}

\clearpage
\def\arxivfont{\rm}
\bibliographystyle{plainnat}

\bibliography{refs}

\end{document}

%% file: macros.tex
\allowdisplaybreaks
\usepackage{graphicx}
\usepackage{tikz}
\usetikzlibrary{shapes.geometric, arrows, positioning}
\usepackage{tikz-cd}
\usepackage{times}
\usepackage{courier}
\usepackage{physics}
\usepackage{xcolor}
\usepackage{natbib}
\usepackage{mdframed}
\usepackage{nicefrac}
\usepackage{booktabs}
\usepackage{lipsum}
\usepackage{titlesec}
\usepackage{wrapfig,lipsum,booktabs}
\usepackage{blindtext}
\usepackage{algorithm}
\usepackage{algorithmicx}
\usepackage{algpseudocode}
\usepackage{titletoc}
\usepackage{todonotes}
\usepackage{titletoc}
\usepackage{setspace}
\usepackage{amsfonts}
\usepackage{subcaption}

\usepackage{dsfont} %
\newcommand{\one}{\mathds{1}}

\usepackage[nameinlink,capitalize,noabbrev]{cleveref}

\usepackage{CJK}

\usepackage{array}
\usepackage{makecell}

\usepackage{slashed}

\def\({\left(}
\def\){\right)}
\def\[{\left[}
\def\]{\right]}

\usepackage{dashbox}
\usepackage{xcolor}
\usepackage{colortbl}
\definecolor{lightyellow}{rgb}{1.0, 0.95, 0.7}
\definecolor{Blue}{rgb}{0, 0, 0.8}
\definecolor{blue}{rgb}{0,0,1}
\definecolor{darkgreen}{rgb}{0,0.40,0}
\definecolor{firebrick}{rgb}{0.698,0.133,0.133}

\definecolor{colorA}{rgb}{1,0,0}
\definecolor{colorB}{rgb}{0,0.3,1}
\definecolor{colorC}{rgb}{0.9,0.8,0.2}
\definecolor{colorD}{rgb}{0,0.65,0}
\definecolor{lesslightgray}{rgb}{0.5,0.5,0.5}
\definecolor{light-gray}{gray}{0.95}

\let\tilde\widetilde
\let\hat\widehat

\newcommand{\EE}{\mathbb{E}}

\newcommand{\calD}{\mathcal{D}}

\newcommand{\calO}{\mathcal{O}}

\newcommand{\calT}{\mathcal{T}}

\newcommand{\RR}{\mathbb{R}}
\newcommand{\sT}{ \mathsf{T} }

\newcommand{\diag}{\mathop{\rm{diag}}}

\newcommand{\Softmax}{\mathop{\rm{Softmax}}}

\def\P{\mathbb{P}}
\def\R{\mathbb{R}}
\def\B{\mathbb{B}}

\let\cite\citep 

\usepackage{amsthm}
\makeatletter
\def\th@remark{%
  \thm@headfont{\bfseries}%
  \normalfont %
  \thm@preskip\topsep \divide\thm@preskip\tw@
  \thm@postskip\z@ %
}
\makeatother
\usepackage[many]{tcolorbox}
\theoremstyle{definition}
\newtheorem{theorem}{Theorem}
\tcolorboxenvironment{theorem}{
  breakable,
  colback=black!10,
  colframe=white,%
  width=\dimexpr\linewidth+10pt\relax,%
  enlarge left by=-5pt,%
  enlarge right by=-5pt,%
  boxsep=5pt,%
  boxrule=0pt,
  left=0pt,right=0pt,top=0pt,bottom=0pt,
  sharp corners,
  before skip=\topsep,
  after skip=\topsep
}
\newtheorem{lemma}{Lemma}
\tcolorboxenvironment{lemma}{
  breakable,
  colback=black!10,
  colframe=white,%
  width=\dimexpr\linewidth+10pt\relax,%
  enlarge left by=-5pt,%
  enlarge right by=-5pt,%
  boxsep=5pt,%
  boxrule=0pt,
  left=0pt,right=0pt,top=0pt,bottom=0pt,
  sharp corners,
  before skip=\topsep,
  after skip=\topsep
}
\newtheorem{corollary}{Corollary}[theorem]
\tcolorboxenvironment{corollary}{
  breakable,
  colback=black!10,
  colframe=white,%
  width=\dimexpr\linewidth+10pt\relax,%
  enlarge left by=-5pt,%
  enlarge right by=-5pt,%
  boxsep=5pt,%
  boxrule=0pt,
  left=0pt,right=0pt,top=0pt,bottom=0pt,
  sharp corners,
  before skip=\topsep,
  after skip=\topsep
}

\theoremstyle{definition}
\newtheorem{definition}{Definition}

\tcolorboxenvironment{definition}{
  breakable,
  colback=black!10,
  colframe=white,%
  width=\dimexpr\linewidth+10pt\relax,%
  enlarge left by=-5pt,%
  enlarge right by=-5pt,%
  boxsep=5pt,%
  boxrule=0pt,
  left=0pt,right=0pt,top=0pt,bottom=0pt,
  sharp corners,
  before skip=\topsep,
  after skip=\topsep
}
\newtheorem{remark}{Remark}

\tcolorboxenvironment{assumption}{
  breakable,
  colback=black!10,
  colframe=white,%
  width=\dimexpr\linewidth+10pt\relax,%
  enlarge left by=-5pt,%
  enlarge right by=-5pt,%
  boxsep=5pt,%
  boxrule=0pt,
  left=0pt,right=0pt,top=0pt,bottom=0pt,
  sharp corners,
  before skip=\topsep,
  after skip=\topsep
}
\newtheorem{problem}{Problem}
\tcolorboxenvironment{problem}{
  breakable,
  colback=black!10,
  colframe=white,%
  width=\dimexpr\linewidth+10pt\relax,%
  enlarge left by=-5pt,%
  enlarge right by=-5pt,%
  boxsep=5pt,%
  boxrule=0pt,
  left=0pt,right=0pt,top=0pt,bottom=0pt,
  sharp corners,
  before skip=\topsep,
  after skip=\topsep
}

\newtheorem{question}{Question}
\tcolorboxenvironment{question}{
  breakable,
  colback=black!10,
  colframe=white,%
  width=\dimexpr\linewidth+0pt\relax,%
  enlarge left by=0pt,%
  enlarge right by=0pt,%
  boxsep=5pt,%
  boxrule=0pt,
  left=0pt,right=0pt,top=0pt,bottom=0pt,
  sharp corners,
  before skip=\topsep,
  after skip=\topsep
}
\crefname{question}{Question}{Questions}

\crefname{theorem}{Theorem}{Theorems}
\crefname{proposition}{Proposition}{Propositions}
\crefname{lemma}{Lemma}{Lemmas}
\crefname{corollary}{Corollary}{Corollaries}
\crefname{definition}{Definition}{Definitions}
\crefname{assumption}{Assumption}{Assumptions}
\crefname{remark}{Remark}{Remarks}
\crefname{problem}{Problem}{Problems}
\crefname{property}{Property}{property}

\numberwithin{equation}{section}

\usepackage{lipsum}

\newcommand*{\annot}[1]{\tag*{\footnotesize{\textcolor{black!50}{\big(#1\big)}}}}

\makeatletter
\let\save@mathaccent\mathaccent
\newcommand*\if@single[3]{%
    \setbox0\hbox{${\mathaccent"0362{#1}}^H$}%
    \setbox2\hbox{${\mathaccent"0362{\kern0pt#1}}^H$}%
    \ifdim\ht0=\ht2 #3\else #2\fi
}
\newcommand*\rel@kern[1]{\kern#1\dimexpr\macc@kerna}
\newcommand*\widebar[1]{\@ifnextchar^{{\wide@bar{#1}{0}}}{\wide@bar{#1}{1}}}
\newcommand*\wide@bar[2]{\if@single{#1}{\wide@bar@{#1}{#2}{1}}{\wide@bar@{#1}{#2}{2}}}
\newcommand*\wide@bar@[3]{%
    \begingroup
    \def\mathaccent##1##2{%
        \let\mathaccent\save@mathaccent
        \if#32 \let\macc@nucleus\first@char \fi
        \setbox\z@\hbox{$\macc@style{\macc@nucleus}_{}$}%
        \setbox\tw@\hbox{$\macc@style{\macc@nucleus}{}_{}$}%
        \dimen@\wd\tw@
        \advance\dimen@-\wd\z@
        \divide\dimen@ 3
        \@tempdima\wd\tw@
        \advance\@tempdima-\scriptspace
        \divide\@tempdima 10
        \advance\dimen@-\@tempdima
        \ifdim\dimen@>\z@ \dimen@0pt\fi
        \rel@kern{0.6}\kern-\dimen@
        \if#31
        \overline{\rel@kern{-0.6}\kern\dimen@\macc@nucleus\rel@kern{0.4}\kern\dimen@}%
        \advance\dimen@0.4\dimexpr\macc@kerna
        \let\final@kern#2%
        \ifdim\dimen@<\z@ \let\final@kern1\fi
        \if\final@kern1 \kern-\dimen@\fi
        \else
        \overline{\rel@kern{-0.6}\kern\dimen@#1}%
        \fi
    }%
    \macc@depth\@ne
    \let\math@bgroup\@empty \let\math@egroup\macc@set@skewchar
    \mathsurround\z@ \frozen@everymath{\mathgroup\macc@group\relax}%
    \macc@set@skewchar\relax
    \let\mathaccentV\macc@nested@a
    \if#31
    \macc@nested@a\relax111{#1}%
    \else
    \def\gobble@till@marker##1\endmarker{}%
    \futurelet\first@char\gobble@till@marker#1\endmarker
    \ifcat\noexpand\first@char A\else
    \def\first@char{}%
    \fi
    \macc@nested@a\relax111{\first@char}%
    \fi
    \endgroup
    }
\makeatother
\let\bar\widebar

\makeatletter
\newcommand*{\redefinesymbolwitharg}[1]{%
  \expandafter\let\csname ltx#1\expandafter\endcsname\csname #1\endcsname
  \@namedef{#1}{\@ifnextchar{^}{\@nameuse{#1@}}{\@nameuse{#1@}^{}}}%
  \expandafter\def\csname #1@\endcsname^##1##2{%
     \csname ltx#1\endcsname\ifx!##1!\else^{##1}\fi\mathopen{}\mathclose\bgroup\left(##2\aftergroup\egroup\right)
     }%
}
\makeatother
\redefinesymbolwitharg{sin}
\redefinesymbolwitharg{cos}

\newcommand{\vertiii}[1]{{\left\vert\kern-0.25ex\left\vert\kern-0.25ex\left\vert #1 
\right\vert\kern-0.25ex\right\vert\kern-0.25ex\right\vert}}

%% file: title.tex
In-Context Deep Learning via Transformer Models

%% file: 0abstract.tex
We investigate the transformer's capability to simulate the training process of deep models via in-context learning (ICL), i.e., \textit{in-context deep learning}.
Our key contribution is providing a positive example of using a transformer to train a deep neural network by gradient descent in an implicit fashion via ICL. 
Specifically, we provide an explicit construction of a $(2N+4)L$-layer transformer capable of simulating $L$ gradient descent steps of an $N$-layer ReLU network through ICL.
We also give the theoretical guarantees for the approximation within any given error and the convergence of the ICL gradient descent.
Additionally, we extend our analysis to the more practical setting using $\Softmax$-based transformers. 
We validate our findings on synthetic datasets for 3-layer, 4-layer, and 6-layer neural networks.
The results show that ICL performance matches that of direct training.

%% file: 1intro.tex
We study transformers' ability to simulate the training process of deep models.
This analysis is not only practical but also timely.
On one hand, transformers and deep models   \cite{brown2020language, radford2019language} 
are so powerful, popular and form a new machine learning paradigm --- foundation models.
These large-scale machine learning models, trained on vast data, provide a general-purpose foundation for various tasks with minimal supervision \cite{team2023gemini, touvron2023llama, zhang2022opt}.
On the other hand, the high cost of pretraining these models often makes them prohibitive outside certain industrial labs \cite{jiang2024mixtral, bi2024deepseek, achiam2023gpt}. 
In this work, we aim to advance the ``one-for-many'' modeling philosophy of foundation model paradigm \cite{bommasani2021opportunities} by considering the following research problem:
\begin{question}\label{q:central}
    Is it possible to train one deep model with the ICL of another foundation model?
\end{question}
The implication of \cref{q:central} is profound: if true, one foundation model could lead to many others without pertaining.
In this work, we provide an affirmative example for \cref{q:central}.
Specifically, we show that transformers are capable of simulating the training of a deep ReLU-based feed-forward neural network with provable guarantees through ICL.
Our analysis assumes that we have well-pretrained the transformer using the data generated by the deep network.
We require the deep network to maintain consistent hyperparameters (e.g., model width and depth) during the pretraining and testing. 
However, during the testing, we vary the parameter distribution and input data distribution of the deep network to generate data for the transformer.

In ICL, the models learn to solve new tasks during inference by using task-specific examples provided as part of the input prompt, rather than through parameter updates \cite{wei2023larger,bubeck2023sparks,achiam2023gpt,bai2023transformers, min2022rethinking,garg2022can}.
Unlike standard supervised learning, ICL enables models to adapt to new tasks during inference using only the provided examples.
In this work, the new task of our interest is algorithmic approximation via ICL \cite{bai2023transformers,zhang2023trained,wangcontext}.
Specifically, we aim to use transformer's ICL capability to replace/simulate the standard supervised training algorithms for $N$-layer networks. 
To be concrete, we formalize the learning problem of how transformers learn (i) a given function and (ii) a machine learning algorithm (e.g., gradient descent) via ICL, following \cite{bai2023transformers}.

\textbf{(i) ICL for Function $f$.}
Let $f: \R^d \rightarrow \R$ be the function of our interest.
Suppose we have a dataset $\mathcal{D}_n \coloneqq \left\{ (x_i, y_i) \right\}_{i \in [n]} $, where $\left\{ x_i \right\}_{i \in [n]} \subseteq \R^d$ and $\left\{ y_i \right\}_{i \in [n]} \subseteq \R$ are the input and output of $f$, respectively. 
Let $x_{n+1}$ be the test input.
The goal of ICL is to use a transformer, denoted by $\calT$, to predict $y_{n+1}$ based on the test input and the in-context dataset autoregresively: $\hat{y}_{n+1}\sim \calT(\calD_n,x_{n+1})$.
The goal is for the prediction $\hat{y}_{n+1}$ to be close to $y_{n+1} =f(x)$.

\textbf{(ii) ICL for Gradient Descent of a Parametrized Model $f(w,\cdot)$.}
\citet{bai2023transformers} generalize \textbf{(i)} to include algorithmic approximations of Gradient Descent (GD) training algorithms and explore how transformers simulate gradient descent during inference without parameter updates.
They term the simulated GD algorithm ``In-Context Gradient Descent (ICGD).''
In essence, ICGD enables transformers to approximate gradient descent on a loss function $L_n(w)$ for a parameterized model $f(w, \cdot)$ based on a dataset $\mathcal{D}_n$. 
Traditional gradient descent updates $w$ iteratively as $w_{t+1} = w_t - \eta \nabla L_n(w_t)$. 
In contrast, ICGD uses a transformer $\mathcal{T}$ to simulate these updates within a forward pass. 
Given example data $\mathcal{D}_n$ and test input $x_{n+1}$, the transformer performs gradient steps in an implicit fashion by inferring parameter updates through its internal representations, using input context without explicit weight changes.
Please see \cref{sec:ICGD} for explicit formulation.

In this work, we investigate the case where $f(w, \cdot)$ is a deep feed-forward neural network.
We defer the detailed problem setting to \cref{sec:preliminaries}.
In comparison to standard ICGD \cite{bai2023transformers}, ICGD for deep feed-forward networks is not trivial.
This is due to two technical challenges: 
\begin{itemize}[leftmargin=2.4em, itemsep=-0.3pt]
    \item [(C1)] 
    Analytical feasibility of gradient computation for these thick networks.
    \item [(C2)] 
    Explicit construction capable of approximating ICGD for such layers and their gradients.
\end{itemize}
To this end, 
we present the first explicit expression for gradient computation of $N$-layer feed-forward network  (\cref{lem:decomposition_gd_m}).
Importantly, its term-by-term tractability provides key insights for the detailed construction of a specific transformer to train this network via ICGD (\cref{thm:icl_gd_m}). 

\textbf{Contributions.}
Our contributions are threefold:
\begin{itemize}[itemsep=-0.7pt]
    \item \textbf{Approximation by ReLU-Transformer.}
    For simplicity, we begin with the ReLU-based transformer.
    For a broad class of smooth empirical risks, we construct a $(2N+4)L$-layer transformer to approximate $L$ steps of in-context gradient descent on the $N$-layer feed-forward networks with the same input and output dimensions (\cref{thm:icl_gd_m}).
    We then extend this to accommodate varying dimensions (\cref{thm:icl_gd_m_io}).
    We also provide the theoretical guarantees for the approximation within any given error (\cref{coro:error_ICGD_m}) and the convergence of the ICL gradient descent (\cref{lem:error_general_m}).
    \item \textbf{Approximation by Softmax-Transformer.}
    We extend our analysis to the $\Softmax$-transformer to better reflect realistic applications.
    The key technique is to ensure a qualified approximation error at each point to achieve universal approximation capabilities of the $\Softmax$-based Transformer (\cref{lemma:univerality_softmax}).
    We give a construction of a $4L$-layer $\Softmax$ transformer to approximate $L$ steps of gradient descent, and guarantee the approximation and the convergence (\cref{thm:icgd_soft_main}).
    \item \textbf{Experimental Validation.}
    We validate our theory with ReLU- and $\Softmax$-transformers, specifically, ICGD for the $N$-layer networks (\cref{thm:icl_gd_m}, \cref{thm:icgd_soft_main}, and \cref{thm:icl_gd_m_io}). 
    We assess the ICL capabilities of transformers by training 3-, 4-, and 6-layer networks in  \cref{sec:experiments}.
    The numerical results show that the performance of ICL matches that of training $N$-layer networks.
    However, a minor limitation is that the trained transformers do not always achieve the theoretical construction.
\end{itemize}
\textbf{Organization.}
We present our main results in \cref{thm:icl_gd_m}. \cref{sec:preliminaries} covers the preliminaries. \cref{sec:method} presents the problem setup and the ICL approximation of GD steps for an $N$-layer feed-forward network with both ReLU-Transformer and $\Softmax$-Transformer.
\cref{sec:experiments} presents the experimental results, with additional details in \cref{app:sec_exp}. 
The appendix includes related work (\cref{sec:related_works}), detailed proofs for \cref{sec:method} (\cref{app:sec_proof_main}), and an application to train diffusion models via ICL (\cref{app:sec_icl_diff}).

\textbf{Notations.}
We use lower case letters to denote vectors and  upper case letters to denote matrices.
The index set $\{ 1, ..., I \}$ is denoted by $[ I ]$, where $I \in \mathbb{N}^+$. 
For any matrices $A \in \R^{n \times n}$, let $\ell_p$ norm of $A$ be induced by vector $\ell_p$-norm, defined as 
$\|A\|_p := \sup \{ \|Ax\|_p : x \in \R^n ~{\rm with}~ \|x\|_p = 1 \}$. 
We use $A[i,j]$ to denote the element in $i$-th row and $j$-th column of matrix $A$.
For any matrices $A \in \R^{m \times n}$ and $B \in \R^{m \times n}$,  let $\odot$ denotes the Hadamard product: $(A \odot B)[i,j] := A[i,j] \cdot B[i,j]$.
For any matrices $A \in \R^{m \times n}$ and $B \in \R^{p \times q}$, let $\otimes$ denote the Kronecker product: 
\begin{align*}
    A \otimes B :=
    \begin{bmatrix}
        &A[1,1] B &\cdots &A[1,n] B \\
        &\vdots &\ddots &\vdots \\
        &A[m,1] B &\cdots &A[m,n] B
    \end{bmatrix}.
\end{align*}

%% file: 2preliminary.tex
\label{sec:ICGD}
We present the ideas we built upon:
In-Context Gradient Descent (ICGD).

\textbf{(i) ICL for Function $f$.}
Let $f: \R^d \rightarrow \R$ be the function of our interest.
Suppose we have a dataset $\mathcal{D}_n \coloneqq \left\{ (x_i, y_i) \right\}_{i \in [n]} $, where $\left\{ x_i \right\}_{i \in [n]} \subseteq \R^d$ and $\left\{ y_i \right\}_{i \in [n]} \subseteq \R$ are the input and output of $f$, respectively. 
Let $x_{n+1}$ be the test input.
The goal of ICL is to use a transformer, denoted by $\calT$, to predict $y_{n+1}$ based on the test input and the in-context dataset autoregresively: $\hat{y}_{n+1}\sim \calT(\calD_n,x_{n+1})$.
For convenience in our analysis, we adopt the ICL notation from \cite{bai2023transformers}. Specifically, we shorthand $(\mathcal{D}_n, x_{n+1})$ into an input sequence (i.e., prompt) of length $n+1$ and represent it as a compact matrix $H \in \mathbb{R}^{D \times (n+1)} := [h_1, \ldots, h_{n+1}]$ in the form: 
\begin{align}
    \label{eqn:input}
    H \coloneqq & ~ \begin{bmatrix}
    x_1 & x_2 & \cdots & x_n & x_{n+1} \\
    y_1 & y_2 & \cdots & y_n & 0 \\
    q_1 & q_2 & \cdots & q_n & q_{n+1}
    \end{bmatrix} \in \mathbb{R}^{D \times (n+1)},\notag \\
    q_i \coloneqq & ~
    \begin{bmatrix}
    0_{D-(d+3)} \\
    1 \\
    t_i
    \end{bmatrix} 
    \in \mathbb{R}^{D-(d+1)}.
\end{align}
We use $q_i$ to fill in the remain $D-(d+1)$ entries in addition to $x_i\in \R^d$ and $y_i\in\R$. 
The last entry $t_i\coloneqq\one(i < n+1)$ of $q_i$ is the position indicator to distinguish the $n$ in-context examples and the test data.
The \textbf{problem of ``ICL for $f$''} is to show the existence of a transformer $\mathcal{T}$ that, when given $H$, outputs $\mathcal{T}(H)\in \mathbb{R}^{D \times (n+1)}$ of the same shape, and the ``$(d+1, n+1)$ entry of $\mathcal{T}(H)$'' provides the prediction $\hat{y}_{n+1}$.
The goal is for the prediction $\hat{y}_{n+1}$ to be close to $y_{n+1} =f(x)$
measured by some proper loss.

\textbf{(ii) ICL for Gradient Descent of a Parametrized Model $f(w,\cdot)$.}
We aim to use ICL to simulate the standard supervised training procedure for $N$-layer neural networks. 
To achieve this, we introduce the concept of In-Context Gradient Descent (ICGD) for a parameterized model.
Consider a  machine learning model $f(w,\cdot): \R^{D_w} \times \R^d \rightarrow \R^d$, parametrized by $w \in \R^{D_w}$.
Given a dataset $\mathcal{D}_n \coloneqq \left\{ (x_i, y_i) \right\}_{i \in [n]} \overset{\text{iid}}{\sim} \P$, a typical learning task is to find parameters $w^\star$ such that $f(w^\star,\cdot)$ becomes closest to the true data distribution $\P$. 
Then, for any test input $x_{n+1}$, we predict:
$\hat{y}_{n+1} = f(w^\star, x_{n+1})$.
To find $w^\star$, \citet{bai2023transformers} configure a transformer to implement gradient descent on $f(w, \cdot)$ through ICL, simulating optimization algorithms  during inference  without explicit parameter updates.
We formalize this \textbf{In-Context Gradient Descent (ICGD)} problem: using a pretrained model to simulate gradient descent on $f(w, \cdot)$ w.r.t. the provided context $(\calD_n, x_{n+1})$.

\begin{problem}[In-Context Gradient Descent (ICGD) on Model $f(w, \cdot)$ \cite{bai2023transformers}]
\label{prob:icgd}
Let $\epsilon > 0$ and $L \geq 1$. 
Consider a machine learning model $f(w, x): \mathbb{R}^{D_w} \times \mathbb{R}^d \rightarrow \mathbb{R}^d$ parameterized by $w \in \mathbb{R}^{D_w}$. Given a dataset $\mathcal{D}_n \coloneqq \left\{ (x_i, y_i) \right\}_{i \in [n]} \overset{\text{iid}}{\sim} \P$ with $(x_i, y_i) \in \mathbb{R}^d \times \mathbb{R}^d$, define the empirical risk function:
\begin{align}\label{eqn:loss}
\mathcal{L}_n(w) \coloneqq \frac{1}{2n} \sum_{i=1}^n \ell(f(w, x_i), y_i),
\end{align}
where $\ell: \mathbb{R}^d \times \mathbb{R}^d \rightarrow \mathbb{R}$ is a loss function.
Let $\mathcal{W} \subseteq \mathbb{R}^{D_w}$ be a closed domain, and  ${\rm Proj}_{\mathcal{W}}$ denote the projection onto $\mathcal{W}$. 
The problem of ``ICGD on model $f(w,\cdot)$'' is to find a transformer $\mathcal{T}$ with $L$ blocks, each approximating one step of gradient descent using $T$ layers. 
For any input $H^{(0)}\in\R^{D \times (n+1)}$ in the form of \eqref{eqn:input}, the transformer $\mathcal{T}(H^{(0)})$ approximates $L$ steps of gradient descent. 
Specifically, for $l \in [L]$ and $i \in [n+1]$, the output at layer $Tl$ is: $h_i^{(Tl)} = [x_i; y_i; \bar{w}^{(l)}; \mathbf{0}; 1; t_i]$,
where, with $\bar{w}^{(0)} = \mathbf{0}$,
\begin{align}\label{eqn:w_recursion}
\bar{w}^{(l)} = {\rm Proj}_{\mathcal{W}} \left( \bar{w}^{(l-1)} - \eta \left( \nabla \mathcal{L}_n(\bar{w}^{(l-1)}) + \epsilon^{(l-1)} \right) \right)  
\end{align}
is updated recursively,
and $\| \epsilon^{(l-1)} \|_2 \leq \epsilon$ represents the approximation error at step $l - 1$.

\end{problem}

\cref{prob:icgd} aims to find a transformers $\mathcal{T}$ to  perform $L$ steps gradient
descent on loss $\mathcal{L}_n(w)$ in an implicit fashion (i.e., no explicit parameter update). 
More precisely, \citet{bai2023transformers} configure $\mathcal{T}$ with $L$ identical blocks, each approximating one gradient descent step using $T$ layers. 
In this work, we investigate the case where $f(w, \cdot)$ is an ``$N$-layer neural network.''

\textbf{Transformer.}
We defer the standard definition of transformer to \cref{sec:transformer}.

%% file: 3N_NN_approx.tex
We now show that transformers is capable of implementing gradient descent on $N$-layer neural networks through ICL. 
In \cref{sec:L_NN_grad_m}, we define the $N$-layer ReLU neural network and state its ICGD problem.
In \cref{sec:explicit_grad}, we
derive explicit gradient descent expression for $N$-layer NN. 
In \cref{sec:ICL_for_LNN_m}, we construct ReLU-Transformer executing gradient descent on $N$-layer NN via ICL.
In \cref{sec:softmax}, we show the existence of $\Softmax$-Transformer capable of performing in-context gradient descent on $N$-layer NN.

\subsection{Problem Setup: ICGD for \texorpdfstring{$N$}{}-Layer Neural Networks}
\label{sec:L_NN_grad_m}
To begin, we introduce the construction of our $N$-Layer Neural Network which we aims to implement gradient descent on its empirical loss function. 

\begin{definition}
[$N$-Layer Neural Network]
\label{def:N_nn_m}
An $N$-Layer Neural Network comprises $N-1$ hidden layers and $1$ output layer, all constructed similarly.
Let $r:\R \rightarrow \R$ be the activation function.
For the hidden layers: 
for any $i \in [n+1], j \in 
[N-1]$, and $ k \in [K]$, 
the output for the first $j$ layers w.r.t. input $x_i \in \R^d$, denoted by ${\rm pred}_h(x_i;j)\in \R^K$, is defined as recursive form:
\begin{align*}
    {\rm pred}_h 
    (x_i ; 1)[k]
    := & ~ 
    r(v_{1_{k}}^{\top} x_i), \\
    {\rm pred}_h 
    (x_i ; j)[k]
    :=  & ~ 
    r(v_{j_{k}}^{\top} {\rm pred}_h 
    (x_i ; j-1)),  
    \end{align*}  
where  $v_{1_k} \in \R^d$  and $v_{j_k}\in\R^K$ for $j\in 
\{2,\ldots,N-1\}$ are the 
$k$-th parameter vectors in the first layer and the $j$-th layer, respectively.
For the output layer ($N$-th layer), 
the output for the first $N$ layers (i.e the entire neural network) w.r.t. input $x_i \in \R^d$, denoted by ${\rm pred}_o(x_i ;w, N) \in \R^d$, is defined for any $k \in [d]$ as follows:
\begin{align}
    {\rm pred}_o(x_i ;w, N)[k]
    := r(v_{N_{k}}^{\top} 
    {\rm pred}_h 
    (x_i ; N-1)),
\end{align} 
where $v_{N_k}\in\R^K$ are the $k$-th parameter vectors in the $N$-th layer and $w \in \R^{2dK + (N-2)K^2}$ denotes the vector containing all parameters in the neural network, 
\begin{align}   
\label{eqn:w_N_m}
w :=  \begin{bmatrix}
        v_{1_1}^\top, \ldots, 
        v_{1_K}^\top, \ldots, v_{j_k}^\top, \ldots, 
        v_{{N}_1}^\top, \ldots, 
        v_{{N}_d}^\top 
    \end{bmatrix}^\top.
\end{align}
\end{definition}

\begin{remark}[Prediction Function for $j$-th layer on $i$-th Data: $p_i(j)$]
\label{remark:p_ij_m}
For simplicity,  we abbreviate the output from the first $j$-th layer of the $N$-layer neural networks NN with input $x_i$ as $p_i(j)$,
\begin{align}
    \label{eqn:p_m}
    p_i(j) :=
    \begin{cases}
        x_i \in \R^{d}, &\text{ for } j=0 \\
        {\rm pred}_h (x_i ; j) \in \R^{K}, &\text{ for } j \in [N-1]\\
        {\rm pred}_o (x_i; w, N)
        \in \R^{d}, &\text{ for } j = N.
    \end{cases}
\end{align}
Additionally, we define
\begin{align*}
   p_i := [p_i(1); \ldots; p_i(N)] \in 
\R^{(N-1)K+d} .
\end{align*} 
\end{remark}
We formalize the problem of using a transformer to simulate gradient descent algorithms for training the $N$-layer NN defined in \cref{def:N_nn_m}, by optimizing loss  \eqref{eqn:loss}.
Specifically, we consider the ICGD (\cref{prob:icgd}) with the parameterized model $f(w, \cdot) \coloneqq {\rm pred}_o(\cdot; w, N)$.

\begin{problem}[ICGD on $N$-Layer Neural Networks]
    \label{prob:icl_n_m}
    Let the $N$-layer neural networks, activation function $r$, and prediction function $p_i(j)$ for all layers follow \cref{def:N_nn_m} and \cref{remark:p_ij_m}. 
    Assume we under the identical setting as \cref{prob:icgd}, considering model $f(w,\cdot) := {\rm pred}_o(\cdot; w, N)$
    and specifying $\mathcal{W}$ is a closed domain such that for any $j \in [N-1]$ and $k \in [K]$,
    \begin{align}
    \label{eqn:domain_w_m}
        \mathcal{W} \subset \left\{
        w = [v_{j_k}] \in \mathbb{R}^{D_N} : \|v_{j_k}\|_2 \leq B_v
        \right\}.      
    \end{align}
    The problem of ``ICGD on $N$-layer neural networks'' is to find a $TL$ layers transformer $\mathcal{T}$, 
    capable of implementing  
    $L$ steps gradient descent as in \cref{prob:icgd}. 
\end{problem}

\begin{remark}[Why Bounded Domain $\mathcal{W}$?]
    For using a sum of ReLU to approximate functions like $r$, which is illustrated in the consequent section, we need to avoid gradient exploding.
    Therefore, we require $\mathcal{W}$ to be a bounded domain, and utilize ${\rm Proj}_{\mathcal{W}}$ to project $w$ into bounded domain $\mathcal{W}$.
\end{remark}

\subsection{Explicit Gradient Descent of \texorpdfstring{$N$}{}-Layer Neural Networks}
\label{sec:explicit_grad}

Intuitively, \cref{prob:icl_n_m} asks whether there exists a transformer capable of simulating the gradient descent algorithm on the loss function of an $N$-layer neural network. 
We answer \cref{prob:icl_n_m} by providing an explicit construction for such a transformer $\calT$ in \cref{thm:icl_gd_m}.
To facilitate our proof, we first introduce the necessary notations for explicit expression of the gradient $\nabla_w \mathcal{L}_n(w)$.

\begin{definition}[Abbreviations]
\label{def:abbv_m}
Fix $i \in [n+1]$, and consider an $N$-layer neural network with activation function $r$ and prediction function $p_i(j)$ as defined in \cref{def:N_nn_m}.

\begin{itemize}[leftmargin=0.5em]
    \item Let $D_j \in \mathbb{R}$ denote the total number of parameters in the first $j$ layers. By \eqref{eqn:w_N_m}, we have:
    \begin{align*}
    D_j = \begin{cases}
        0, & j = 0 \\
        dK, & j = 1 \\
        (j-1)K^2 + dK, & 2 \leq j \leq N-1 \\
        (N-2)K^2 + 2dK, & j = N.
    \end{cases}
    \end{align*}

    \item The parameter vector $w := 
    \begin{bmatrix}
        v_{1_1}^\top, \ldots, 
        v_{1_K}^\top, \ldots,
        v_{{N-1}_1}^\top, \ldots, 
        v_{{N-1}_K}^\top,  
        v_{{N}_1}^\top, \ldots, 
        v_{{N}_d}^\top 
    \end{bmatrix}^\top$  
    follows \eqref{eqn:w_N_m}. 
    Define $ \phi_i \coloneqq \left(\pdv
    { \ell(p_i(N), y_i)}{p_i(N)} \cdot \pdv{p_i(N)}{w} \right)^\top \in \mathbb{R}^{D_N}$. 
    For any $j \in [N]$, let $A_i(j)$ denote the derivative of $\ell(p_i(N), y_i)$ with respect to the parameters in the $j$-th layer: $ A_i(j) =\phi_i[D_{j-1}:D_j]$, 
    where $\phi_i[a:b]$ selects elements from the $a$-th to $b$-th position in $\phi_i$.

    \item For activation function $r(t)$, let $r'(t)$ be its derivative. Define $r'_i(j) \in \mathbb{R}^K$ as:
    \begin{align*}
    r'_i(j)[k] := r'(v_{{j+1}_k}^\top p_i(j)).
    \end{align*}

    \item Define $r'_i := [r'_i(0); \ldots; r'_i(N-1)]$ and $R_i(j)$ as:
    \begin{align*}
    R_i(j) :=   
    \begin{cases}
        \mathrm{diag}\{r'(v_{{j+1}_1}^\top p_i(j)), \ldots, r'(v_{{j+1}_K}^\top p_i(j))\}, \; j \leq N-2 \\
        \mathrm{diag}\{r'(v_{{j+1}_1}^\top p_i(j)), \ldots, r'(v_{{j+1}_d}^\top p_i(j))\} , \; j=N-1.
    \end{cases}
    \end{align*}
    where $ R_i(j) \in \mathbb{R}^{K \times K}$ for  $j\in \{0,\dots,N-2\}$ and $R_i(j)\in \mathbb{R}^{d \times d}$ for $j = N-1$.

    \item For any $j \in [N]$, let $V_j$ denote the parameters in the $j$-th layer as:
    \begin{align*}
        V_j := 
        \begin{cases}
            \begin{bmatrix} v_{1_1}, \ldots, v_{1_K} \end{bmatrix}^\top \in \R^{K \times d}, & j=1 \\
            \begin{bmatrix} v_{j_1}, \ldots, v_{j_K} \end{bmatrix}^\top \in \R^{K \times K}, & j \in {2,\ldots,N-1} \\
            \begin{bmatrix} v_{N_1}, \ldots, v_{N_d} \end{bmatrix}^\top \in \R^{d \times K}, & j=N.
        \end{cases}
    \end{align*}
\end{itemize}
\end{definition}

\cref{def:abbv_m} splits the gradient of $\mathcal{L}_n(w)$ into $N$ parts.
This makes $\nabla_w \mathcal{L}_n(w)$ more interpretable and tractable, since all parts follows a recursion formula according to chain rule.
With above notations, we calculate the gradient descent step \eqref{eqn:w_recursion} of $N$-layer neural network as follows:
\begin{lemma}
[Decomposition of One Gradient Descent Step]
\label{lem:decomposition_gd_m}
    Fix any $B_v, \eta > 0$.
    Suppose loss function $\mathcal{L}_n(w)$ on $n$ data points $\{(x_i,y_i)\}_{i \in [n]}$ follows \eqref{eqn:loss}.
    Suppose closed domain $\mathcal{W}$ and projection function
    ${\rm Proj}_{\mathcal{W}}(w)$ follows \eqref{eqn:domain_w_m}. 
    Let $A_i(j), r'_i(j), R_i(j), V_j$ be as defined in \cref{def:abbv_m}.
    Then the explicit form of gradient $\nabla \mathcal{L}_n(w)$ becomes
    \begin{align}\label{eqn:decomposition_gd_m}
        \nabla \mathcal{L}_n(w) =  
        \frac{1}{2n} \sum_{i=1}^{n}
         \begin{bmatrix}
            A_i(1) \\
            \vdots \\
            A_i(N)
         \end{bmatrix},
    \end{align}
    where $A_i(j)$ denote the derivative of $\ell(p_i(N), y_i)$ with respect to the parameters in the $j$-th layer, 
    \begin{align*}
    A_i(j) = 
    \begin{cases}
    (R_i(N-1)  V_{N}  \ldots  R_i(j-1) 
    \begin{bmatrix}
    \textbf{I}_{K \times K} \otimes 
    p_i(j-1)^{\top} 
    \end{bmatrix}
    )^{\top} \cdot (\pdv {\ell(p_i(N), y_i)}{p_i(N)})^\top, \quad j \neq N \\[.5em]
    (R_i(N-1) \cdot
    \begin{bmatrix}
    \textbf{I}_{d \times d} \otimes 
    p_i(N-1)^{\top} 
    \end{bmatrix}
    )^{\top} \cdot (\pdv {\ell(p_i(N), y_i)}{p_i(N)})^\top, \quad j = N.
    \end{cases}
    \end{align*}
\end{lemma}
\begin{proof}[Proof Sketch]
Using the chain rule and product rule, we decompose the gradient as follows:
$\nabla_w \mathcal{L}_n(w) = \frac{1}{2n} \sum_{i=1}^N [\pdv{p_i(N)}{w}]^\top \cdot
[\pdv{ \ell(p_i(N), y_i)}{p_i(N)}]^\top $.
Thus, we only need to compute 
$\pdv{p_i(N)}{w}$.
By \cref{def:N_nn_m} and the chain rule, we prove that $\pdv{p_i(N)}{w}$ satisfies the recursive formulation \eqref{eqn:A_k_p3_m}. 
Combining these, we derive
the explicit form of gradient $\nabla_w \mathcal{L}_n(w)$,
and the gradient step follows directly.  
Please see \cref{proof:lem:decomposition_gd_m} for a detailed proof.
\end{proof}
It is hard to calculate the elements in $A_i(j)$ in a straightforward mannar, we calculate each parts of it successively.
We define the intermediate terms $s_i(j)$ and $u$ as follows
\begin{definition}[Definition of intermediate terms]
    \label{def:s_m}
    Let $A_i(j), r'_i(j), R_i(j), V_j$ be as defined in \cref{def:abbv_m}.
    For any $t,y \in \R^d$, we define vector function $u(t,y):= (\pdv {\ell(t, y)}{t})^\top : \R^{d} \times \R^{d} \rightarrow \R^d$. 
    Moreover, for any $j \in [N], i\in [n+1]$, we define $s_i(j)$ as
    \begin{align*}
        s_i(j) := \begin{cases}
            R_i(j-1) V_{j+1}^{\top} \ldots R_i(N-2) V_{N}^{\top} \cdot R_i(N-1)
            \cdot u(p_i(N),y_i) \in \R^K, \quad & j \neq N \\
            R_i(N-1) \cdot u(p_i(N),y_i)   \in \R^d, \quad \hspace{1.9em} & j=N.
        \end{cases}
    \end{align*}
\end{definition}
Let $\odot$ denotes Hadamard product. 
For any $j \in [N-1], i \in [N+1]$, \cref{def:s_m} leads to 
\begin{align}
    s_i(j) = & ~ r'_i(j-1) \odot (V_{j+1}^{\top} \cdot s_i(j+1)),
    \label{eqn:s_recursion_m}
    \end{align}
Moreover, by \cref{def:s_m}, it holds
\begin{align}
    \label{eqn:a_and_s}
    A_i(j) = 
    \begin{cases}
        \begin{bmatrix}
        \textbf{I}_{K \times K} \otimes 
        p_i(j-1)
        \end{bmatrix} \cdot s_i(j), 
        & j \neq N, \\
        \begin{bmatrix}
        \textbf{I}_{d \times d} \otimes 
        p_i(N-1)
        \end{bmatrix} \cdot s_i(N), & j = N.
    \end{cases}
\end{align}

\subsection{Transformers Approximate Gradient Descent of \texorpdfstring{$N$}{}-Layer Neural Networks In-Context}
\label{sec:ICL_for_LNN_m}

For using neural networks to approximate \eqref{eqn:loss}, which contains smooth functions changeable, we need to approximate these smooth functions by simple combination of activation functions.  
Our key approximation theory is the sum of ReLUs for any smooth function \cite{bai2023transformers}.

\begin{definition}[Approximability by Sum of ReLUs, Definition 12 of \cite{bai2023transformers}]
\label{def:sum_of_relus_m}
Let $z\in \R^k$.
We say a function $g:\R^k \rightarrow \R$ is $(\epsilon_{\text{approx}}, R, H, C)$-approximable by sum of ReLUs if there exist a ``$(H,C)$-sum of ReLUs'' function $f_{H,C}(z)$ defined as
\begin{align*}
f_{H,C}(z) = \sum_{h=1}^{H} c_h \sigma(a_h^{\top} [z; 1]) ,
\end{align*} 
with $\sum_{h=1}^{H} |c_h| \leq C$, 
 $\max_{h \in [H]} \|a_h\|_1 \leq 1$, $ 
a_h \in \R^{k+1}$, and $ \quad c_h \in \R$, 
such that 
\begin{align*}
    \sup_{z \in [-R, R]^k} |g(z) - f_{H,C}(z)| \leq \epsilon_{\rm approx}.
\end{align*}
\end{definition}

\begin{figure}[t!]
\tikzstyle{block} = [rectangle, draw, minimum width=1.5cm, minimum height=1.0cm, text centered]
\tikzstyle{arrow} = [thick,->,>=stealth]
\tikzstyle{dashedarrow} = [thick,dashed,->,>=stealth]
\tikzstyle{label} = [font=\small]
\tikzstyle{labeltext} = [font=\small, midway, left]
    \centering
    \begin{tikzpicture}[node distance=1.0cm and 1.0cm]
    \node (input) [block] {Inputs};
    \node (tfan) [block, right=of input] {${\rm TF}_{\theta}^{N}$};
    \node (tf1a) [block, right=0.6cm of tfan] {${\rm TF}_{\theta}^{1}$};
    \node (tf1b) [block, right=0.6cm of tf1a] {${\rm TF}_{\theta}^{1}$};
    \node (ewmln) [block, right=0.6cm of tf1b] {${\rm EWML}_{\theta}^{N}$};
    \node (tf2) [block, right=1.5cm of ewmln] {${\rm TF}_{\theta}^{2}$};
    
    \node (xyw) [block, below=1.2cm of input] {$\{(x_i,y_i)\}_{i \in [n]}, w$};
    \node (pij) [block, below=1.2cm of tfan] {$p_i(j)$};
    \node (rij) [block, below=1.2cm of tf1a] {$r'_i(j)$};
    \node (gi) [block, below=1.2cm of tf1b] {$u(p_i(N), y_i)$};
    \node (si) [block, below=1.2cm of ewmln] {$s_i(j)$};
    \node (proj) [block, below=1.2cm of tf2] {${\rm Proj}_{\mathcal{W}}(w - \eta \nabla \mathcal{L}_n(w))$};
    
    \draw [arrow] (input) -- (tfan);
    \draw [arrow] (tfan) -- (tf1a);
    \draw [arrow] (tf1a) -- (tf1b);
    \draw [arrow] (tf1b) -- (ewmln);
    \draw [arrow] (ewmln) -- (tf2);
    \draw [arrow] (xyw) -- (input);
    
    \draw [dashedarrow] (tfan) -- node[labeltext] {\cref{lem:aprox_p_m}} (pij);
    \draw [dashedarrow] (tf1a) -- node[labeltext] {\cref{lem:aprox_r_p_m}} (rij);
    \draw [dashedarrow] (tf1b) -- node[labeltext] {\cref{lem:aprox_pl_m}} (gi);
    \draw [dashedarrow] (ewmln) -- node[labeltext] {\cref{lem:aprox_s_j_m}} (si);
    \draw [dashedarrow] (tf2) -- node[labeltext] {\cref{thm:icl_gd_m}} (proj);
    
    \end{tikzpicture}
    \caption{\textbf{One Step In-Context Gradient Descent (ICGD) with $(2N+4)$-layer Transformer.}
    This illustration presents the backpropagation process within an ICGD in a transformer model with $2N+4$ layers.
    It simulates a single gradient descent step for an $N$-layer neural network, trained with loss $\mathcal{L}_n$ and datasets $\{(x_i, y_i)\}_{i \in [n]}$.
    The term $p_i(j)$ denotes the output after the $j$-th layer for input $x_i$.
    The terms $r'_i(j)$, $u(p_i(N), y_i)$, and $s_i(j)$ are intermediate gradient terms of gradient $\nabla \mathcal{L}_n(w)$ from the chain rule.
    The expression ${\rm Proj}_{\mathcal{W}}(w - \eta \nabla \mathcal{L}_n(w))$ shows one gradient descent step.
    Here, $\eta$ is the learning rate, and $\mathcal{W}$ denotes the bounded domain for the $N$-layer NN parameters $w$.}
    \label{fig:theory_overview}
\end{figure}

\textbf{Overview of Our Proof Strategy.}
\cref{lem:decomposition_gd_m} and \cref{def:sum_of_relus_m} motivate the following strategy: term-by-term approximation for our gradient descent step \eqref{eqn:decomposition_gd_m}. 
Please see \cref{fig:theory_overview} for a high-level visualization.

\begin{enumerate}[leftmargin=3.5em]
    \item [\textbf{Step 1.}]
    Given $(x_i,w)$, we use $N$ attention layers to approximate the output of
    the first $j$ layers with input $x_i$, $p_i(j)\coloneqq {\rm pred}_h(x_i ; j) \in \R^k$ (\cref{def:N_nn_m}) for any $j \in [N]$. 
    Then we use $1$ attention layer to approximate chain-rule intermediate terms $r'_i(j-1)[k] := 
    r'(v_{{j}_k}^\top p_i(j-1))$ (\cref{def:abbv_m}) for any 
    $i \in [n]$, $j \in [N] $ and $k \in [K]$: \cref{lem:aprox_p_m} and \cref{lem:aprox_r_p_m}.
    \item [\textbf{Step 2.}]
    Given $(r'_i, p_i, w)$, we use an MLP layer to approximate 
    $u(p_i(N),y_i)$ (\cref{def:s_m}), for $i \in [n]$, and use $N$ element-wise multiplication layers to approximate $s_i(j)$ (\cref{def:s_m}), for any 
    $j \in[N]$: 
    \cref{lem:aprox_pl_m} and 
    \cref{lem:aprox_s_j_m}.
    Moreover, \cref{lem:error_g_s_m} shows the closeness result for
    approximating $s_i(j)$, which leads to the final error accumulation in \cref{thm:icl_gd_m}.
    \item [\textbf{Step 3.}]
    Given $(p_i, r'_i, g_i s_i(j), w)$, we use an attention layer to approximate $w - \eta \nabla \mathcal{L}_n(w)$.
    Then we use an MLP layer to approximate ${\rm Proj}_{\mathcal{W}}(w)$.
    And implementing $L$ steps gradient descent by a 
    $(2N+4)L$-layer neural network
    ${\rm NN}_{\theta}$ 
    constructed based on 
    \textbf{Step 1 and 2}.
    Finally, we arrive our main result: \cref{thm:icl_gd_m}. 
    Furthermore, \cref{lem:error_general_m} shows closeness results to the true gradient descent path. 
\end{enumerate}

\textbf{Step 1.}
We start with approximation for $p_i(j)$.
\begin{lemma}[Approximate $p_i(j)$]
    \label{lem:aprox_p_m}
    Let upper bounds $B_v, B_x > 0$ such that
    for any $k \in [K], j \in [N] ~\text{and}~ i \in [n]$, $\|v_{j_k}\|_2 \leq B_v$,
    and $\|x_i\|_2 \leq B_x$.
    For any $j \in [N], i \in [n]$, define
    \begin{align*}
        B_r^j :=  \max_{\abs{t} \leq B_v B_r^{j-1}} \abs{r(t)},
        \, B_r^0 := B_x ,\,\text{and}\,
        B_r := \max_{j} B_r^j.
    \end{align*}
    Let function $r(t)$ be $(\epsilon_r, R_1, M_1, C_1)$-approximable for $R_1 = \max \{B_v B_r, 1\}$, $M_1 \leq \Tilde{\mathcal{O}}(C_1^2 \epsilon_r^{-2})$, where $C_1$ depends only on $R_1$ and the $C^2$-smoothness of $r$.
    Then, for any $\epsilon_r>0$, there exist $N$ attention layers ${\rm Attn}_{\theta_1}, \ldots, {\rm Attn}_{\theta_N}$ such that for any input $h_i \in \R^D$ takes from \eqref{eqn:input}, they map
    \begin{align*}
        h_i = [x_i; y_i; w; \bar{p}_i(1); \ldots; \bar{p}_i(j-1); \mathbf{0}; 1; t_i] \xrightarrow
        {{\rm Attn}_{\theta_j}} 
        \Tilde{h_i} = [x_i; y_i; w; \bar{p}_i(1); \ldots; \bar{p}_i(j); \mathbf{0}; 1; t_i],
    \end{align*}
    where $\bar{p}_i(j)$ is  approximation for $p_i(j)$ (\cref{def:N_nn_m}).
    In the expressions of $h_i$ and $\Tilde{h}_i$, the dimension of $\mathbf{0}$ differs.
    Specifically, the $\mathbf{0}$ in $h_i$ is larger than in $\Tilde{h}_i$.
    The dimensional difference between these $\mathbf{0}$ vectors equals the dimension of $\bar{p}_i(j)$.
    Suppose function $r$ is $L_r$-smooth in bounded domain $\mathcal{W}$, then for any $i \in [n+1]$, $j \in [N]$, $\bar{p}_i(j)$ such that
    \begin{align}
    \label{eqn:error_1_m}
        \bar{p}_i(j) = & ~ p_i(j) + \epsilon(i,j), ~ \|\epsilon(i,j)\|_2 \leq 
        \begin{cases}
            (\sum_{l=0}^{j-1} 
            K^{l/2} L_r^l B_v^l) \sqrt{K} \epsilon_r ~,  \: 1 \leq j \leq N-1 \\ 
            (\sum_{l=0}^{N-1} 
            K^{l/2} L_r^l B_v^l) \sqrt{d} \epsilon_r 
            ~, \: j=N.
        \end{cases}
    \end{align} 
    Additionally, for any $j \in [N]$, the norm of parameters $B_{\theta_j}$ defined as \eqref{eqn:tf_norm}  such that $B_{\theta_j} \leq 
        1 + K C_1$.
\end{lemma}
\begin{proof}
    Please see \cref{proof:lem:aprox_p_m} for a detailed proof.
\end{proof}

Notice that the form of error accumulation in \cref{lem:aprox_p_m} is complicated. 
For the ease of later presentations, we define the upper bound of coefficient in \eqref{eqn:error_1_m} as  
\begin{align}
\label{eqn:E_r_m}
    E_r := \max_{j \in [N]}
    \frac{\|\epsilon(i,j)\|_2}{\epsilon_r} 
    = \max_{j \in [N]}
    \{(\sum_{l=0}^{j-1} 
    K^{l/2} L_r^l B_v^l) \sqrt{K}, (\sum_{l=0}^{N-1} 
    K^{l/2} L_r^l B_v^l) \sqrt{d} \},
\end{align}
such that  \eqref{eqn:error_1_m} becomes
\begin{align}        \label{eqn:e_r_m}
        \bar{p}_i(j) = p_i(j) + \epsilon(i,j), \quad  \|\epsilon(i,j)\|_2 \leq 
        E_r \epsilon_r.
\end{align}
Moreover, we abbreviate $\bar{p}_i := [\bar{p}_i(1); \ldots; \bar{p}_i(N)] \in \R^{(N-1)K+d}$,
such that the output of ${\rm Attn}_{\theta_1} \circ \cdots \circ {\rm Attn}_{\theta_N}$ is 
    \begin{align}
        h_i = [x_i; y_i; w; \bar{p}_i; \mathbf{0}; 1; t_i].
        \label{eqn:h_i_p_m}
    \end{align}
Then, the next lemma approximates $r'_i(j)$ base on $\bar{p}_i(j)$ obtained in \cref{lem:aprox_p_m}.

\begin{lemma}[Approximate $r'_i(j)$]
    \label{lem:aprox_r_p_m}
    Let upper bounds $B_v, B_x > 0$ such that
    for any $k \in [K], j \in [N] ~\text{and}~ i \in [n]$, $\|v_{j_k}\|_2 \leq B_v$,
    and $\|x_i\|_2 \leq B_x$.
    For any $j \in [N], i \in [n]$, define
    \begin{align*}
        B_r'^j :=  \max_{\abs{t} \leq B_v B_{r'}^{j-1}} \abs{r'(t)},
        \, B_{r'}^0 := B_x,\,\text{and}\,
        B_{r'} :=  \max_{j} B_{r'}^j.
    \end{align*}
    Suppose function $r'(t)$ is $(\epsilon_{r'}, R_2, M_2, C_2)$-approximable for $R_2 = \max \{B_v B_{r'}, 1\}$, $M_2 \leq \Tilde{\mathcal{O}}(C_2^2 \epsilon_r'^{-2})$, where $C_2$ depends only on $R_2$ and the $C^2$-smoothness of $r'$.
    Then, for any $\epsilon_r > 0$, there exist an attention layer 
    $ {\rm Attn}_{\theta_{N+1}}$ such that for any input $h_i \in \R^D$ takes from \eqref{eqn:h_i_p_m}, it maps
    \begin{align*}
        h_i = [x_i; y_i; w; \bar{p}_i; \mathbf{0}; 1; t_i]  \xrightarrow
        {{\rm Attn}_{\theta_{N+1}}} 
        \Tilde{h_i} = [x_i; y_i; w; \bar{p}_i; \bar{r}'_i; \mathbf{0}; 1; t_i],
    \end{align*}
    where $\bar{r}'_i(j)$ is  approximation for $r'_i(j)$ (\cref{def:abbv_m}) and $\bar{r}'_i := [\bar{r}'_i(0); \ldots; \bar{r}'_i(N-1)] \in \R^{(N-2)K+d}$.
    Similar to \cref{lem:aprox_p_m}, in the expressions of $h_i$ and $\Tilde{h}_i$, the dimension of $\mathbf{0}$ differs.
    In addition, let $E_r$ be defined in \eqref{eqn:e_r_m}, for any $i \in [n+1]$, $j \in [N], k \in [K]$, $\bar{r}'_i(j)$ such that 
    \begin{align}
        \bar{r}'_i(j-1)[k] = 
        r'_i(j-1)[k] + \epsilon(i,j,k),  ~ \abs{\epsilon(i,j,k)} \leq 
        \epsilon_{r'} +
        L_{r'} B_v E_r \epsilon_r,
        \label{eqn:error_2_m}
    \end{align}
    where $\epsilon_r$ denotes the error generated in approximating
    $r$ by sum of ReLUs $\bar{r}$ follows \eqref{eqn:r_bar_m}.
    Additionally, the norm of parameters $B_{\theta_{N+1}}$ defined as \eqref{eqn:tf_norm} such that
    $B_{\theta_{N+1}} \leq
    1 + K (N-1) C_2$.
\end{lemma}
\begin{proof}[Proof Sketch]
    By \cref{lem:aprox_p_m}, we obtain $\bar{p}_i(j)$, the approximation for $p_i(j)$ \eqref{eqn:p_m}.
    Using $\bar{p}_i(j)$, we construct an Attention layer to approximate $r'_i(j)$.
    We then establish upper bounds for the errors $\abs{\bar{r}'_i(j)[k] - 
    r'_i(j)[k]}$ by applying Cauchy-Schwarz inequality and \cref{lem:aprox_p_m}.
    Finally we present the norms \eqref{eqn:tf_norm} of the Transformers constructed.
    Please see \cref{proof:lem:aprox_r_p_m} for a detailed proof.
\end{proof}

Let ${\rm Attn}_{\theta_j}
(j \in [N])$ be as defined in \cref{lem:aprox_p_m}, then \cref{lem:aprox_r_p_m} implies that for the input takes from \cref{prob:icl_n_m}, the output of ${\rm Attn}_{\theta_1} \circ \cdots \circ {\rm Attn}_{\theta_{N+1}}$ is 
\begin{align}
    h_i = [x_i; y_i; w; \bar{p}_i; \bar{r}'_i; \mathbf{0}; 1; t_i].
    \label{eqn:h_i_r_m}
\end{align}

\textbf{Step 2.} Now, we construct an approximation for $u(p_i(N),y_i) = (\pdv {\ell(p_i(N), y_i)}{p_i(N)})^\top$.

\begin{lemma}[Approximate 
$u(p_i(N),y_i)$]
    \label{lem:aprox_pl_m}
    Let upper bounds $B_v, B_x, >0$ such that
    for any $k \in [K], j \in [N] ~\text{and}~ i \in [n]$, $\|v_{j_k}\|_2 \leq B_v$,
    and $\|x_i\|_2 \leq B_x$.
    For any $k \in [d]$, suppose function $u(t,y)[k]$ be $(\epsilon_{l}, R_3, M_3^k, C_3^k)$-approximable for $R_3 = \max
    \{B_v B_r, B_y, 1\}$, $M_3 \leq \Tilde{\mathcal{O}}((C_3^k)^2 \epsilon_l^{-2})$, where $C_3^k$ depends only on $R_3^k$ and the $C^3$-smoothness of $u(t,y)[k]$.
    Then, there exists an MLP layer 
    $ \rm{MLP}_{\theta_{N+2}}$ such that for any input sequences $h_i \in \R^D$ takes from \eqref{eqn:h_i_r_m}, 
    it maps
    \begin{align*}
        & ~ h_i = [x_i; y_i; w; \bar{p}_i; \bar{r}'_i; \mathbf{0}; 1; t_i] \xrightarrow
        {{\rm{MLP}}_{\theta_{N+2}}} 
        \Tilde{h_i} = [x_i; y_i; w; \bar{p}_i; \bar{r}'_i; g_i; \mathbf{0}; 1; t_i],
    \end{align*}
    where $g_i \in \R^d$ is an approximation for $u(p_i(N),y_i)$. 
    For any $k \in [d]$, assume $u(p_i(N),y_i)$ is $L_l$- Lipschitz continuous.  
    Then the approximation $g_i$ such that,
    \begin{align}
        \label{eqn:error_3_m}
        g_i[k] = u(p_i(N),y_i)[k] + \epsilon(i,k), 
    \end{align}
    with $\abs{\epsilon(i,k)} \leq 
        \epsilon_l + L_l E_r \epsilon_r$.
     Additionally, the parameters $\theta_{N+2}$ such that
     \begin{align*}
         B_{\theta_{N+2}} \leq \max \{R_3 +1, C_3\}.
     \end{align*}
\end{lemma}
\begin{proof}[Proof Sketch]
    By \cref{def:N_nn_m}, we provide term-by-term approximations for $p_i(j)$ as forward propagation. 
    Specifically, we construct Attention layers to implement forward propagation algorithm.
    Then we establish upper bounds for the errors $\|\bar{p}_i(j) - p_i(j)\|_2$ inductively. Finally, we present the norms \eqref{eqn:tf_norm} of the Transformers constructed.
    Please see \cref{proof:lem:aprox_pl_m} for a detailed proof.
\end{proof}

Let ${\rm Attn}_{\theta_j}
(j \in [N+1])$ be as defined in \cref{lem:aprox_p_m} and \cref{lem:aprox_r_p_m}, then for any input sequences $h_i \in \R^D$ takes from \eqref{eqn:input}, the output of ${\rm Attn}_{\theta_1} \circ \cdots \circ {\rm Attn}_{\theta_{N+1}} \circ \rm{MLP}_{\theta_{N+2}}$ is 
\begin{align}
    h_i = [x_i; y_i; w; \bar{p}_i; \bar{r}'_i; g_i; \mathbf{0}; 1; t_i].
    \label{eqn:h_i_g_m}
\end{align}
Before introducing our next approximation lemma, we define an element-wise multiplication layer, since attention mechanisms and MLPs are unable to compute self-products (e.g., output $xy$ from input $[x; y]$).
To enable self-multiplication, we introduce a function $\gamma$. 
This function, for any square matrix, preserves the diagonal elements and sets all others to zero.

\begin{definition}[Operator Function $\gamma$]
    \label{def:gamma_m}
    For any square matrix $A \in 
    \R^{n \times n}$, define 
    \begin{align*}
        \gamma(A) := {\rm diag}(A[1,1], \ldots A[n,n]) \in \R^{n \times n}.
    \end{align*}
\end{definition}

By \cref{def:gamma_m}, we introduce the following element-wise multiplication layer, capable of performing self-multiplication operations such as the Hadamard product.

\begin{definition}[Element-wise Multiplication Layer]
\label{def:ewml_m}
    Let $\gamma$ be defined as \cref{def:gamma_m}.
    An element-wise multiplication layer with $m$ heads is denoted as ${\rm Attn}_{\theta}(\cdot)$ with parameters $\theta=\{Q_m, K_m, V_m\}_{m \in [M]}$. 
    On any input sequence $H \in \R^{D \times n}$, 
\begin{align}
\label{eqn:ewml_m}
{\rm EWML}_{\theta}(H) = H +  \sum_{i=1}^m  (V_m H) \cdot \gamma((Q_m H)^{\top} 
(K_m H)).
\end{align}
where $Q_m,K_m,V_m\in\R^{D\times D}$ and $\gamma(\cdot)$ is operator function follows \cref{def:gamma_m}.
In vector form, for for each token $h_i\in\R^D$ in $H$, it outputs
$[{\rm EWML}_{\theta}(H)]_i = h_i +  \sum_{m=1}^M  \gamma (\langle Q_m h_i, K_m h_i \rangle) \cdot V_m h_i$.
In addition, we define $L$-layer neural networks 
\begin{align*}
    {\rm EWML}_{\theta}^{L} := {\rm EWML}_{\theta_1} \circ \cdots \circ
{\rm EWML}_{\theta_L}.
\end{align*}
\end{definition}

\begin{remark}[Necessary for Element-Wise Multiplication Layer]
As we shall show in subsequent
sections, ELML is capable of implementing multiplication in $h_i$. 
Specifically, it allows us to multiply some elements in $h_i$ in \cref{lem:aprox_s_j_m}.
By \cref{def:attn}, it is impossible for transformer layers to achieve our goal without any other assumptions.
\end{remark}
Similar to \eqref{eqn:tf_norm}, we define the norm for $L$-layer transformer 
${\rm EWML}_{\theta}^{L}$ as: 
\begin{align}
\label{eqn:ewml_norm_m}
B_{\theta} :=  
\max_{m \in [M], l \in [L]} 
\{ \|Q_{m}^l\|_1, 
\| K_{m}^l\|_1, 
\| V_{m}^l\|_1 \}.
\end{align}
Then, given the approximations for $p_i(j)$ and $r'_i(j)$, we use $N$ element-wise multiplication layer (\cref{def:ewml_m}) to approximate $s_i(j)$, the chain-rule intermediate terms defined as \cref{def:s_m}.

\begin{lemma}[Approximate $s_i(j)$]
    \label{lem:aprox_s_j_m}
    Recall that $s_i(j) = r'_i(j-1) \odot (V_{j+1}^{\top} \cdot s_i(j+1))$ follows \cref{def:s_m}. 
    Let the initial input take from \eqref{eqn:h_i_g_m}.
    Then, there exist $N$ element-wise multiplication layers: ${\rm EWML}_{\theta_{N+3}}, \ldots, {\rm EWML}_{\theta_{2N+2}}$ such that for input sequences, 
    $j \in [N]$,
    \begin{align*}
        h_i = [x_i; y_i; w; \bar{p}_i; \bar{r}'_i; g_i; \bar{s}_i(N); \ldots; \bar{s}_i(j+1); \mathbf{0}; 1; t_i],
    \end{align*}
    they map ${\rm EWML}_{\theta_{2N+3-j}}(h_i)
        = [x_i; y_i; w; \bar{p}_i; \bar{r}'_i; g_i; \bar{s}_i(N); \ldots; \bar{s}_i(j); \mathbf{0}; 1; t_i]$,
    where the approximation $\bar{s}_i(j)$ is defined as recursive form: for any $i \in [n+1], j \in[N]$,
    \begin{align}
        \bar{s}_i(j) 
        := \begin{cases}
            \bar{r}'_i(j-1) \odot (V_{j+1}^{\top} \cdot \bar{s}_i(j+1)), & j\in[N-1] \\
            \bar{r}'_i(N-1) \odot g_i, & j=N.
        \end{cases}
        \label{eqn:s_bar_i(j)_m}
    \end{align}
    Additionally, for any $j \in [N]$,  $B_{\theta_{N+2+j}}$ defined in \eqref{eqn:tf_norm} satisfies $B_{\theta_{N+2+j}} \leq 1$.
\end{lemma}
\begin{proof}
    Please see \cref{proof:lem:aprox_s_j_m} for a detailed proof.
\end{proof}

Let ${\rm Attn}_{\theta_j}
(j \in [N+1]), {\rm MLP}_{\theta_{N+2}} $ be as defined in \cref{lem:aprox_p_m}, \cref{lem:aprox_r_p_m} and 
\cref{lem:aprox_pl_m} respectively.
Define $\bar{s}_i := 
[\bar{s}_i(N); \ldots; \bar{s}_i(1)] \in \R^{(N-1)K+d}$, then for any input sequences $h_i \in \R^D$ takes from \cref{prob:icl_n_m}, the output of neural network
\begin{align}
    {\rm Attn}_{\theta_1} \circ \cdots \circ
    {\rm Attn}_{\theta_{N+1}}
    \circ
    {\rm MLP}_{\theta_{N+2}}
    \circ
    {\rm EWML}_{\theta_{N+3}}
    \circ \cdots \circ {\rm EWML}_{\theta_{2N+2}},  
    \label{eqn:nn_1_m}
\end{align}
is 
\begin{align}
    h_i = [x_i; y_i; w; \bar{p}_i; \bar{r}'_i; 
    \bar{s}_i; \mathbf{0}; 1; t_i].
    \label{eqn:h_i_s_m}
\end{align}
For the sake of simplicity, we consider  ReLU Attention layer and MLP layer are both a special kind of transformer.
In this way, by \cref{def:tf}, \eqref{eqn:nn_1_m} becomes
\begin{align*}
 {\rm TF}_{\theta}^{N+2} \circ {\rm EWML}_{\theta}^{N}.    
\end{align*}
Next we calculate the error accumulation $\abs{\bar{s}_i(j)[k] - s_i(j)[k]}$ based on \cref{lem:aprox_r_p_m} and \cref{lem:aprox_pl_m}.

\begin{lemma}[Error for $\bar{s}_i(j)$]
\label{lem:error_g_s_m}
    Suppose the upper bounds $B_v, B_x> 0$ such that
    for any $k \in [K], j \in [N] ~\text{and}~ i \in [n]$, $\|v_{j_k}\|_2 \leq B_v$,
    and $\|x_i\|_2 \leq B_x$.
    Let $r'_i(j) \in \R^K$ such that $r'_i(j)[k] := 
    r'(v_{{j+1}_k}^\top p_i(j))$ follows \cref{def:abbv_m}.
    Let $s_i(j) = R_i(j-1) V_{j+1}^{\top} \ldots R_i(N-2) V_{N}^{\top} \cdot R_i(N-1) u$ follows \cref{def:s_m}.
    Let $\bar{r}'_i(j), g_i, \bar{s}_i(j)$ be the approximations for $r'_i(j), u(p_i(N),y_i), s_i(j)$  follows
    \cref{lem:aprox_r_p_m},
    \cref{lem:aprox_pl_m} and \cref{lem:aprox_s_j_m} respectively.
    Let $B_{r'}$ be the upper bound of $\bar{r}'_i(j)[k]$ and $r'_i(j)[k]$ as defined in \cref{lem:aprox_r_p_m}.
    Let $B_l$ be the upper bound of $g_i$ and $u(p_i(N),y_i)$ as defined in \cref{lem:aprox_pl_m}.
    Then for any $i \in [n+1], j \in [N], k \in [K]$, 
    \begin{align*}
       \bar{s}_i(j)[k] 
        \leq  & ~   B_s, \\
        \abs{\bar{s}_i(j)[k] -  s_i(j)[k]}  
        \leq  & ~  E_s^{r} \epsilon_r + E_s^{r'} \epsilon_{r'} + E_s^{l} \epsilon_l,
    \end{align*}
    where
    $B_s$ is the upper bound of $\bar{s}_i(j)[k]$ and $E_s^r, E_s^{r'}, E_s^l$ are the
    coefficients of $\epsilon_r, \epsilon_{r}', \epsilon_l$ in the upper bounds of $\abs{\bar{s}_i(j)[k] - s_i(j)[k]}$, respectively.
\end{lemma}

\begin{proof}
    Please see \cref{proof:lem:error_g_s_m} for a detailed proof.
\end{proof}

\cref{lem:error_g_s_m} offers the explicit form of the error $\abs{\bar{s}_i(j)[k] - s_i(j)[k]}$, which is crucial for calculating the error $\|\nabla_w \bar{\mathcal{L}}_n(w) - \nabla_w \mathcal{L}_n(w)\|_2$ in \cref{thm:icl_gd_m}.

\textbf{Step 3.} Combining the above, we prove the existence of a neural network, that implements $L$ in-context GD steps on our $N$-layer neural network. 
And finally we arrive our main result: a neural network $\mathcal{T}$ for \cref{prob:icl_n_m}.

\begin{theorem}[In-Context Gradient Descent on $N$-layer NNs]
\label{thm:icl_gd_m}
    Fix any $B_v, \eta, \epsilon > 0, L \geq 1$.
    For any input sequences takes from $\eqref{eqn:input}$, their exist upper bounds $B_x,B_y$ such that for any $i \in [n]$, $\|y_i\|_2 \leq B_y$, $\|x_i\|_2 \leq B_x$.
    Assume functions $r(t)$, $r'(t)$ and $u(t,y)[k]$ are $L_r,L_{r'},L_l$-Lipschitz continuous.
    Suppose $\mathcal{W}$ is a closed domain such that for any $j \in [N-1]$ and $k \in [K]$,
    \begin{align*}
        \mathcal{W} \subset \left\{
        w = [v_{j_k}] \in \mathbb{R}^{D_N} : \|v_{j_k}\|_2 \leq B_v
        \right\},      
    \end{align*}
    and ${\rm Proj}_{\mathcal{W}}$ project $w$ into bounded domain $\mathcal{W}$. 
    Assume
    ${\rm Proj}_{\mathcal{W}} = {\rm MLP}_{\theta}$ for some MLP layer with hidden dimension 
    $D_w$ parameters $\|\theta\| \leq C_{w}$.
    If functions $r(t)$, $r'(t)$ and $u(t,y)[k]$ are $C^4$-smoothness, then for any $\epsilon > 0$, there exists a transformer model ${\rm NN}_{\theta}$ with $(2N+4)L$ hidden layers
    consists of $L$ neural network blocks 
    ${\rm TF}_{\theta}^{N+2} \circ
    {\rm EWML}_{\theta}^{N} \circ
    {\rm TF}_{\theta}^{2}$,
    \begin{align*}
        {\rm NN}_{\theta} 
        := & ~ {\rm TF}_{\theta}^{N+2} \circ
        {\rm EWML}_{\theta}^{N} \circ
        {\rm TF}_{\theta}^{2},
    \end{align*}
    such that the heads number $M^l$, parameter dimensions $D^l$, and the parameter norms
    $B_{\theta^l}$ suffice
    \begin{align*}
    \max_{l \in [(2N+4)L]} M^l &\leq \Tilde{O}(\epsilon^{-2}), \\
    \max_{l \in [(2N+4)L]} D^l &\leq O(NK^2) + D_w, \\
    \max_{l \in [(2N+4)L]} B_{\theta^l} &\leq O(\eta) + C_w + 1,
    \end{align*}
    where $\Tilde{O}(\cdot)$ hides the constants that depend on $d,K,N$, the radius parameters $B_x,B_y,B_v$ and the smoothness
    of $r$ and $\ell$. 
    And this neural network such that for any input sequences $H^{(0)}$, take from \eqref{eqn:input}, ${\rm NN_{\theta}}(H^{(0)})$ implements $L$ steps in-context gradient descent on risk Eqn~\eqref{eqn:loss}: For every $l \in [L]$, the $(2N+4)l$-th layer outputs $h_i^{((2N+4)l)} = [x_i; y_i; \bar{w}^{(l)}; \mathbf{0}; 1; t_i]$ for every $i \in [n+1]$, and approximation gradients
    $\bar{w}^{(l)}$ such that 
    \begin{align*}
        \bar{w}^{(l)} = 
        {\rm Proj}_{\mathcal{W}} 
        (\bar{w}^{(l-1)} - \eta \nabla \mathcal{L}_n(\bar{w}^{(l-1)}) + \epsilon^{(l-1)}),
    \end{align*}
    where $\bar{w}^{(0)} = \mathbf{0}$, and $\|\epsilon^{(l-1)}\|_2 \leq \eta \epsilon$ is an error term.
\end{theorem}
\begin{proof}[Proof Sketch]
     Let the first $2N+2$ layers of ${\rm NN}_{\theta}$ are Transformers and EWMLs constructed in \cref{lem:aprox_p_m}, 
    \cref{lem:aprox_r_p_m},
    \cref{lem:aprox_pl_m}, and
    \cref{lem:aprox_s_j_m}.
    Explicitly, we design the last two layers to implement the gradient descent step (\cref{lem:decomposition_gd_m}).
    We then establish the upper bounds for error $\|\nabla_w \bar{\mathcal{L}}_n(w) - \nabla_w \mathcal{L}_n(w)\|_2$, where $\nabla_w \bar{\mathcal{L}}_n(w)$, derived from the outputs of ${\rm NN}_{\theta}$, approximates
    $\nabla_w \mathcal{L}_n(w)$.
    Next, for any $\epsilon>0$, we select appropriate parameters $\epsilon_l$, $\epsilon_r$ and $\epsilon_{r'}$ to ensure that  $\|\nabla_w \bar{\mathcal{L}}_n(\bar{w}^{(l-1)}) - \nabla_w \mathcal{L}_n(\bar{w}^{(l-1)})\|_2 
    \leq \epsilon$ holds for any $l \in [L]$.
    
    Please see \cref{proof:thm:icl_gd_m} for a detailed proof.
\end{proof}

We summarize and visualize the backpropagation process within an ICGD in a transformer model with $2N+4$ layers in \cref{fig:theory_overview}.
As a direct result, the neural networks ${\rm NN}_{\theta}$ constructed earlier is able to approximate the true gradient descent trajectory 
$\{w^l_{\rm GD}\}_{l\geq0}$, defined by $w^0_{\rm GD} = \mathbf{0}$ and 
$w^{l+1}_{\rm GD} = w^l_{\rm GD} - \eta \nabla_w \mathcal{L}_n(w^l_{\rm GD})$ for any 
$l \geq 0$.  
Consequently, \cref{thm:icl_gd_m} motivates us to investigate the error accumulation under setting 
\begin{align*}
    \bar{w}^{(l)} = 
    {\rm Proj}_{\mathcal{W}} 
    (\bar{w}^{(l-1)} - \eta \nabla \mathcal{L}_n(\bar{w}^{(l-1)}) + \epsilon^{(l-1)}),
\end{align*}
where $\bar{w}^{(0)} = \mathbf{0}$, and $\|\epsilon^{(l-1)}\|_2 \leq \eta \epsilon$ represents error terms.
Moreover, \cref{coro:error_ICGD_m} shows ${\rm NN}_{\theta}$ constructed in \cref{thm:icl_gd_m} implements $L$ steps ICGD with exponential error accumulation to the true GD paths.

\begin{corollary}[Error for implementing ICGD on $N$-layer neural network]
\label{coro:error_ICGD_m}
    Fix $L \geq 1$, under the same setting as \cref{thm:icl_gd_m}, $(2N+4)L$-layer neural networks
    ${\rm NN}_{\theta}$ approximates
    the true gradient descent trajectory 
    $\{w^l_{\rm GD}\}_{l\geq0} \in \R^{D_N}$ with the error accumulation
    $\|\bar{w}^{l}-w^{l}_{\rm GD}\|_2
    \leq L_f^{-1} (1 + n L_f)^l \epsilon$,
    where $L_f$ denotes the Lipschitz constant of $\mathcal{L}_n(w)$ within $\mathcal{W}$.
\end{corollary}
\begin{proof}
    Please see \cref{proof:coro:error_ICGD_m} for a detailed proof.
\end{proof}

\section{In-Context Deep Learning with Softmax Transformers}
\label{sec:softmax}
In this section, we extend our analysis 
from ReLU-transformers to more practical $\Softmax$-transformers for ICGD of $N$-layer neural network (\cref{app:sec_soft}).
Specifically, we establish the existence of $\Softmax$-transformers capable of performing ICGD for $N$-layer neural networks in \cref{thm:icgd_soft_main} and give more details in \cref{app:sec_soft}.
\begin{theorem}[In-Context Gradient Descent of $\Softmax$-Transformer]
\label{thm:icgd_soft_main}
    Fix any $B_w, \eta, \epsilon > 0, L \geq 1$.
    For any input sequences takes from $\eqref{eqn:input}$, their exist upper bounds $B_x,B_y$ such that for any $i \in [n]$, $\|y_i\|_{\max} \leq B_y$, $\|x_i\|_{\max} \leq B_x$.
    Suppose $\mathcal{W}$ is a closed domain such that
    $\|w\|_{\max} \leq B_w$
    and ${\rm Proj}_{\mathcal{W}}$ project $w$ into bounded domain $\mathcal{W}$. 
    Assume
    ${\rm Proj}_{\mathcal{W}} = {\rm MLP}_{\theta}$ for some MLP layer.
    Define $l(w, x_i,y_i)$ as a loss function with $L$-Lipschitz gradient.
    Let $\mathcal{L}_n(w) = \frac{1}{n}\sum_{i=1}^n 
    \ell(w, x_i,y_i)$ denote the empirical loss function, then there exists a $\Softmax$-transformer ${\rm NN}_{\theta}$,  such that for any input sequences $H^{(0)}$, take from \eqref{eqn:input}, ${\rm NN_{\theta}}(H^{(0)})$  implements $L$ steps in-context gradient descent on $\mathcal{L}_n(w)$: 
    For every $l \in [L]$, the $4l$-th layer outputs $h_i^{(4l)} = [x_i; y_i; \bar{w}^{(l)}; \mathbf{0}; 1; t_i]$ for every $i \in [n+1]$, and approximation gradients
    $\bar{w}^{(l)}$ with $\bar{w}^{(0)} = \mathbf{0}$ such that 
    \begin{align*}
        \bar{w}^{(l)} = 
        {\rm Proj}_{\mathcal{W}} 
        (\bar{w}^{(l-1)} - \eta \nabla \mathcal{L}_n(\bar{w}^{(l-1)}) + \epsilon^{(l-1)}), 
    \end{align*}
    where $\|\epsilon^{(l-1)}\|_2 \leq \eta \epsilon$ is an error term.
\end{theorem}

\begin{proof}[Proof Sketch]
    By our assumption ${\rm Proj}_{\mathcal{W}} = {\rm MLP}_{\theta}$,  we only need to find a transformer to implement gradient descent $w^+ := w - \eta \nabla \mathcal{L}_n(w)$.
    For ant input takes from \eqref{eqn:input_l}, let  function $f: \R^{D \times n} \rightarrow \R^{D \times n}$ maps $w$ into $w - \eta \nabla \mathcal{L}_n(w)$ and preserve other elements. 
    By \cref{lem:approx_k_v}, their exist a transformer block $f_{\Softmax}$ capable of approximating $f$ with any desired small error. 
    Therefore, $f_{\Softmax} \circ {\rm MLP}$ suffices our requirements.
    
    Please see \cref{app:proof_thm_icgd_soft} for a detailed proof.
\end{proof}

\section{Numerical Studies}
\label{sec:experiments}
In this section, we conduct experiments to verify the capability of ICL to learn feed-forward neural networks, and give details in \cref{app:sec_exp}.
We conduct the experiments based on 3-, 4- and 6-layer NN using both ReLU- and $\Softmax$-Transformer.
The main objective is to validate the performance of ICL matches that of training $N$-layer networks, i.e., the results in \cref{thm:icl_gd_m}, \cref{thm:icgd_soft_main}, and \cref{thm:icl_gd_m_io}.
However, a minor limitation is that the trained transformers do not always achieve the theoretical construction.

\begin{figure*}[ht]
    \centering
    \begin{subfigure}[b]{0.465\textwidth}
        \includegraphics[width=\textwidth]{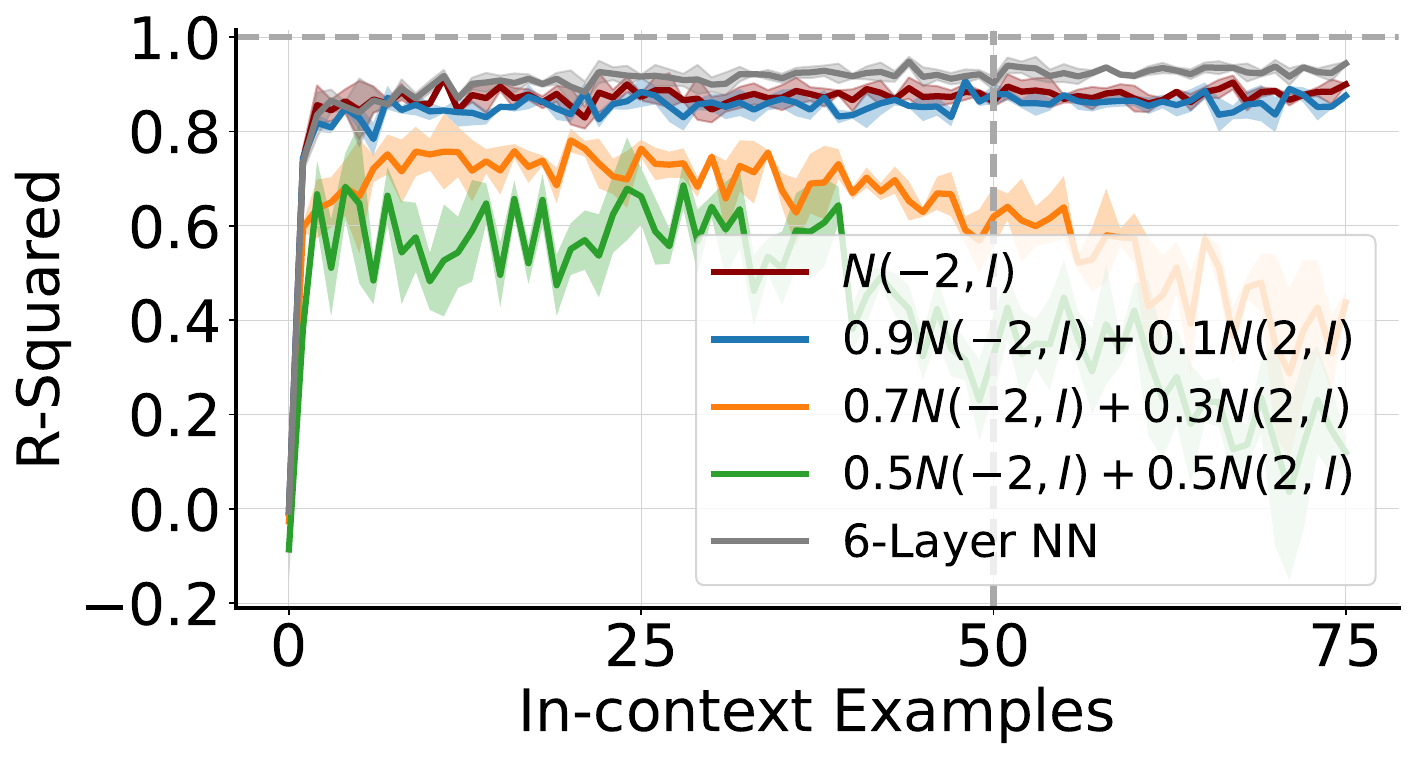}
        \caption{ReLU-Transformer}
        \label{fig:6_l_relu_main}
    \end{subfigure}
    \hfill
    \begin{subfigure}[b]{0.45\textwidth}
        \includegraphics[width=\textwidth]{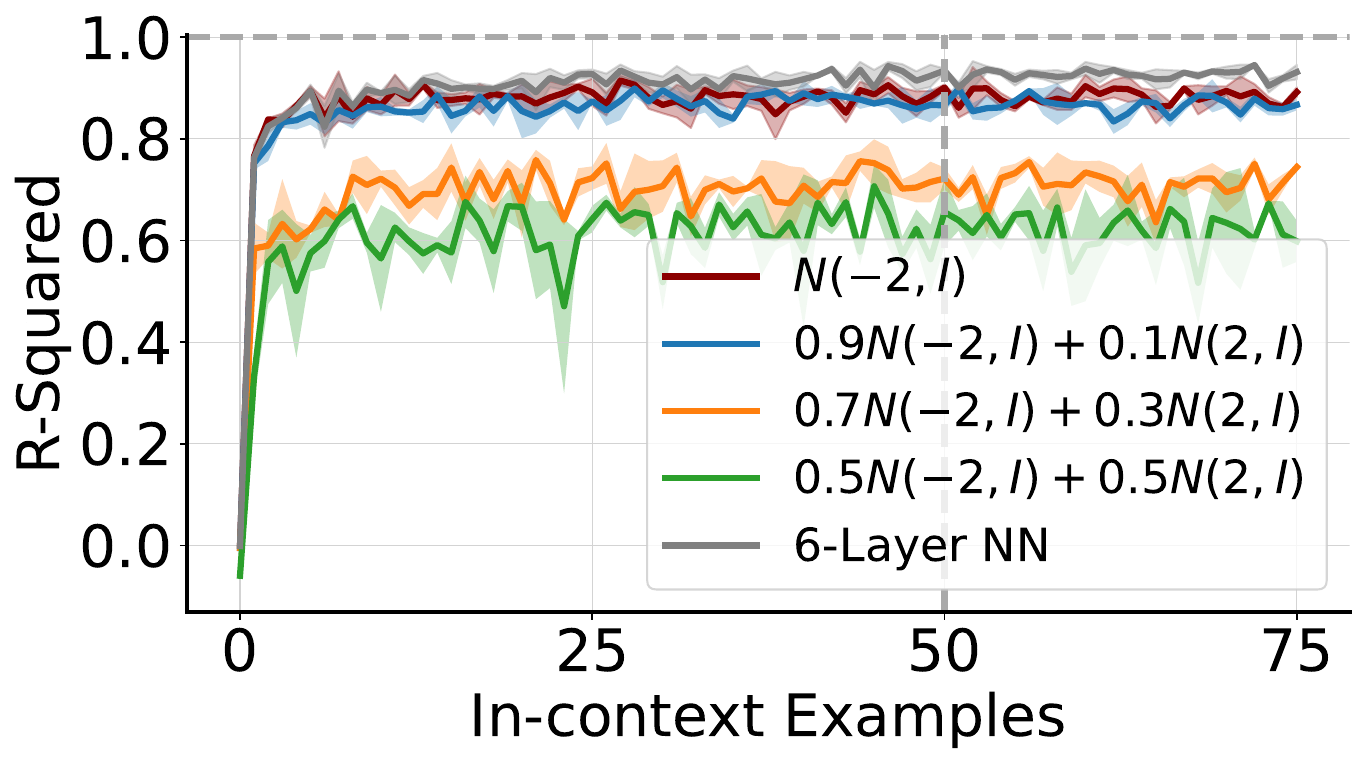}
        \caption{Softmax-Transformer}
        \label{fig:6_l_soft_main}
    \end{subfigure}
    \caption{\textbf{Performance of ICL in ReLU-Transformer and Softmax-Transformer:} 
    ICL learns 6-layer NN and achieves R-squared values comparable to those from training with prompt samples.
    }
    \label{fig:relu_soft_main}
\end{figure*}

Specifically, we sample the input of feed-forward network $x \in \RR^{d}$ from the Gaussian mixture distribution: $w_1 N(-2, I_{d}) + w_2 N(2, I_{d})$, where $w_1, w_2 \in \RR$, and d=20.
We consider the network $f : \RR^d \rightarrow \RR$ as a 3-, 4-, or 6-layer NN.
We generate the true output by $y=f(x)$.
For the pertaining data, we use 50 in-context examples, and sample them from $N(-2, I_{d})$.
For the testing data, we use 75 in-context examples, and sample them from four distributions: (i) $\omega_1 = 1,  \omega_2 = 0$, (ii) $\omega_1 = 0.9,  \omega_2 = 0.1$, (iii) $\omega_1 = 0.7,  \omega_2 = 0.3$, (iv) $\omega_1 = 0.5,  \omega_2 = 0.5$.
We show the results of 6-layer NN in \cref{fig:relu_soft_main}.

\section{Conclusion}
\label{sec:conclusion}
\input{4conclusion}

%% file: 4conclusion.tex
We provide an explicit characterization of the ICL capabilities of both ReLU- and $\Softmax$-transformer in approximating the gradient descent training process of a $N$-layer feed-forward neural network.
Our results include approximation  (\cref{thm:icl_gd_m} and \cref{thm:icgd_soft_main}) and convergence (\cref{coro:error_ICGD_m}) guarantees.
We also provide experimental validation.

\textbf{Extensions.}
We further extend our analysis  
from $N$-layer networks with the same input and output dimensions to scenarios with arbitrary dimensions (\cref{app:sec_diff_io}).

\textbf{Applications.} We apply our results to learn the score function of the diffusion model through ICL in \cref{app:sec_icl_diff}.

\textbf{Related Work and Limitations.}
Please see the related works, a detailed comparison with \cite{wangcontext}, broader impact, and limitations in \cref{sec:broad_limit_relate}.

%% file: impact.tex
By the theoretical nature of this paper, we do not expect immediate negative social impact.

%% file: x_acknowledgments.tex
The authors would like to thank Zhijia Li, Mimi Gallagher, Sara Sanchez, Dino Feng and Andrew Chen for helpful discussions; Hude Liu,
Hong-Yu Chen, Jennifer Zhang, and Teng-Yun Hsiao for collaborations on related topics;
and Jiayi Wang for facilitating experimental deployments.
JH also thanks the Red Maple Family for their support. 
The authors also thank the anonymous reviewers and program chairs for their constructive comments.

JH is partially supported by the Walter P. Murphy Fellowship.
HL is partially supported by NIH R01LM1372201  AbbVie and Dolby.
This research was supported in part through the computational resources and staff contributions provided for the Quest high performance computing facility at Northwestern University which is jointly supported by the Office of the Provost, the Office for Research, and Northwestern University Information Technology.
The content is solely the responsibility of the authors and does not necessarily represent the official
views of the funding agencies.

%% file: appendix.tex
{
\setlength{\parskip}{-0em}
\startcontents[sections]
\printcontents[sections]{ }{1}{}
}
\clearpage

\section{Related Work, Broader Impact, Further Discussion and Limitations}
\label{sec:broad_limit_relate}
In this section, we show the related works, broader impact and limitations.

\subsection{Related Work}
\label{sec:related_works}

\paragraph{In-Context Learning.}
Large language models
(LLMs) demonstrate the in-context learning (ICL)
ability \cite{brown2020language},
an ability to flexibly adjust their prediction based on
additional data given in context.
In recent years, a number of studies investigate enhancing ICL capabilities \cite{chen2022improving,gu2023pre,shi2023context}, exploring influencing factors \cite{shin2022effect,yoo2022ground}, and interpreting ICL theoretically \cite{xie2021explanation,wies2024learnability,panwar2023context,li2023closeness,bai2023transformers,dai2022can}.
The works most relevant to ours are as follows.
\cite{von2023transformers} showed that linear attention-only Transformers with manually set parameters closely resemble models trained via gradient descent.
\cite{bai2023transformers} providing a more efficient construction for in-context
gradient descent and established quantitative error
bounds for simulating multi-step gradient descent.
However, these results focused on simple ICL algorithms or specific tasks like least squares, ridge regression, and gradient descent on two-layer neural networks. 
These algorithms are inadequate for practical applications. 
For example: (i) Approximating the diffusion score function requires neural networks with multiple layers \cite{chen2023score}. 
(ii) Approximating the indicator function requires at least $3$-layer networks \cite{safran2017depth}. 
Therefore, the explicit construction of transformers to implement in-context gradient descent (ICGD) on deep models is necessary to better align with real-world in-context settings.
Our work achieves this by analyzing the gradient descent on $N$-layer neural networks through the use of ICL.
We provide a more efficient construction for in-context gradient descent.
Furthermore, we extend our analysis to $\Softmax$-transformer in \cref{app:sec_soft} to better align with real-world uses.

\paragraph{In-Context Gradient Descent on Deep Models \cite{wangcontext,panigrahi2023trainable}.} 
A work similar to ours is \cite{wangcontext}.
It constructs a family of transformers with flexible activation functions to implement multiple steps of ICGD on deep neural networks. 
This work emphasizes the generality of activation functions and demonstrates the theoretical feasibility of such constructions.
Our work adopts a different approach by enhancing the efficiency of transformers and better aligning with practical applications.
We introduce the following novelties:
\begin{itemize}
    \item {\bf More Structured and Efficient Transformer Architecture.} 
    While the work \cite{wangcontext} uses a $O(N^2L)$-layer transformer to approximate $L$ gradient descent steps on $N$-layer neural networks, our approach achieves more efficient simulation for ICGD.
    We approximate specific terms in the gradient expression to reduce computational costs, requiring only a $(2N + 4)L$-layer transformer for $L$ gradient descent steps. 
    Our method focuses on selecting and approximating the most impactful intermediate terms in the explicit gradient descent expression (\cref{lem:aprox_r_p_m,lem:aprox_pl_m,lem:aprox_s_j_m}), optimizing layer complexity to $O(NL)$.
    
    \item {\bf Less Restrictive Input and Output Dimensions for $N$-layer Neural Networks.} The work \cite{wangcontext} simplifies the output of $N$-layer networks to a scalar. 
    Our work expands this by considering cases where output dimensions exceed one, as detailed in \cref{app:sec_diff_io}. 
    This includes scenarios where input and output dimensions differ.

    \item {\bf More Practical Transformer Model.}
    The work \cite{wangcontext} discusses activation functions in the attention layer that meet a general decay condition (\cite[Definition 2.3]{wangcontext}) without considering the $\Softmax$ activation function. 
    We extend our analysis to include $\Softmax$-transformers.
    Our analysis reflects more realistic applications, as detailed in \cref{app:sec_soft}.

    \item {\bf More Advanced and Complicated Applications.}
    The work \cite{wangcontext} discusses the applications to functions, including indicators, linear, and smooth functions. 
    We explore more advanced and complicated scenarios, i.e., the score function in diffusion models discussed in \cref{app:sec_icl_diff}. 
    The score function \cite{chen2023score} falls outside the smooth function class. 
    This enhancement broadens the applicability of our results.
    
\end{itemize}

Another work similar to ours is \cite{panigrahi2023trainable}.
It proposes a new efficient construction, Transformer in Transformer (TINT), to allow a transformer to simulate and finetune more complex models (e.g., one transformer).
The main distinction between ours and \cite{panigrahi2023trainable} lies in the different aims: 
Our approach focuses on using a standard transformer for the simulator (with a minor modification: the ``element-wise multiplication layer''), and we provide a theoretical understanding of how a standard transformer can learn the ICGD of an $N$-layer network using ICL.
In contrast, the work \cite{panigrahi2023trainable} aims to build even stronger transformers by introducing several structural modifications that enable running gradient descent on auxiliary transformers.
While it demonstrates in-context gradient descent for a more advanced model, i.e., one transformer, our work offers the following potential advantages:
\begin{itemize}
    \item \textbf{Explicit Transformer Construction.}
    We provide an explicit construction of the transformer, whereas the work \cite{panigrahi2023trainable} does not detail the explicit construction of model parameters within their transformer.

    \item \textbf{Exact Gradient Descent.}
    We compute the exact and explicit gradient descent for an $N$-layer network (\cref{lem:decomposition_gd_m}). 
    Building on this, we employ the transformer's ICL to perform gradient descent on all parameters. 
    However, the work \cite{panigrahi2023trainable} stops the gradient computation through attention scores in the self-attention layer and only updates the value parameter in the self-attention module. 
    Additionally, it uses Taylor expansion to approximate the gradient.

    \item \textbf{Rigorous Error and Convergence Guarantees.}
    We provide rigorous gradient descent approximation errors (for multiple steps) and convergence guarantees for the ICGD on an $N$-layer network (\cref{coro:error_ICGD_m} and \cref{lem:error_general_m}). 
    However, the work \cite{panigrahi2023trainable} only presents the gradient approximation error for each specific part of the parameters in a single step.

    \item \textbf{Attention Layer Better Aligned with Practice.}
    Our analysis is based on ReLU-attention (\cref{thm:icl_gd_m}) or Softmax-attention (\cref{thm:icgd_soft}), whereas the work \cite{panigrahi2023trainable} utilizes linear attention. Our choice of attention layer better aligns with practical applications.
\end{itemize}

\subsection{Broader Impact}
This theoretical work aims to shed light on the foundations of large transformer-based models and is not expected to have negative social impacts.

\subsection{Further Discussion}
\label{sec:further_discussion}
We provide an interpretation and example of how to explicitly instantiate the constants for ReLU approximations in \cref{lem:aprox_p_m,lem:aprox_r_p_m,lem:aprox_pl_m}. 
The key reason is that the function approximated by the sum of ReLUs is simple in our context, such as the Sigmoid activation function. 
For such simple functions, it is straightforward to derive an explicit construction. 

Here, we take the Sigmoid activation function as an example and propose one explicit construction method. Let $r(z)$ denote the Sigmoid function.
\begin{itemize}
    \item 
    \textbf{Segment the Input Domain.} 
    For example, divide the domain $[-10, 10]$ smaller intervals such as $[-10, -9]$, $[-9, -8]$, $\dots$, $[9, 10]$.

    \item 
    \textbf{Approximate Each Segment Locally Using a Linear Function via Linear Interpolation.}
    For instance, in the domain $[9,10]$, approximate $r(z)$ using a linear function $a_1 z + c_1$, where $a_1$ and $c_1$ are calculated as follows:
    $a_1 = (r(10)-r(9)) / (10-9)$, and $c_1 = r(9) - a_1 * 9$.

    \item 
    \textbf{Approximate Linear Function $a_1 z + c_1 (z \in [9,10])$ Using a Sum of ReLU Terms.}
    This step involves two substeps, which are straightforward to implement:
    (i) Approximate the indicator function for $z \in [9,10]$ using a sum of ReLU terms.
    (ii) Approximate the constant $c_1$ using the sum of ReLU. This is because bias terms are not included in the sum of ReLU terms in \cref{def:sum_of_relus_m}. The bias term $c_1$ must be approximated using an additional sum of ReLU terms.

    \item 
    \textbf{Combine All the Sum of ReLU Approximators Across All Segments.}
    Finally, integrate the approximations for all segments to construct the complete approximation.

    \item 
    \textbf{Estimation of the Parameters in \cref{def:sum_of_relus_m}.} $\epsilon_{\rm approx} = 0.625$, $R=10$, $H=80$, and $C=25$.
\end{itemize}
Furthermore, to achieve higher precision in the approximation, it is sufficient to use finer segmentations.

\subsection{Limitations}
Our work has the following six limitations:
\begin{itemize}
    \item Although we provide a theoretical guarantee for the ICL of the $\Softmax$-Transformer to approximate gradient descent in $N$-layer NN, characterizing the weight matrices construction in $\Softmax$-Transformer remains challenging.
    This motivates us to rethink transformer universality and explore more accurate proof techniques for ICL in $\Softmax$-Transformer, which we leave for future work.

    \item The hidden dimension and MLP dimension of the transformer in \cref{thm:icl_gd_m} are both $\Tilde{O}(NK^2)+D_w$, which is very large.
    The reason for the large dimensions is that if we use ICL to perform ICGD on the $N$-layer network, we need to allow the transformer to realize the $N$-layer network parameters. 
    This means that it is reasonable for the input dimension to be so large. However, it is possible to reduce the hidden dimension and MLP dimension of the transformer through smarter construction.
    We leave this for future work.

    \item The generalization capabilities are limited compared with traditional transformers. 
    In our setting, the pretraining task refers to using in-context examples generated by an $N$-layer network for a given $N$. 
    Specifically, during pretraining, the distribution of the $N$-layer network parameters is predetermined (e.g., $N(0, I)$). 
    The input data distribution of $N$-layer network for generating the in-context examples is also predetermined (e.g., $N(-2, I)$). 
    The generalization capabilities include the following two aspects:
    (i) Varying the input data distribution for the $N$-layer network to generate the in-context examples. For example, we change the input data distribution from $N(-2, I)$ to $0.9N(-2, I) + 0.1N(2, I)$ during the testing in \cref{subsec:obj_1_2}.
    (ii) Varying the distribution of the $N$-layer network parameters. 
    For example, we change the distribution from $N(0, I)$ to $N(0.5, I)$ in \cref{subsec:obj_3}. 
    The above points lead to differences between the distributions of in-context examples during pretraining and testing. 
    However, we must generate the in-context examples by the $N$-layer network with the same hyperparameters, including the network width and depth.
    We leave the theoretical analysis of broader generalization capabilities for future work.

    \item In theory, the FLOPs \cite{hoffmann2022training} required to perform one forward pass of the transformer are greater than those required for the direct training of an $N$-layer network.
    (i) For the forward pass of the transformer, the FLOPs for in-context learning (ICL) are $O(nLN^3K^5 / \epsilon^2)$, where $\epsilon$ is the approximation error in the sum of ReLU.
    (ii) For direct training of the $N$-layer network, the FLOPs without ICL are $O(nLNK^2)$.
    Therefore, the FLOPs required for ICL exceed those needed for direct training of the $N$-layer network. However, experimental results in \cref{app:sec_exp} demonstrate that the transformer with ICL can achieve the performance of a trained $6$-layer network using fewer FLOPs in practice (3.3 billion vs. 7.6 billion FLOPs). This finding encourages further exploration of more efficient architectures. We also leave this topic for future research.

    \item 
    The empirically trained transformer differs from the transformer constructed in our theoretical analysis. 
    Our experiments confirm the existence of a transformer capable of simulating gradient descent (GD) steps for $N$-layer neural networks through in-context learning (ICL). 
    Despite this discrepancy, the limitation does not affect the primary contribution: establishing the theoretical existence of this transformer by explicit construction.

    \item There are two minor differences between the transformer used in the theoretical analysis and a standard transformer:
    (i) The transformer used in the theoretical analysis incorporates an element-wise multiplication layer, a specialized variant of self-attention that retains only the diagonal score and allows efficient implementation. 
    (ii) It does not alternate self-attention and MLP layers. 
    We emphasize that this also qualifies as a standard transformer because we view either an attention or an MLP layer as equivalent to an attention plus MLP layer due to the residual connections.
    
\end{itemize}

\clearpage
\section{Supplementary Theoretical Backgrounds}

Here we present some ideas we built on.

\subsection{Transformers}
\label{sec:transformer}
Lastly, we introduce key components for constructing a transformer for ICGD: ReLU-Attention, MLP, and element-wise multiplication layers. 
We begin with the ReLU-Attention layer.

\begin{definition}[ReLU-Attention Layer]
\label{def:attn}
For any input sequence $H \in \R^{D \times n}$,  an $M$-head ReLU-attention layer with parameters $\theta=\{Q_m, K_m, V_m\}_{m \in [M]}$ outputs
\begin{align*}
{\rm Attn}_{\theta}(H) \coloneqq H + \frac{1}{n} \sum_{m=1}^M  (V_m H) \cdot \sigma((Q_m H)^{\top} 
(K_m H)),
\end{align*}
where $Q_m,K_m,V_m\in\R^{D\times D}$ and $\sigma(\cdot)$ is element-wise ReLU activation function.
In vector form, for each token $h_i\in\R^D$ in $H$, it outputs $[{\rm Attn}_{\theta}(H)]_i = h_i + \frac{1}{n} \sum_{m=1}^M  \sum_{s=1}^{n} \sigma (\langle Q_m h_i, K_m h_s \rangle) \cdot V_m h_s$.
\end{definition}
Notably, \cref{def:attn} uses normalized ReLU activation $\sigma/n$, instead of the standard $\Softmax$.
We adopt this for technical convenience following \cite{bai2023transformers}. 
Next we define the  MLP layer.

\begin{definition} [MLP Layer]
\label{def:mlp}
For any input sequence $H \in \R^{D \times n}$,  an $d'$-hidden dimensions MLP layer with parameters
$\theta = (W_1, W_2)$ outputs ${\rm MLP}_{\theta}(H) := H + W_2 
    \sigma(W_1 H)$,
where $W_1 \in \R^{d' \times D}$, $W_2 \in \R^{D \times d'}$ and $\sigma(\cdot):\R\rightarrow\R$ is element-wise ReLU activation function. 
In vector form, for each token $h_i\in\R^D$ in $H$, it outputs 
$\text{MLP}_{\theta}(H)_i := h_i + W_2 
\sigma(W_1 h_i)$.
\end{definition}

Then, we consider a transformer architecture with $L \geq 1$ transformer layers, each consisting of a self-attention
layer followed by an MLP layer.

\begin{definition}[Transformer]
\label{def:tf}
For any input sequence $H \in \R^{D \times n}$,  an $L$-layer transformer with parameters $\theta =\{\theta_{\rm Attn},\theta_{\rm MLP}\}$ outputs
\begin{align*}
    {\rm TF_{\theta}^{L}}(H) := {\rm MLP}_{{\theta}^{(L)}_{\rm mlp}} \circ {\rm Attn}_{{\theta}^{(L)}_{\rm attn}}\ldots {\rm MLP}_{{\theta}^{(1)}_{\rm mlp}}\circ {\rm Attn}_{{\theta}^{(1)}_{\rm attn}}(H),   
\end{align*}
where $\theta =\{\theta_{\rm Attn},\theta_{\rm MLP}\}$ consists of Attention layers $\theta_{\rm Attn} = 
\{(Q_m^l, K_m^l, V_m^l)\}_
{l \in [L],m \in [M^l]}$ and MLP layers $\theta_{\rm MLP} = \{(W_1^l,W_2^l)\}_{l \in [L]}$. 
Above, for any $l \in [L], m \in [M^l]$, $Q_m^l,K_m^l,V_m^l \in\R^{D\times D}$ and $(W_1^l,W_2^l) \in \R^{d' \times D} \times \R^{D \times d'}$
In this section, we consider ReLU Attention layer and MLP layer are both a  special kind of $1$-layer transformer, which is for technical convenience. 
\end{definition}

For later proof use, we define the norm for $L$-layer transformer 
$\rm TF_{\theta}$ as:
\begin{align}
\label{eqn:tf_norm}
B_{\theta}
:= \max_{l \in [L]} 
\left\{ \max_{m \in [M]} \left\{ \|Q_{m}^l\|_1, 
\| K_{m}^l\|_1 \right\} + \sum_{i=1}^m \|V_{m}^l\|_1 + \|W_1\|_1 + \|W_2\|_1 \right\}.
\end{align}

The choice of operation norm and max/sum operation is for convenience in later proof only, as our result depends only on $B_{\theta}$.

\subsection{ReLU Provably Approximates Smooth \texorpdfstring{$k$}{}-Variable Functions}

Following lemma expresses that the smoothness enables the approximability of sum of ReLU.

\begin{lemma}
[Approximating Smooth $k$-Variable Functions, modified from Proposition A.1 of \cite{bai2023transformers}]
\label{lem:approx_k_v}
For any $\epsilon, C_l >0, R \geq 1$.
If function $g:\R^k \rightarrow \R$ such that for $s := \lceil (k-1)/2 \rceil + 1$, $g$ is a $C^s$ function on $B_{\infty}^k(R)$, and for all $i \in \{0,1,\ldots,s\}$,
\begin{align*}
    \sup_{z \in B_{\infty}^k(R)} \|\nabla^i g(z)\|_{\infty}
    \leq L_i, \quad
    \max_{0 \leq i \leq s}
    L_i R^i \leq C_l,
\end{align*}
then function $g$ is $(\epsilon, R, H, C)$-approximable by sum of ReLUs (\cref{def:sum_of_relus_m}) with $H \leq C(k) C_l^2 \log(1+C_l/\epsilon)/\epsilon^2$ and $C \leq C(k) C_l$ where $C(k)$ is a constant that depends only on $k$.
\end{lemma}

\clearpage
\section{Proofs of Main Text}
\label{app:sec_proof_main}

\subsection{Proof of \texorpdfstring{\cref{lem:decomposition_gd_m}}{}}
\label{proof:lem:decomposition_gd_m}

\begin{lemma}
[\cref{lem:decomposition_gd_m} Restated: Decomposition of One Gradient Descent Step]
    Fix any $B_v, \eta > 0$.
    Suppose loss function $\mathcal{L}_n(w)$ on $n$ data points $\{(x_i,y_i)\}_{i \in [n]}$ follows \eqref{eqn:loss}.
    Suppose closed domain $\mathcal{W}$ and projection function
    ${\rm Proj}_{\mathcal{W}}(w)$ follows \eqref{eqn:domain_w_m}.
    Let $A_i(j), r'_i(j), R_i(j), V_j$ be as defined in \cref{def:abbv_m}.
    Then the explicit form of gradient $\nabla \mathcal{L}_n(w)$ becomes
    \begin{align*}
        \nabla \mathcal{L}_n(w) =  
        \frac{1}{2n} \sum_{i=1}^{n}
         \begin{bmatrix}
            A_i(1) \\
            \vdots \\
            A_i(N)
         \end{bmatrix},
    \end{align*}
    where $A_i(j)$ denote the derivative of $\ell(p_i(N), y_i)$ with respect to the parameters in the $j$-th layer, 
    \begin{align*}
        A_i(j) = \begin{cases}
            (R_i(N-1) \cdot V_{N} \cdot \ldots \cdot R_i(j-1) \cdot
            \begin{bmatrix}
            \textbf{I}_{K \times K} \otimes 
            p_i(j-1)^{\top} 
            \end{bmatrix}
            )^{\top} \cdot (\pdv {\ell(p_i(N), y_i)}{p_i(N)})^\top, & j \neq N \\
            (R_i(N-1) \cdot
            \begin{bmatrix}
            \textbf{I}_{d \times d} \otimes 
            p_i(N-1)^{\top} 
            \end{bmatrix}
            )^{\top} \cdot (\pdv {\ell(p_i(N), y_i)}{p_i(N)})^\top, & j = N.
        \end{cases}
    \end{align*}
\end{lemma}

\begin{proof}[Proof of \cref{lem:decomposition_gd_m}]
We start with calculating 
$\nabla_w \mathcal{L}_n(w)$. 
By chain rule and \eqref{eqn:loss}, 
\begin{align*}
    \underbrace{\nabla_w \mathcal{L}_n(w)}_{\R^{D_N \times 1}}
    = & ~ \frac{1}{2n} \sum_{i=1}^n 
    \underbrace{[\pdv{}{w} p_i(N)]^\top}
    _{\R^{D_N \times d}} \cdot
    \underbrace{[\pdv{}{p_i(N)} 
    \ell(p_i(N), y_i)]^\top}_{\R^{d \times 1}}
    \annot{By \eqref{eqn:loss}
    and chain rule}
\end{align*}
Thus we only need to calculate 
$\pdv{}{w} p_i(N)$.
For a vector $x$ and a function $r : \R \rightarrow \R$, we use $\textbf{r}(x)$ to denote the vector that $i$-th coordinate is $r(x_i)$.
Let $R_i(j), V_j$ follows \cref{def:abbv_m}, then it holds
\begin{align}
    \underbrace{\pdv{p_i(N)}{w}}_{\R^{d \times D_N}} \nonumber 
    = & ~ \underbrace{\pdv{\textbf{r}(\overbrace{V_{N}}^{\R^{d \times K}} \cdot \overbrace{p_i(N-2)}^{\R^K})}{w}}
    _{\R^{d \times D_N}} 
    \annot{By \cref{def:N_nn_m}} \nonumber \\
    = & ~ \underbrace{\pdv{\textbf{r}
    (V_{N} \cdot p_i(N-1))}{V_{N} \cdot 
    p_i(N-1)}}_{\R^{d \times d}} \cdot
    \underbrace{\pdv{V_{N} \cdot p_i(N-1)}{w}}_{\R^{d \times D_N}} \annot{By chain rule}\nonumber \\
    = & ~ \diag \{
    r'(v_{{N}_1}^{\top} p_i(N-1)),\ldots, r'(v_{{N}_K}^{\top} p_i(N-1))\} 
    \cdot \pdv{ V_{N} \cdot p_i(N-1)}{w} \annot{By \cref{def:abbv_m}} \nonumber\\
    = & ~ R_i(N-1) \cdot \pdv{ V_{N} \cdot p_i(N-1)}{w}.
    \label{eqn:A_k_p1_m}
\end{align}
Notice that for any $k \in [d]$, $v_{N_{k}}$ is a part of $w$, thus 
\begin{align}
    \pdv{v_{N_{k}}}{w} = & ~ 
    [\overbrace{\mathbf{0}}
    ^{D_{N-1}+(k-1)K} \overbrace{\mathbf{I}}^{K}
    \overbrace{\mathbf{0}}^{D_N - D_{N-1} - kK}] \in 
    \R^{d \times D_N}.
    \label{eqn:drv_v_m}
\end{align}
Therefore, letting $\otimes$ denotes Kronecker product, it holds
\begin{align}
    & ~ \pdv{V_{N} \cdot p_i(N-1)}{w} \nonumber \\
    = & ~ \begin{bmatrix}
        v_{{N}_1}^{\top} \cdot 
        \pdv{p_i(N-1)}{w} +p_i(N-1)^{\top} \cdot \pdv{v_{{N}_1}}{w} \nonumber \\
        \vdots \nonumber \\
        v_{{N}_d}^{\top} \cdot .
        \pdv{p_i(N-1)}{w} +p_i(N-1)^{\top} \cdot \pdv{v_{{N}_d}}{w}
    \end{bmatrix} \annot{By chain rule and product rule}
    \nonumber \\
    = & ~ V_{N} \cdot 
    \pdv{p_i(N-1)}{w} +
    \begin{bmatrix}
        \textbf{0}_{D_{N-1}} ; \textbf{I}_{K \times K} \otimes p_i(N-1)^{\top}
    \end{bmatrix},
    \label{eqn:A_k_p2_m}
\end{align}
where the last step follows from the definition of 
$V_{N}$ (i.e., \cref{def:abbv_m}) and
\eqref{eqn:drv_v_m}.

Substituting \eqref{eqn:A_k_p2_m} into \eqref{eqn:A_k_p1_m}, we obtain 
\begin{align*}
    \pdv{p_i(N)}{w} 
    =  R_i(N-1) \cdot
    (V_{N} \cdot 
    \pdv{p_i(N-1)}{w} +
    \begin{bmatrix}
        \textbf{0}_{D_{N-1}} ; \textbf{I}_{d \times d} \otimes p_i(N-1)^{\top}
    \end{bmatrix}). 
\end{align*}

Similarly, for any $j \in [N]$, we prove
\begin{align}
    \label{eqn:A_k_p3_m}
    \pdv{p_i(j)}{w}  
    =  R_i(j-1) \cdot
    (V_{j} \cdot 
    \pdv{p_i(j-1)}{w} +
    \begin{bmatrix}
        \textbf{0}_{D_{j-1}} ; \textbf{I}_{K \times K} \otimes p_i(j-1)^{\top};
        \textbf{0}_{D_{N}-D_{j}}
    \end{bmatrix}). 
\end{align}

By the recursion formula \eqref{eqn:A_k_p3_m}, for any $j \in [N-1]$, we calculate $A_i(j)$ as follows,
\begin{align*}
    A_i(j) = & ~ 
    \left( \left(\pdv
    { \ell(p_i(N), y_i)}{p_i(N)} \cdot \pdv{p_i(N)}{w} \right)^\top \right)
    [D_{j-1}:D_j]  \annot{By \cref{def:abbv_m}}\nonumber \\
    = & ~ (\pdv{p_i(N)}{w})^\top  
    \cdot (\pdv {\ell(p_i(N), y_i)}{p_i(N)})^\top [D_{j-1}:D_j] 
    \annot{By transpose property}\\
    = & ~ (\pdv{p_i(N)}{w})^\top 
    [*,D_{j-1}:D_j]
    \cdot (\pdv {\ell(p_i(N), y_i)}{p_i(N)})^\top  \\
    = & ~ (R_i(N-1) \cdot V_{N} \cdot \ldots \cdot R_i(j-1) \cdot
    \begin{bmatrix}
        \textbf{I}_{K \times K} \otimes 
        p_i(j-1)^{\top} 
    \end{bmatrix}
    )^{\top} \cdot (\pdv {\ell(p_i(N), y_i)}{p_i(N)})^\top,
    \annot{By \eqref{eqn:A_k_p3_m}}
\end{align*}
where $M[*,a:b]$ denotes a sub-matrix of $M$, which includes all the columns but only the rows from the $a$-th row to the $b$-th row of $A$. 
Similarly, for $j=N$, it holds
\begin{align*}
    A_i(N) = (R_i(N-1) \cdot
    \begin{bmatrix}
        \textbf{I}_{d \times d} \otimes 
        p_i(N-1)^{\top} 
    \end{bmatrix}
    )^{\top} \cdot (\pdv {\ell(p_i(N), y_i)}{p_i(N)})^\top.
\end{align*}
Thus we completes the proof.
\end{proof}

\subsection{Proof of \texorpdfstring{\cref{lem:aprox_p_m}}{}}
\label{proof:lem:aprox_p_m}

\begin{lemma}[\cref{lem:aprox_p_m} Restated: Approximate $p_i(j)$]
    Let upper bounds $B_v, B_x> 0$ such that
    for any $k \in [K], j \in [N] ~\text{and}~ i \in [n]$, $\|v_{j_k}\|_2 \leq B_v$,
    and $\|x_i\|_2 \leq B_x$.
    For any $j \in [N], i \in [n]$, define
    \begin{align*}
        B_r^j :=  \max_{\abs{t} \leq B_v B_r^{j-1}} \abs{r(t)},
        \quad B_r^0 := B_x ,\quad\text{and}\quad
        B_r := & ~ \max_{j} B_r^j.
    \end{align*}

    Let function $r(t)$ be $(\epsilon_r, R_1, M_1, C_1)$-approximable for $R_1 = \max \{B_v B_r, 1\}$, $M_1 \leq \Tilde{\mathcal{O}}(C_1^2 \epsilon_r^{-2})$, where $C_1$ depends only on $R_1$ and the $C^2$-smoothness of $r$.
    Then, for any $\epsilon_r>0$, there exist $N$ attention layers ${\rm Attn}_{\theta_1}, \ldots, {\rm Attn}_{\theta_N}$ such that for any input $h_i \in \R^D$ takes from \eqref{eqn:input}, they map
    \begin{align*}
        h_i = [x_i; y_i; w; \bar{p}_i(1); \ldots; \bar{p}_i(j-1); \mathbf{0}; 1; t_i] \xrightarrow
        {{\rm Attn}_{\theta_j}} 
        \Tilde{h_i} = [x_i; y_i; w; \bar{p}_i(1); \ldots; \bar{p}_i(j); \mathbf{0}; 1; t_i],
    \end{align*}
    where $\bar{p}_i(j)$ is  approximation for $p_i(j)$ (\cref{def:N_nn_m}).
    In the expressions of $h_i$ and $\Tilde{h}_i$, the dimension of $\mathbf{0}$ differs.
    Specifically, the $\mathbf{0}$ in $h_i$ is larger than in $\Tilde{h}_i$.
    The dimensional difference between these $\mathbf{0}$ vectors equals the dimension of $\bar{p}_i(j)$.
    Suppose function $r$ is $L_r$-smooth in bounded domain $\mathcal{W}$, then for any $i \in [n+1]$, $j \in [N]$, $\bar{p}_i(j)$ such that
    \begin{align*}
        \bar{p}_i(j) = p_i(j) + \epsilon(i,j), \quad  \|\epsilon(i,j)\|_2 \leq 
        \begin{cases}
            (\sum_{l=0}^{j-1} 
            K^{l/2} L_r^l B_v^l) \sqrt{K} \epsilon_r ~, &  \quad 1 \leq j \leq N-1 \\ 
            (\sum_{l=0}^{N-1} 
            K^{l/2} L_r^l B_v^l) \sqrt{d} \epsilon_r 
            ~, & \quad j=N
        \end{cases}.
    \end{align*} 
    Additionally, for any $j \in [N]$, the norm of parameters $B_{\theta_j}$ defined as \eqref{eqn:tf_norm}  such that
    \begin{align*}
        B_{\theta_j} \leq 
        1 + K C_1.
    \end{align*}
\end{lemma}

\begin{proof}[Proof of \cref{lem:aprox_p_m}]
    First we need to give a approximation for activation function $r(t)$.
    By our assumption and \cref{def:sum_of_relus_m}, $r(t)$ is $(\epsilon_r, R_1, M_1, C_1)$-approximable by sum of ReLUs, there exists:
    \begin{align}
        \bar{r}(t) 
        = \sum_{m=1}^{M_1} c_m^1 \sigma(\langle a_m^1, [t;1]\rangle) ~ \text{with}~
        \sum_{m=1}^{M_1} 
        \abs{c_m^1} \leq C_1, ~
        \|a_m^1\|_1 \leq 1, ~
        \forall m \in [M_1],
        \label{eqn:r_bar_m}
    \end{align}
    such that $\sup_{t \in [-R_1, R_1]}
    \abs{\bar{r}(t)-r(t)} \leq \epsilon_r$.
    Let $\bar{p}_i(0) := p_i(0) = x_i$.
    Similar to $p_i(j)$ follows \cref{def:N_nn_m}, we pick $\bar{p}_i(j)$ such that for any $j \in [N]$,
    \begin{align}
        \bar{p}_i(j)[k] := 
            \bar{r}(v_{j_{k}}^{\top} 
            \bar{p}_i(j-1)).
        \label{eqn:p_bar_m}
    \end{align}
    Fix any $j \in [N]$, suppose the input sequences $h_i = [x_i; y_i; w; \bar{p}_i(1); \ldots; \bar{p}_i(j-1); 
    \mathbf{0}; 1; t_i]$.
    Then for every $m \in [M_1], k \in [K] ({\rm or}~ k \in [d] ~{\rm if}~ j=N)$, we define matrices $Q_{m,k}^{j},K_{m,k}^{j},V_{m,k}^{j} \in \R^{D \times D}$ such that for all $i \in [n+1]$,
    \begin{align}
        Q_{m,k}^{j} h_i = \begin{bmatrix}
            a_m^1[1] \cdot \bar{p}_i(j-1) \\
            a_m^1[2] \\
           \mathbf{0}
        \end{bmatrix}, \quad
        K_{m,k}^{j} h_i = \begin{bmatrix}
            v_{j_k} \\
            1 \\
           \mathbf{0}
        \end{bmatrix}, \quad
        V_{m,k}^{j} h_i = c_m^1 e_{j,k}^1~, 
        \label{eqn:matrix_con_m}
    \end{align}
    where $e_{j,k}^1$  denotes the position unit vector of element $\bar{p}_i(j)[k]$
    because this position only depends on $j,k$.
    Since input $h_i = [x_i; y_i; w; \bar{p}_i(1); \ldots; \bar{p}_i(j-1); 
    \mathbf{0}; 1; t_i]$, those matrices indeed exist.
    In fact, it is simple to check that
    \begin{align}
        Q_{m,k}^{j} = & \begin{bmatrix}
            &\mathbf{0} & a_m^1[1] \mathbf{I}_K(j)
            &\mathbf{0} &\mathbf{0} 
            &\mathbf{0} \\
            &\mathbf{0} &\mathbf{0} &\mathbf{0} &a_m^1[2] &\mathbf{0} \\ 
            &\mathbf{0} &\mathbf{0} &\mathbf{0} &\mathbf{0}
            &\mathbf{0}
        \end{bmatrix}, \nonumber\\
        K_{m,k}^{j} = & \begin{bmatrix}
            &\mathbf{0} &\mathbf{I}_K(j,k) 
            &\mathbf{0} &\mathbf{0} 
            &\mathbf{0} \\
            &\mathbf{0} &\mathbf{0} &\mathbf{0} &1 &\mathbf{0} \\ 
            &\mathbf{0} &\mathbf{0} &\mathbf{0} &\mathbf{0}
            &\mathbf{0}
        \end{bmatrix}, \nonumber\\
        V_{m,k}^{j} = & \begin{bmatrix}
            &\mathbf{0} &\mathbf{0} &\mathbf{0} &\mathbf{0} \\
            &\mathbf{0} &\mathbf{0} &c_m^1(j,k) &\mathbf{0}  \\
            &\mathbf{0} &\mathbf{0} &\mathbf{0} &\mathbf{0}
        \end{bmatrix},
        \label{eqn:matrix_form_m}
    \end{align}
    are suffice to \eqref{eqn:matrix_con_m}.
    $\mathbf{I}_K(j), \mathbf{I}_K(j,k), c_m^1(j,k)$ represents their positions are related to variables in parentheses.
    In Addition, by \eqref{eqn:tf_norm}, notice that they have operator norm bounds 
    \begin{align*}
        \max_{j,m,k} \|Q_{m,k}^{j}\|
        _1
        \leq  1, \quad
        \max_{j,m,k} \|K_{m,k}^{j}\|
        _1
        \leq  1, \quad
        \max_{j} \sum_{k,m} \|V_{m,k}^{j}\|_1
        \leq  K C_1.
    \end{align*}
    Consequently, for any $j \in [N]$, $B_{\theta_j} \leq 1 + C_1$.

    By our construction follows \eqref{eqn:matrix_con_m}, a simple calculation shows that
    \begin{align*}
        & ~ \sum_{m \in[M_1], k \in[K]} \sigma(\langle Q_{m,k}^{j} h_i, K_{m,k}^{j} h_s \rangle) V_{m,k}^{j} h_s \\
        = & ~ \sum_{k=1}^K \sum_{m=1}^{M_1} 
        c_m^1 \sigma(\langle a_m^1, [v_{j_k}^{\top} 
        \bar{p}_i(j-1);1]\rangle) e_{j,k}^1 \annot{By our construction \eqref{eqn:matrix_con_m}} \\
        = & ~ \sum_{k=1}^K (\bar{r}(v_{j_k}^{\top} \bar{p}_i(j-1))) 
        e_{j,k}^1 \annot{By definition of $\bar{r}$ follows \eqref{eqn:r_bar_m}}\\
        = & ~
        [\mathbf{0};
        \bar{p}_i(j);\mathbf{0}].
        \annot{By definition of $\bar{p}_i(j)$ follows \eqref{eqn:p_bar_m}}
    \end{align*}
    Therefore, by definition of ReLU Attention layer follows \cref{def:attn}, the output $\Tilde{h}_i$ becomes
    \begin{align*}
        \Tilde{h}_i = & ~[{\rm Attn}_{\theta_j}(h_i)] \\
        = & ~ h_i + \frac{1}{n+1} \sum_{s=1}^{n+1} \sum_{m \in[M_1], k \in[K]} \sigma(\langle Q_{m,k}^{j} h_i, K_{m,k}^{j} h_s \rangle) V_{m,k}^{j} h_s \\
        = & ~ h_i + \frac{1}{n+1} \sum_{s=1}^{n+1} (n+1) [\mathbf{0};
        \bar{p}_i(j);\mathbf{0}]\\
        = & ~ [x_i; y_i; w; \bar{p}_i(1); \ldots; \bar{p}_i(j-1); 
        \mathbf{0}; 1; t_i] +
        [\mathbf{0},
        \bar{p}_i(j),\mathbf{0}]
        \\
        = & ~ [x_i; y_i; w; \bar{p}_i(1); \ldots; \bar{p}_i(j-1); \bar{p}_i(j); \mathbf{0}; 1; t_i].
    \end{align*}
    Therefore, let the attention layer 
    $\theta_j = \{(Q_{m,k}^{j},K_{m,k}^{j},
    V_{m,k}^{j})\}_{(k,m)}$, we construct $
    {\rm Attn}_{\theta_j}$ such that
    \begin{align*}
        h_i = [x_i; y_i; w; \bar{p}_i(1); \ldots; \bar{p}_i(j-1); \mathbf{0}; 1; t_i] \xrightarrow
        {{\rm Attn}_{\theta_j}} 
        \Tilde{h_i} = [x_i; y_i; w; \bar{p}_i(1); \ldots; \bar{p}_i(j); \mathbf{0}; 1; t_i].
    \end{align*}
    
    In addition, by setting $R_1 = \max 
    \{B_v B_r , 1\} $ , the lemma then follows directly by induction on $j$.
    For the base case $j=1$, it holds
    \begin{align*}
        \abs{\bar{p}_i(1)[k]-p_i(1)[k]} 
        = & ~  \abs{\bar{r}_i
        (v_{1_{k}}^{\top} x_i)[k]-r(v_{1_{k}}^{\top} 
        x_i)} \annot{By \cref{def:N_nn_m}} \\
        \leq & ~ \epsilon_r.
        \annot{By definition of $\bar{r}$ follows \eqref{eqn:r_bar_m}}
    \end{align*}
    Suppose the claim holds for iterate $j-1$ and  function $r$ is $L_r$-smooth in bounded domain $\mathcal{W}$.
    Then for iterate $j$, 
    \begin{align*}
      & ~ \abs{\bar{p}_i(j)[k]-p_i(j)[k]} \\
      \leq & ~ \abs{\bar{p}_i(j)[k]-r(v_{j_{k}}^{\top} 
      \bar{p}_i(j-1))} + \abs{r(v_{j_{k}}^{\top} 
      \bar{p}_i(j-1)) - p_i(j)[k]}
      \annot{By triangle inequality} \\
      \leq & ~ \epsilon_r + L_r 
      \|v_{j_{k}}^{\top}\|_2 
      \|\bar{p}_i(j-1)-p_i(j-1)\|_2 \annot{By \eqref{eqn:r_bar_m} and Cauchy–Schwarz
      inequality} \\
      \leq & ~ \epsilon_r + \sqrt{K} L_r B_v 
      (\epsilon_r \sum_{l=0}^{j-2} 
      K^{l/2} L_r^l B_v^l)
      \annot{By inductive hypothesis} \\
      \leq & ~ \epsilon_r 
      \sum_{l=0}^{j-1} 
      K^{l/2} L_r^l B_v^l,
    \end{align*}
    Thus, it holds
    \begin{align*}
        \|\bar{p}_i(j) - p_i(j)\|_2
        = & ~  \sqrt{\sum_{k=1}^{K} \abs{\bar{p}_i(j)[k]-p_i(j)[k]}^2} \\
        \leq  & ~ \sqrt{K} (\epsilon_r \sum_{l=0}^{j-1} 
        K^{l/2} L_r^l B_v^l).
    \end{align*}
    This finishes the induction.
    Then for the output layer $j=N$, it holds
    \begin{align*}
        \|\bar{p}_i(N) - p_i(N)\|_2
        = & ~  \sqrt{\sum_{k=1}^{d} \abs{\bar{p}_i(N)[k]-p_i(N)[k]}^2} \\
        \leq  & ~ \sqrt{d} (\epsilon_r \sum_{l=0}^{N-1} 
        K^{l/2} L_r^l B_v^l).
    \end{align*}
    Thus we complete the proof.
\end{proof}

\subsection{Proof of \texorpdfstring{\cref{lem:aprox_r_p_m}}{}}
\label{proof:lem:aprox_r_p_m}

\begin{lemma}[\cref{lem:aprox_r_p_m} Restated: Approximate $r'_i(j)$]
    Let upper bounds $B_v, B_x > 0$ such that
    for any $k \in [K], j \in [N] ~\text{and}~ i \in [n]$, $\|v_{j_k}\|_2 \leq B_v$,
    and $\|x_i\|_2 \leq B_x$.
    For any $j \in [N], i \in [n]$, define
    \begin{align*}
        B_r'^j :=  \max_{\abs{t} \leq B_v B_{r'}^{j-1}} \abs{r'(t)},
        \quad B_{r'}^0 := B_x,\quad\text{and}\quad
        B_{r'} :=  \max_{j} B_{r'}^j.
    \end{align*}
    Suppose function $r'(t)$ is $(\epsilon_{r'}, R_2, M_2, C_2)$-approximable for $R_2 = \max \{B_v B_{r'}, 1\}$, $M_2 \leq \Tilde{\mathcal{O}}(C_2^2 \epsilon_r'^{-2})$, where $C_2$ depends only on $R_2$ and the $C^2$-smoothness of $r'$.
    Then, for any $\epsilon_r > 0$, there exist an attention layer 
    $ {\rm Attn}_{\theta_{N+1}}$ such that for any input $h_i \in \R^D$ takes from \eqref{eqn:h_i_p_m}, it maps
    \begin{align*}
        h_i = [x_i; y_i; w; \bar{p}_i; \mathbf{0}; 1; t_i] \xrightarrow
        {{\rm Attn}_{\theta_{N+1}}} 
        \Tilde{h_i} = [x_i; y_i; w; \bar{p}_i; \bar{r}'_i; \mathbf{0}; 1; t_i],
    \end{align*}
    where $\bar{r}'_i(j)$ is  approximation for $r'_i(j)$ (\cref{def:abbv_m}) and $\bar{r}'_i := [\bar{r}'_i(0); \ldots; \bar{r}'_i(N-1)] \in \R^{(N-2)K+d}$.
    Similar to \cref{lem:aprox_p_m}, in the expressions of $h_i$ and $\Tilde{h}_i$, the dimension of $\mathbf{0}$ differs.
    In addition, let $E_r$ be defined in \eqref{eqn:e_r_m}, for any $i \in [n+1]$, $j \in [N], k \in [K]$, $\bar{r}'_i(j)$ such that 
    \begin{align*}
        \bar{r}'_i(j-1)[k] = 
        r'_i(j-1)[k] + \epsilon(i,j,k), \quad  \abs{\epsilon(i,j,k)} \leq 
        \epsilon_{r'} +
        L_{r'} B_v E_r \epsilon_r,
    \end{align*}
    where $\epsilon_r$ denotes the error generated in approximating
    $r$ by sum of ReLUs $\bar{r}$ follows \eqref{eqn:r_bar_m}.
    Additionally, the norm of parameters $B_{\theta_{N+1}}$ defined as \eqref{eqn:tf_norm} such that
    $B_{\theta_{N+1}} \leq
    1 + K (N-1) C_2$.
\end{lemma}

\begin{proof}[Proof of \cref{lem:aprox_r_p_m}]
    By \cref{def:abbv_m}, recall that 
    for any $j \in [N], i \in [n+1],k \in [K]$,
    \begin{align}
        r'_i(j)[k] = 
        r'(v_{{j+1}_k}^\top p_i(j)).
        \label{eqn:r_i(j)_m}
    \end{align}
    Therefore we need to give a approximation for $r'$.
    By our assumption and \cref{def:sum_of_relus_m}, $r'(t)$ is $(\epsilon_{r'}, R_2, M_2, C_2)$-approximable by sum of relus. 
    In other words, there exists:
    \begin{align}
        \bar{r}'(t) 
        = \sum_{m=1}^{M_2} c_m^2 \sigma(\langle a_m^2, [t;1]\rangle) ~ \text{with}~
        \sum_{m=1}^{M_2} 
        \abs{c_m^2} \leq C_2, ~
        \|a_m^2\|_2 \leq 1, ~
        \forall m \in [M_2],
        \label{eqn:r_p_bar_m}
    \end{align}
    such that $\sup_{t \in [-R_2, R_2]}
    \abs{\bar{r}'(t)-r'(t)} \leq \epsilon_{r'}$.
    Similar to \eqref{eqn:r_i(j)_m}, we pick $\bar{r}'_i(j)$ such that
    \begin{align}
        \bar{r}'_i(j)[k] := 
        \bar{r}'(v_{{j+1}_k}^\top \bar{p}_i(j)).
        \label{eqn:r_p_i(j)_m}
    \end{align}
    To ensure \eqref{eqn:r_p_i(j)_m}, we construct our attention layer as follows:
    for every $j \in [N], m \in [M_2], k \in [K]$, we define matrices $Q_{j,m,k}^{N+1},K_{j,m,k}^{N+1},
    V_{j,m,k}^{N+1} \in \R^{D \times D}$ 
    such that
    \begin{align}
        Q_{j,m,k}^{N+1} h_i = \begin{bmatrix}
            a_m^2[1] \cdot \bar{p}_i(j-1) \\
            a_m^2[2] \\
           \mathbf{0}
        \end{bmatrix}, \quad
        K_{j,m,k}^{N+1} h_i = \begin{bmatrix}
            v_{j_k} \\
            1 \\
           \mathbf{0}
        \end{bmatrix}, \quad
        V_{j,m,k}^{N+1} h_i = c_m^2 e_{j,k}^2~, 
        \label{eqn:matrix_con_2_m}
    \end{align}
    for all $i \in [n+1]$ and  $e_{j,k}^2$ denotes the position unit vector of element $\bar{r}'_i(j)[k]$.
    Since input $h_i = [x_i; y_i; w; \bar{p}_i; \mathbf{0}; 1; t_i]$, similar to \eqref{eqn:matrix_form_m}, those matrices indeed exist. 
   In addition, they have operator norm bounds 
    \begin{align*}
        \max_{j,m,k} \|Q_{j,m,k}^{N+1}\|_1
        \leq  1, \quad
        \max_{j,m,k} \|K_{j,m,k}^{N+1}\|_1
        \leq  1, \quad
        \sum_{j,m,k} \|V_{j,m,k}^{N+1}\|_1
        \leq K (N-1) C_2.
    \end{align*}
    Consequently,  by definition of parameter norm follows \eqref{eqn:tf_norm},
    $B_{\theta_{N+1}} \leq 1 + K (N-1) C_2$.

    A simple calculation shows that
    \begin{align*}
        & ~ \sum_{j \in [N], m \in [M_2], k \in [K]} \sigma(\langle Q_{j,m,k}^{N+1} h_i, K_{j,m,k}^{N+1} h_s \rangle) V_{j,m,k}^{N+1} h_s \\
       = & ~ \sum_{j=1}^{N} \sum_{k=1}^K \sum_{m=1}^{M_2}  c_m^2 \sigma(\langle a_m^2, [v_{j_k}^{\top} \bar{p}_i(j-1);1]\rangle) e_{j,k}^2 \annot{By our construction follows \eqref{eqn:matrix_con_2_m}}\\
        = & ~ \sum_{j=1}^{N} \sum_{k=1}^K (\bar{r}'(v_{j_k}^{\top}
        \bar{p}_i(j-1))) 
        e_{j,k}^2 \annot{By definition of $\bar{r}'$ follows \eqref{eqn:r_bar_m}} \\
        = & ~[\mathbf{0};\bar{r}'_i(0);
        \ldots;
        \bar{r}'_i(N-1);\mathbf{0}]
        \annot{By definition of $\bar{r}'_i(j)$ follows \eqref{eqn:r_p_i(j)_m}}\\
        = & ~ [\mathbf{0};\bar{r}'_i;\mathbf{0}],
        \annot{By definition of $\bar{r}'_i$}
    \end{align*}
    Therefore, by definition of ReLU Attention layer follows \cref{def:attn}, the output $\Tilde{h}_i$ becomes
    \begin{align*}
        \Tilde{h}_i = & ~[{\rm Attn}_{\theta_N}(h_i)] \\
        = & ~ h_i + \frac{1}{n+1} \sum_{s=1}^{n+1} \sum_{j \in [N-1], m \in[M_2], k \in[K]} \sigma(\langle Q_{j,m,k}^{N} h_i, K_{j,m,k}^{N} h_s \rangle) V_{j,m,k}^{N} h_s  \\
        = & ~ h_i + \frac{1}{n+1} \sum_{s=1}^{n+1} (n+1) 
        [\mathbf{0};
        \bar{r}'_i;\mathbf{0}] \\
        = & ~ [x_i; y_i; w; \bar{p}_i; 
        \mathbf{0}; 1; t_i] +
        [\mathbf{0};
        \bar{r}'_i;\mathbf{0}]\\
        = & ~ [x_i; y_i; w; \bar{p}_i; \bar{r}'_i; \mathbf{0}; 1; t_i].
    \end{align*}
    Next, we calculate the error accumulation in this approximation layer. 
    By our assumption, $R_2 = \max 
    \{B_v B_{r'} , 1\}$.
    Thus, for any $j \in 
    [N], k \in [K], i \in [n+1]$, it holds
    \begin{align*}
        v_{j_k}^{\top} p_i(j-1) \leq R_2.
    \end{align*}
    As our assumption, we suppose function $r'$ is $L_r$-smooth in bounded domain $\mathcal{W}$.
    Combining above, the upper bound of error accumulation $\abs{\bar{r}'_i(j)[k] - r'_i(j)[k]} $ becomes
    \begin{align*}
        & ~ \abs{\bar{r}'_i(j)[k] - 
        r'_i(j)[k]} \\
        \leq & ~ \abs{\bar{r}'_i(j)[k]-r'(v_{j_{k}}^{\top} 
        \bar{p}_i(j-1))} + \abs{r'(v_{j_{k}}^{\top} 
        \bar{p}_i(j-1)) - 
        r'_i(j)[k]} \annot{By triangle inequality} \\
        \leq & ~ \epsilon_{r'} + 
        L_{r'} \|v_{j_{k}}^{\top}\|_2 
        \|\bar{p}_i(j-1) -
        p_i(j-1)\|_2
        \annot{By \eqref{eqn:r_p_bar_m} and Cauchy–Schwarz
        inequality} \\
        \leq & ~ \epsilon_{r'} +
        L_{r'} B_v E_r \epsilon_r.
        \annot{By definition of $E_r$ follows \eqref{eqn:e_r_m}}
    \end{align*}
     Thus we complete the proof.
\end{proof}

\subsection{Proof of \texorpdfstring{\cref{lem:aprox_pl_m}}{}}
\label{proof:lem:aprox_pl_m}

\begin{lemma}[\cref{lem:aprox_pl_m} Restated: Approximate 
$\partial_1 \ell(p_i(N),y_i)$]
    Let upper bounds $B_v, B_x, >0$ such that
    for any $k \in [K], j \in [N] ~\text{and}~ i \in [n]$, $\|v_{j_k}\|_2 \leq B_v$,
    and $\|x_i\|_2 \leq B_x$.
    For any $k \in [d]$, suppose function $u(t,y)[k]$ be $(\epsilon_{l}, R_3, M_3^k, C_3^k)$-approximable for $R_3 = \max
    \{B_v B_r, B_y, 1\}$, $M_3 \leq \Tilde{\mathcal{O}}((C_3^k)^2 \epsilon_l^{-2})$, where $C_3^k$ depends only on $R_3^k$ and the $C^3$-smoothness of $u(t,y)[k]$.
    Then, there exists an MLP layer 
    $ \rm{MLP}_{\theta_{N+2}}$ such that for any input sequences $h_i \in \R^D$ takes from \eqref{eqn:h_i_r_m}, 
    it maps
    \begin{align*}
        h_i = [x_i; y_i; w; \bar{p}_i; \bar{r}'_i; \mathbf{0}; 1; t_i] \xrightarrow
        {{\rm{MLP}}_{\theta_{N+2}}} 
        \Tilde{h_i} = [x_i; y_i; w; \bar{p}_i; \bar{r}'_i; g_i; \mathbf{0}; 1; t_i],
    \end{align*}
    where $g_i \in \R^d$ is an approximation for $u(p_i(N),y_i)$. 
    For any $k \in [d]$, assume $u(p_i(N),y_i)$ is $L_l$- Lipschitz continuous.  
    Then the approximation $g_i$ such that,
    \begin{align*}
        g_i[k] = u(p_i(N),y_i)[k] + \epsilon(i,k), \quad\text{with} \quad
        \abs{\epsilon(i,k)} \leq 
        \epsilon_l + L_l E_r \epsilon_r.
    \end{align*}
     Additionally, the parameters $\theta_{N+2}$ such that 
     $B_{\theta_{N+2}} \leq \max \{R_3 +1, C_3\}$.
\end{lemma}

\begin{proof}[Proof of \cref{lem:aprox_pl_m}]
    By our assumption and \cref{def:sum_of_relus_m}, for any $k \in [d]$, function $u[k](t,y)$ is $(\epsilon_{l}, R_3, M_3^k, C_3^k)$-approximable by sum of relus, there exists : 
    \begin{align}
        g_k(t,y) 
        = \sum_{m=1}^{M_3^k} c_m^{3,k} \sigma(\langle a_m^{3,k}, [t;y;1]\rangle) ~ \text{with}~
        \sum_{m=1}^{M_3^k} 
        \abs{c_m^{3,k}} \leq C_3, ~
        \|a_m^{3,k}\|_2 \leq 1, ~
        \forall m \in [M_3^k],
        \label{eqn:g_m}
    \end{align}
    such that $\sup_{(t,y) \in 
    [-R_3, R_3]^2} \abs{g_k(t,y) - u[k](t,y)} \leq \epsilon_l$.
    Then we construct our MLP layer.

    Let $M_3 := \sum_{k=1}^d M_3^k$, we pick matrices $W_1^{N+1} \in 
    \R^{M_3 \times D}, W_2^{N+1} \in
    \R^{D \times M_3}$ 
    such that for any $i \in [n+1], m \in [M_3]$,
    \begin{align}
        W_1^{N+1} h_i = & ~
        \begin{bmatrix}
            a_1^{3,1}[1] \cdot \bar{p}_i(N) + a_1^{3,1}[2] \cdot y_i + 
            a_1^{3,1}[3] - R_3(1-t_i) \\
            \vdots \\
            a_{M_3^1}^{3,1}[1] \cdot \bar{p}_i(N) + a_{M^1}^{3,1}[2] \cdot y_i + 
            a_{M_3^1}^{3,1}[3] - R_3(1-t_i) \\
            \vdots \\
            a_1^{3,d}[1] \cdot \bar{p}_i(N) + a_1^{3,d}[2] \cdot y_i + 
            a_1^{3,d}[3] - R_3(1-t_i) \\
            \vdots \\
            a_{M_3^d}^{3,d}[1] \cdot \bar{p}_i(N) + a_{M^d}^{3,d}[2] \cdot y_i + 
            a_{M_3^d}^{3,d}[3] - R_3(1-t_i)
        \end{bmatrix} \in \R^{M_3}, \nonumber \\
        W_2^{N+1}[j,m] = & ~ 
        c_m^{3,k} \cdot 1\{j = D_{g}^k, 
        M_3^{k-1} < m \leq M_3^k \}, 
        \label{eqn:matrix_con_3_m}
    \end{align}
    where $D_g^k$ denotes the position of element $g_i[k]$.
    Since input $h_i = [x_i; y_i; w; \bar{p}_i; \bar{r}'_i; \mathbf{0}; 1; t_i]$, similar to \eqref{eqn:matrix_form_m}, those matrices indeed exist.
    Furthermore, by \eqref{eqn:tf_norm}, they have operator norm bounds 
    \begin{align*}
        \|W_1^{N+1}\|_1
        \leq R_3 + 1, \quad
        \|W_2^{N+1}\|_1 \leq C_3
    \end{align*}
    Consequently, $B_{\theta_{N+2}} \leq 
    \max \{R_3 +1, C_3\}$.

    By our construction \eqref{eqn:matrix_con_3_m}, a simple calculation shows that
    \begin{align*}
        W_2^{N+1} \sigma(W_1^{N+1} h_i)
        = & ~ \sum_{k=1}^d \sum_{m=1}^{M_3^k} \sigma(\langle a_m^{3,k}, [\bar{p}_i(N);y_i;1]\rangle - 
        R_3(1-t_i)) \cdot c_m^{3,k} e_{D_g^k} \\
        = & ~ 1\{t_j=1\} \cdot 
        \begin{bmatrix}
            \mathbf{0} \\
            g_1(\bar{p}_i(N),y_i) \\
            \vdots \\
            g_d(\bar{p}_i(N),y_i) \\
            \mathbf{0}
        \end{bmatrix}.
    \end{align*}
    For $k \in [d]$, we let $g_i[k] = 
    1\{t_j=1\} \cdot g_k(\bar{p}_i(N),y_i) e_{D_g^k}$ for $i \in [n+1]$.
    Hence, $\rm{MLP}_{\theta_{N+2}}$ maps
    \begin{align*}
        h_i = [x_i; y_i; w; \bar{p}_i; \bar{r}'_i; \mathbf{0}; 1; t_i] \xrightarrow
        {\rm{MLP}_{\theta_{N+2}}} 
        \Tilde{h_i} = [x_i; y_i; w; \bar{p}_i; \bar{r}'_i; g_i; \mathbf{0}; 1; t_i],
    \end{align*}
    Next, we calculate the error generated in this approximation.
    By setting $R_3 = \max 
    \{B_v B_r, B_y, 1\}$, for any 
    $i \in [n+1]$, it holds
    \begin{align*}
        p_i(N) \leq R_3,
        \quad y_i \leq R_3
    \end{align*}
    Moreover, as our assumption, we suppose function $\partial_1 \ell$ is $L_l$-smooth in bounded domain $\mathcal{W}$.
    Therefore, by the definition of the function $g$, for each $i \in [n]$, the error becomes
    \begin{align*}
        & ~ \abs{g_i[k] - u(p_i(N),y_i)[k]} \\
        \leq & ~  
        \abs{g_i[k] - u(\bar{p}_i(N),y_i)[k]} + \abs{u(\bar{p}_i(N),y_i)[k] - u(p_i(N),y_i)[k]} 
        \annot{By triangle inequality} \\
        \leq & ~ \epsilon_l
        + L_l \|\bar{p}_i(N) -
        p_i(N)\|_2 \annot{By the definition of $g_k$ follows \eqref{eqn:g_m} and $L_l$-smooth assumption} \\
        \leq & ~ \epsilon_l +
        L_l E_r \epsilon_r,
        \annot{By the definition of $E_r$ follows \eqref{eqn:e_r_m}}.
    \end{align*}
    Combining above, we complete the proof.
\end{proof}

\subsection{Proof of \texorpdfstring{\cref{lem:aprox_s_j_m}}{}}
\label{proof:lem:aprox_s_j_m}

\begin{lemma}[\cref{lem:aprox_s_j_m} Restated: Approximate $\bar{s}_t(j)$]
    Recall that $s_i(j) = r'_i(j-1) \odot (V_{j+1}^{\top} \cdot s_i(j+1))$ follows \cref{def:s_m}. 
    Let the initial input takes from \eqref{eqn:h_i_g_m}.
    Then, there exist $N$ element-wise multiplication layers: ${\rm EWML}_{\theta_{N+3}}, \ldots, {\rm EWML}_{\theta_{2N+2}}$ such that for input sequences, 
    $j \in [N]$,
    \begin{align*}
        h_i = [x_i; y_i; w; \bar{p}_i; \bar{r}'_i; g_i; \bar{s}_i(N); \ldots; \bar{s}_i(j+1); \mathbf{0}; 1; t_i],
    \end{align*}
    they map ${\rm EWML}_{\theta_{2N+3-j}}(h_i)
        = [x_i; y_i; w; \bar{p}_i; \bar{r}'_i; g_i; \bar{s}_i(N); \ldots; \bar{s}_i(j); \mathbf{0}; 1; t_i]$,
    where the approximation $\bar{s}_i(j)$ is defined as recursive form: for any $i \in [n+1], j \in[N-1]$,
    \begin{align*}
        \bar{s}_i(j) := 
        \begin{cases}
            \bar{r}'_i(j-1) \odot (V_{j+1}^{\top} \cdot \bar{s}_i(j+1)), & j\in[N-1] \\
            \bar{r}'_i(N-1) \odot g_i, & j=N.
        \end{cases}
    \end{align*}
    Additionally, for any $j \in [N]$,  $B_{\theta_{N+2+j}}$ defined in \eqref{eqn:tf_norm} satisfies $B_{\theta_{N+2+j}} \leq 1$.
\end{lemma}

\begin{proof}[Proof of \cref{lem:aprox_s_j_m}]
By \cref{lem:aprox_p_m} and \cref{lem:aprox_r_p_m}, we obtain $\bar{p}_i(j)$ and $\bar{r}'_i(j)$, the approximation for $p_i(j)$ \eqref{eqn:p_m} and $r'_i(j)$ respectively.
Using $\bar{p}_i(j)$ and $\bar{r}'_i(j)$, we construct $N$ element-wise multiplication layers to approximate $s_i(j)$.
    
    We give the construction of parameters directly.
    For every $j \in [N-1], k \in [K]$, 
    we define matrices $Q_{k}^{2N+3-j},
    K_{k}^{2N+3-j},V_{k}^{2N+3-j} \in \R^{D \times D}$ such that for all $i \in [n+1]$,
    \begin{align}
        Q_{k}^{2N+3-j} h_i = \begin{bmatrix}
            v_{{j+1}_1}[k] \\
            \vdots \\
            v_{{j+1}_K}[k] \\
           \mathbf{0}
        \end{bmatrix}, \quad
        K_{k}^{2N+3-j} h_i = \begin{bmatrix}
            \bar{s}_i(j+1) \\
           \mathbf{0}
        \end{bmatrix}, \quad
        V_{k}^{2N+3-j} h_i =  
        \bar{r}'_i(j-1)[k] \cdot e_{j,k}^3~,
        \label{eqn:matrix_con_4_m}
    \end{align}
    where $e_{j,k}^3$ denotes the position unit vector of element $\bar{s}_i(j)[k]$.
    
    Since input $h_i = [x_i; y_i; w; \bar{p}_i; \bar{r}'_i; g_i; \bar{s}_i(N); \ldots; \bar{s}_i(j+1); \mathbf{0}; 1; t_i]$, similar to \eqref{eqn:matrix_form_m}, those matrices indeed exist.
    Thus, it is straightforward to check that
    \begin{align*}
        & ~ \sum_{k \in[K]} 
        \gamma(\langle 
        Q_{k}^{2N+3-j} h_i, 
        K_{k}^{2N+3-j} h_i \rangle) V_{k}^{2N+3-j} h_i 
        \\
        = & ~ 
        \sum_{k=1}^K  (V_{j+1}^{\top}[k,*] \cdot \bar{s}_i(j+1)) \bar{r}'_i(j-1)[k]
        e_{j,k}^3 
        \annot{By definition of EWML layer follows \cref{def:ewml_m}} \\
        = & ~ 
        \begin{bmatrix}
            \mathbf{0} \\
            r_i(j-1)[1] V_{j+1}^{\top}[1,*] \cdot \bar{s}_i(j+1) \\
            \vdots \\
            \bar{r}'_i(j-1)[k] V_{j+1}^{\top}[K,*] \cdot \bar{s}_i(j+1) \\
            \mathbf{0}
        \end{bmatrix} 
        \annot{By definition of $e_{j,k}^3$}\\
        = & ~ 
        \begin{bmatrix}
            \mathbf{0} \\
            \bar{r}'_i(j-1) \odot (V_{j+1}^{\top} \cdot \bar{s}_i(j+1)) \\
            \mathbf{0}
        \end{bmatrix} 
        \annot{By definition of hadamard product} \\
        = & ~  [\mathbf{0}
        ;\bar{s}_i(j);\mathbf{0}].
        \annot{By definition of $\bar{s}_i(j)$ follows \eqref{eqn:s_bar_i(j)_m}}
    \end{align*}
    Therefore, by the definition of EWML layer follows \cref{def:ewml_m}, the output
    $\Tilde{h}_i$ becomes
    \begin{align*}
        \Tilde{h}_i = & ~
        [{\rm Attn}_{\theta_{2N+3-j}}(h_i)] \\
        = & ~ h_i + \sum_{m \in[2], k \in[K]} \sigma(\langle 
        Q_{m,k}^{2N+3-j} h_i, K_{m,k}^{2N+3-j} h_s \rangle) V_{m,k}^{2N+3-j} h_s \\
        = & ~ h_i +   [\mathbf{0};s(j);\mathbf{0}]\\
        = & ~ [x_i; y_i; w; \bar{p}_i; \bar{r}'_i; g_i; \bar{s}_i(N-1); \ldots; \bar{s}_i(j+1); \mathbf{0}; 1; t_i] +
        [\mathbf{0};
        \bar{s}_i(j);\mathbf{0}]
        \\
        = & ~ [x_i; y_i; w; \bar{p}_i; \bar{r}'_i; g_i; \bar{s}_i(N-1); \ldots; \bar{s}_i(j); \mathbf{0}; 1; t_i].
    \end{align*}
    Finally we come back to approximate the initial approximation $\bar{s}_i(N)=\bar{r}'_i(N-1) \odot g_i$.
    Notice that $g_i$ and 
    $r'_i(N-1)$ are already in the input $h_i = [x_i; y_i; w; \bar{p}_i; \bar{r}'_i; g_i; \mathbf{0}; 1; t_i]$, thus it is simple to construct ${\rm EWML}_{N+3}$ , similar to \eqref{eqn:matrix_con_4_m}, such that it maps,
    \begin{align*}
        [x_i; y_i; w; \bar{p}_i; \bar{r}'_i; g_i; \mathbf{0}; 1; t_i]
        \xrightarrow{{\rm EWML}_{N+3}}
        [x_i; y_i; w;  \mathbf{0}; 1; \bar{p}_i; \bar{r}'_i; g_i; \bar{s}_i(N); \mathbf{0}; 1; t_i].
    \end{align*}
    
    Since we don't using the sum of ReLU to approximate any variables, these step don't generate extra error.
    Besides, by \eqref{eqn:ewml_norm_m}, matrices have operator norm bounds 
    \begin{align*}
        \max_{j,k} \|Q_{k}^{N+2+j}\|
        _1 \leq  1, \quad
        \max_{j,k} \|K_{k}^{N+2+j}\|
        _1 \leq  1, \quad
        \max_{j,k} \|V_{k}^{N+2+j}\|
        _1 \leq  1 .
    \end{align*}
    Consequently, for any $j \in [N]$, $B_{\theta_{N+2+j}} \leq 1$.
    Thus we complete the proof.
\end{proof}

\subsection{Proof of \texorpdfstring{\cref{lem:error_g_s_m}}{}}
\label{proof:lem:error_g_s_m}

\begin{lemma}[\cref{lem:error_g_s_m} Restated: Error for $g_i \bar{s}_i(j)$]
    Suppose the upper bounds $B_v, B_x> 0$ such that
    for any $k \in [K], j \in [N] ~\text{and}~ i \in [n]$, $\|v_{j_k}\|_2 \leq B_v$,
    and $\|x_i\|_2 \leq B_x$.
    Let $r'_i(j) \in \R^K$ such that $r'_i(j)[k] := 
    r'(v_{{j+1}_k}^\top p_i(j))$ follows \cref{def:abbv_m}.
    Let $s_i(j) = R_i(j-1) V_{j+1}^{\top} \ldots R_i(N-2) V_{N}^{\top} \cdot R_i(N-1) u$ follows \cref{def:s_m}.
    Let $\bar{r}'_i(j), g_i, \bar{s}_i(j)$ be the approximations for $r'_i(j), u(p_i(N),y_i), s_i(j)$  follows
    \cref{lem:aprox_r_p_m},
    \cref{lem:aprox_pl_m} and \cref{lem:aprox_s_j_m} respectively.
    Let $B_{r'}$ be the upper bound of $\bar{r}'_i(j)[k]$ and $r'_i(j)[k]$ as defined in \cref{lem:aprox_r_p_m}.
    Let $B_l$ be the upper bound of $g_i[k]$ and $u(p_i(N),y_i)[k]$ as defined in \cref{lem:aprox_pl_m}.
    Then for any $i \in [n+1], j \in [N], k \in [K]$, 
    \begin{align*}
        & ~ \bar{s}_i(j)[k] 
        \leq  B_s, \\
        & ~  \abs{\bar{s}_i(j)[k] -  s_i(j)[k]}  
        \leq E_s^{r} \epsilon_r + E_s^{r'} \epsilon_{r'} + E_s^{l} \epsilon_l, 
    \end{align*}
    where  
    \begin{align*}
        P := & ~ \max \{\sqrt{K},\sqrt{d}\} \\
        B_s := & ~ 
        \max_{j \in [N]} \{ (P \cdot B_{r'} B_v)^{N-j} 
        B_{r'} B_l\}, \\
        E_s^r := & ~ 
        \max_{j \in [N]} \{L_{r'} E_r P B_s B_v^2 [\sum_{l=0}^{N-j-1} (B_{r'} B_v P)^l] + (B_{r'} B_v P)^{N-j} (B_l L_{r'} B_v E_r + B_{r'} L_l E_r)\}, \nonumber \\
        E_s^{r'} := & ~ 
        \max_{j \in [N]} \{P B_s B_v [\sum_{l=0}^{N-j-1} (B_{r'} B_v P)^l] + (B_{r'} B_v P)^{N-j} B_l \}, \nonumber \\
        E_s^l := & ~ 
        \max_{j \in [N]} 
        \{(B_{r'} B_v P)^{N-j} B_{r'}\}.
    \end{align*} 
    Above, $B_s$ is the upper bound of $\bar{s}_i(j)[k]$ and $E_s^r, E_s^{r'}, E_s^l$ are the
    coefficients of $\epsilon_r, \epsilon_{r}', \epsilon_l$ in the upper bounds of $\abs{\bar{s}_i(j)[k] - s_i(j)[k]}$, respectively.
\end{lemma}

\begin{proof}[Proof of \cref{lem:error_g_s_m}]
    By \cref{lem:aprox_s_j_m}, we manage to approximate $s_i(j)$ by $\bar{s}_i(j)$.
    By triangle inequality, we have
    \begin{align*}
        \abs{\bar{s}_i(j)[k] - s_i(j)[k]} \leq \abs{\bar{r}'_i(n-1)[k] - r'_i(n-1)[k]} \cdot \abs{v_{{n+1}_k}^{\top} 
    \bar{s}_i(n+1)} + \abs{r'_i(n-1)[k]} \cdot \abs{(v_{{n+1}_k}^{\top} \bar{s}_i(n+1)) -(v_{{n+1}_k}^{\top} s_i(n+1))}.
    \end{align*}
    We bound these four terms separately.
    By \cref{lem:aprox_r_p_m}, $\abs{\bar{r}'_i(n-1)[k] - r'_i(n-1)[k]}$ is bounded by $\epsilon_{r'} +
    L_{r'} B_v E_r \epsilon_r$.
    We then use induction to establish upper bounds for $\bar{s}_i(j)[k]$ and $\abs{\bar{s}_i(j)[k] - s_i(j)[k]}$.

    We first use induction to prove the first two statements.
    To begin with, we illustrate the recursion formula for $\bar{s}_i(j)$.
    By \eqref{eqn:s_bar_i(j)_m}, recall that for any $j \in [N]$, 
    \begin{align*}
        \bar{s}_i(j) 
        := \begin{cases}
            \bar{r}'_i(j-1) \odot (V_{j+1}^{\top} \cdot \bar{s}_i(j+1)), & j\in[N-1] \\
            \bar{r}'_i(N-1) \odot g_i, & j=N.
        \end{cases}
    \end{align*}
    We consider applying induction to prove the first statement:
    \begin{align*}
        \bar{s}_i(j)[k] \leq 
        (P \cdot B_{r'} B_v)^{N-n} 
        B_{r'} B_l.
    \end{align*}
    As for the base case, $j = N$:
    \begin{align*}
        \bar{s}_i(N)[k] = \bar{r}'_i(N-1)[k] \cdot g_i[k] \leq B_{r'} B_l.
    \end{align*}
    Therefore, if the statement holds for $j = n+1$, by \eqref{eqn:s_bar_i(j)_m} and our assumption, it holds
    \begin{align*}
        \bar{s}_i(n)[k] 
        = & ~ \bar{r}'_i(n-1)[k] \cdot (v_{{j+1}_k}^{\top} \bar{s}_i(n+1)) \annot{By recursion formula \eqref{eqn:s_bar_i(j)_m}} \\
        \leq & ~ \bar{r}'_i(n-1)[k] \cdot \|v_{{n+1}_k}\|_2 \cdot \|\bar{s}_i(n+1)\|_2 \annot{By Cauchy-schwarz inequality} \\
        \leq & ~ \bar{r}'_i(n-1)[k] \cdot \|v_{{n+1}_k}\|_2 \cdot \max 
        \{\sqrt{K},\sqrt{d}\} \cdot \max_{k} \abs{\bar{s}_i(n+1)[k]} \\
        \leq & ~ (B_{r'} B_{v}) \cdot \max \{\sqrt{K},\sqrt{d}\} \cdot (\max \{\sqrt{K},\sqrt{d}\} \cdot 
        B_{r'} B_v)^{N-n-1} 
        B_{r'} B_l \annot{By inductive hypothesis}\\
        = & ~ (P \cdot B_{r'} B_v)^{N-n} 
        B_{r'} B_l \annot{By definition of $P$ follows \cref{lem:error_g_s_m}}.
    \end{align*}
    Thus, by the principle of induction, the first statement is true for all integers $j \in [N]$. 
    Moreover, by the definition of $B_s$ follows\cref{lem:error_g_s_m}, we know $B_s$ is the upper bound of $\bar{s}_i(j)[k]$.
    Next we apply induction to prove the second statement:
    \begin{align*}
        \abs{\bar{s}_i(j)[k] - s_i(j)[k]}  
        \leq & ~ (\epsilon_{r'} +
        L_{r'} B_v E_r \epsilon_r) P B_v B_s [\sum_{l=0}^{N-n-1} (B_{r'} B_v P)^l] \\ 
        & ~ + (B_{r'} B_v P)^{N-n} [(B_l L_{r'} B_v E_r + B_{r'} L_l E_r) \epsilon_r + B_l \epsilon_{r'} + B_{r'} \epsilon_l].
    \end{align*}
    For the base case, $j = N$: 
    \begin{align*}
        & ~ \abs{\bar{s}_i(N)[k] - 
        s_i(N)[k]} \\
        = & ~ \abs{\bar{r}'_i(N-1)[k] \cdot g_i[k] - r'_i(N-1)[k] \cdot u(p_i(N),y_i)[k]} \annot{By definition \eqref{eqn:s_bar_i(j)_m} and \eqref{eqn:s_recursion_m}}\\
        \leq & ~ \abs{\bar{r}'_i(N-1)[k] - r'_i(N-1)[k]} \cdot \abs{g_i[k]} + \abs{r'_i(N-1)[k]} \cdot \abs{g_i[k]-u(p_i(N),y_i)[k]} \annot{By triangle inequality} \\
        \leq & ~  (\epsilon_{r'} +
        L_{r'} B_v E_r \epsilon_r) B_l + 
        B_{r'} (\epsilon_l + L_l E_r \epsilon_r).
        \annot{By \eqref{eqn:error_2_m} and \eqref{eqn:error_3_m}} \\
        = & ~ (B_l L_{r'} B_v E_r + B_{r'} L_l E_r) \epsilon_r + B_l \epsilon_{r'} + B_{r'} \epsilon_l
    \end{align*}
    Therefore, if the statement holds for $j = n+1$, by \eqref{eqn:s_bar_i(j)_m} and our assumption, it holds
    \begin{align*}
        & ~ \abs{\bar{s}_i(n)[k] - 
        s_i(n)[k]} \\
        = & ~ \abs{\bar{r}'_i(n-1)[k] \cdot (v_{{n+1}_k}^{\top} \bar{s}_i(n+1)) - r'_i(n-1)[k] \cdot (v_{{n+1}_k}^{\top} s_i(n+1))} \annot{By the recursion formula \eqref{eqn:s_recursion_m} and \eqref{eqn:s_bar_i(j)_m}}\\
        \leq & ~ \abs{\bar{r}'_i(n-1)[k] - r'_i(n-1)[k]} \cdot \abs{v_{{n+1}_k}^{\top} 
         \bar{s}_i(n+1)} \\
         & ~ + \abs{r'_i(n-1)[k]} \cdot \abs{(v_{{n+1}_k}^{\top} \bar{s}_i(n+1)) -(v_{{n+1}_k}^{\top} s_i(n+1))} \annot{By triangle inequality}\\
        \leq & ~ 
        (\epsilon_{r'} +
        L_{r'} B_v E_r \epsilon_r) P B_v B_s + B_{r'} B_v \|\bar{s}_i(n+1) - s_i(n+1)\|_2 
        \annot{By error accumulation of approximating $r'$ follows \eqref{eqn:error_2_m} } \\
        \leq & ~ 
        (\epsilon_{r'} +
        L_{r'} B_v E_r \epsilon_r) P B_v B_s + B_{r'} B_v P \max_k \abs{\bar{s}_i(n+1)[k] - s_i(n+1)[k]} \\
        \leq & ~ 
        (\epsilon_{r'} +
        L_{r'} B_v E_r \epsilon_r) P B_v B_s  + B_{r'} B_v P 
        \Big\{(\epsilon_{r'} +
        L_{r'} B_v E_r \epsilon_r) P B_v B_s [\sum_{l=0}^{N-n-2} (B_{r'} B_v P)^l] \\
        & ~ + (B_{r'} B_v P)^{N-n-1} [(B_l L_{r'} B_v E_r + B_{r'} L_l E_r) \epsilon_r + B_l \epsilon_{r'} + B_{r'} \epsilon_l] \Big\} \annot{By inductive hypothesis}\\
        \leq & ~ 
        (\epsilon_{r'} +
        L_{r'} B_v E_r \epsilon_r) P B_v B_s [\sum_{l=0}^{N-n-1} (B_{r'} B_v P)^l] \\ 
        & ~ + (B_{r'} B_v P)^{N-n} [(B_l L_{r'} B_v E_r + B_{r'} L_l E_r) \epsilon_r + B_l \epsilon_{r'} + B_{r'} \epsilon_l].
    \end{align*}
    Thus, by the principle of induction, the second statement is true for all integers $j \in [N-1]$. 
    By the definition of $E_s$ follows 
    \cref{lem:error_g_s_m}, 
    it is simple to check that
    \begin{align*}
        \abs{\bar{s}_i(j)[k] -  s_i(j)[k]}  
        \leq E_s^{r} \epsilon_r + E_s^{r'} \epsilon_{r'} + E_s^{l} \epsilon_l.
    \end{align*}
    Thus we complete the proof.
\end{proof}

\subsection{Proof of \texorpdfstring{\cref{thm:icl_gd_m}}{}}
\label{proof:thm:icl_gd_m}

\begin{theorem}[\cref{thm:icl_gd_m} Restated: In-context gradient descent on $N$-layer NNs]
    Fix any $B_v, \eta, \epsilon > 0, L \geq 1$.
    For any input sequences takes from $\eqref{eqn:input}$, their exist upper bounds $B_x,B_y$ such that for any $i \in [n]$, $\|y_i\|_2 \leq B_y$, $\|x_i\|_2 \leq B_x$.
    Assume functions $r(t)$, $r'(t)$ and $u(t,y)[k]$ are $L_r,L_{r'},L_l$-Lipschitz continuous.
    Suppose $\mathcal{W}$ is a closed domain such that for any $j \in [N-1]$ and $k \in [K]$,
    \begin{align*}
        \mathcal{W} \subset \left\{
        w = [v_{j_k}] \in \mathbb{R}^{D_N} : \|v_{j_k}\|_2 \leq B_v
        \right\},      
    \end{align*}
    and ${\rm Proj}_{\mathcal{W}}$ project $w$ into bounded domain $\mathcal{W}$. 
    Assume
    ${\rm Proj}_{\mathcal{W}} = {\rm MLP}_{\theta}$ for some MLP layer with hidden dimension 
    $D_w$ parameters $\|\theta\| \leq C_{w}$.
    If functions $r(t)$, $r'(t)$ and $u(t,y)[k]$ are $C^4$-smoothness, then for any $\epsilon > 0$, there exists a transformer model ${\rm NN}_{\theta}$ with $(2N+4)L$ hidden layers
    consists of $L$ neural network blocks 
    ${\rm TF}_{\theta}^{N+2} \circ
    {\rm EWML}_{\theta}^{N} \circ
    {\rm TF}_{\theta}^{2}$,
    \begin{align*}
        {\rm NN}_{\theta} := 
        {\rm TF}_{\theta}^{N+2} \circ
        {\rm EWML}_{\theta}^{N} \circ
        {\rm TF}_{\theta}^{2}
        \circ \ldots \circ 
        {\rm TF}_{\theta}^{N+2} \circ
        {\rm EWML}_{\theta}^{N} \circ
        {\rm TF}_{\theta}^{2},
    \end{align*}
    such that the heads number $M^l$, embedding dimensions $D^l$, and the parameter norms
    $B_{\theta^l}$ suffice
    \begin{align*}
        \max_{l \in [(2N+4)L]} M^l \leq \Tilde{O}(\epsilon^{-2}), \quad
        \max_{l \in [(2N+4)L]} D^l \leq O(NK^2) + D_w, \quad
        \max_{l \in [(2N+4)L]} B_{\theta^l} \leq 
        O(\eta) + C_w + 1,
    \end{align*}
    where $\Tilde{O}(\cdot)$ hides the constants that depend on $d,K,N$, the radius parameters $B_x,B_y,B_v$ and the smoothness
    of $r$ and $\ell$. 
    And this neural network such that for any input sequences $H^{(0)}$, take from \eqref{eqn:input}, ${\rm NN_{\theta}}(H^{(0)})$ implements $L$ steps in-context gradient descent on risk Eqn~\eqref{eqn:loss}: For every $l \in [L]$, the $(2N+4)l$-th layer outputs $h_i^{((2N+4)l)} = [x_i; y_i; \bar{w}^{(l)}; \mathbf{0}; 1; t_i]$ for every $i \in [n+1]$, and approximation gradients
    $\bar{w}^{(l)}$ such that 
    \begin{align*}
        \bar{w}^{(l)} = 
        {\rm Proj}_{\mathcal{W}} 
        (\bar{w}^{(l-1)} - \eta \nabla \mathcal{L}_n(\bar{w}^{(l-1)}) + \epsilon^{(l-1)}), \quad \bar{w}^{(0)} = \mathbf{0},
    \end{align*}
    where $\|\epsilon^{(l-1)}\|_2 \leq \eta \epsilon$ is an error term.
\end{theorem}

\begin{proof}[Proof Sketch]
     Let the first $2N+2$ layers of ${\rm NN}_{\theta}$ are Transformers and EWMLs constructed in \cref{lem:aprox_p_m}, 
    \cref{lem:aprox_r_p_m},
    \cref{lem:aprox_pl_m}, and
    \cref{lem:aprox_s_j_m}.
    Explicitly, we design the last two layers to implement the gradient descent step (\cref{lem:decomposition_gd_m}).
    We then establish the upper bounds for error $\|\nabla_w \bar{\mathcal{L}}_n(w) - \nabla_w \mathcal{L}_n(w)\|_2$, where $\nabla_w \bar{\mathcal{L}}_n(w)$, derived from the outputs of ${\rm NN}_{\theta}$, approximates
    $\nabla_w \mathcal{L}_n(w)$.
    Next, for any $\epsilon>0$, we select appropriate parameters $\epsilon_l$, $\epsilon_r$ and $\epsilon_{r'}$ to ensure that  $\|\nabla_w \bar{\mathcal{L}}_n(\bar{w}^{(l-1)}) - \nabla_w \mathcal{L}_n(\bar{w}^{(l-1)})\|_2 
    \leq \epsilon$ holds for any $l \in [L]$.
\end{proof}

\begin{proof}[Proof of \cref{thm:icl_gd_m}]
    We consider the first $N+2$ transformer layers ${\rm TF}_{\theta}^{N+2}$ are layers in \cref{lem:aprox_p_m} ,\cref{lem:aprox_r_p_m} and \cref{lem:aprox_pl_m}.
    Then we let the middle $N$ element-wise multiplication layers ${\rm EWML}_{\theta}^{N}$ be layers
    in \cref{lem:aprox_s_j_m}.
    We only need to check
    approximability conditions.
    By \cref{lem:approx_k_v} and our assumptions, for any $\epsilon_r,\epsilon_{r'},\epsilon_l$, it holds
    \begin{itemize}
        \item Function $r(t)$ is $(\epsilon_r, R_1, M_1, C_1)$-approximable for $R_1 = \max \{B_v B_r, 1\}$, $M_1 \leq \Tilde{\mathcal{O}}(C_1^2 \epsilon_r^{-2})$, where $C_1$ depends only on $R_1$ and the $C^2$-smoothness of $r(t)$.
        \item Function $r'(t)$ is $(\epsilon_{r'}, R_2, M_2, C_2)$-approximable for $R_2 = \max \{B_v B_{r'}, 1\}$, $M_2 \leq \Tilde{\mathcal{O}}(C_2^2 \epsilon_r'^{-2})$, where $C_2$ depends only on $R_2$ and the $C^2$-smoothness of $r'(t)$.
        \item Function $\partial_1 \ell(t,y)$ is $(\epsilon_{l}, R_3, M_3, C_3)$-approximable for $R_3 = \max
        \{B_v B_r, 1\}$, $M_3 \leq \Tilde{\mathcal{O}}(C_3^2 \epsilon_l^{-2})$, where $C_3$ depends only on $R_3$ and the $C^3$-smoothness of $u(t,y)[k]$.
    \end{itemize}
    which suffice approximability conditions in \cref{lem:aprox_p_m}, 
    \cref{lem:aprox_r_p_m} and \cref{lem:aprox_pl_m}.
    
    Now we construct the last two layers to implement $w - \eta \nabla \mathcal{L}_n(w)$ and ${\rm Proj}_{\mathcal{W}}(w)$.
    First we construct a attention layer to approximate $w - \eta \nabla \mathcal{L}_n(w)$.
    For every $m \in [2], j \in [N], k \in [K]$, we consider matrices $Q_{m,j,k}^{2N+3},j_{m,j,k}^{2N+3},
    V_{m,j,k}^{2N+3} \in \R^{D \times D}$ such that
    \begin{align}
        Q_{1,j,k}^{2N+3} h_i = \begin{bmatrix}
            1 \\
           \mathbf{0}
        \end{bmatrix}, \quad
        K_{1,j,k}^{2N+3} h_i = \begin{bmatrix}
            \bar{s}_i(j)[k] \\
           \mathbf{0}
        \end{bmatrix}, \quad
        V_{1,j,k}^{2N+3} h_i =  
        -\frac{\eta (n+1)}{2n}
        \begin{bmatrix}
            \mathbf{0}\\
            \bar{p}_i(j-1) \\
            \mathbf{0}
        \end{bmatrix}, \nonumber \\
         Q_{2,j,k}^{2N+3} h_i = \begin{bmatrix}
            -1 \\
           \mathbf{0}
        \end{bmatrix}, \quad
        K_{2,j,k}^{2N+3} h_i = \begin{bmatrix}
            \bar{s}_i(j)[k] \\
           \mathbf{0}
        \end{bmatrix}, \quad
        V_{2,j,k}^{2N+3} h_i = 
        -\frac{\eta (n+1)}{2n}
        \begin{bmatrix}
            \mathbf{0}\\
            - \bar{p}_i(j-1) \\
            \mathbf{0}
        \end{bmatrix}.
        \label{eqn:matrix_con_5_m}
    \end{align} 
    Furthermore, we define  approximation gradient $\nabla_w \bar{\mathcal{L}}_n(w)$ as follows,
    \begin{align}
        \nabla_w \bar{\mathcal{L}}_n(w) 
        := & ~ -\frac{1}{\eta(n+1)}
        \sum_{t=1}^{n+1} 
        \sum_{m \in[2],
        j \in [N], k \in[K]} \sigma(\langle 
        Q_{m,j,k}^{2N+3} h_i, K_{m,j,k}^{2N+3} h_t \rangle) V_{m,j,k}^{2N+3} h_t \nonumber \\ 
        = & ~ \frac{1}{2n}
        \sum_{t=1}^{n+1}
        \sum_{k=1}^K  \sum_{j=1}^{N} (\sigma( \bar{s}_t(j)[k]) - 
        \sigma(- \bar{s}_t(j)[k])) \begin{bmatrix}
            \mathbf{0}\\
            \bar{p}_t(j-1) \\
            \mathbf{0}
        \end{bmatrix} \nonumber \annot{By our construction
        \eqref{eqn:matrix_con_5_m}}\\
        = & ~ \frac{1}{2n} 
        \sum_{t=1}^{n+1}
        \sum_{k=1}^K  \sum_{j=1}^{N} 
        \bar{s}_t(j)[k] \cdot
        \begin{bmatrix}
            \mathbf{0}\\
            \bar{p}_t(j-1) \\
            \mathbf{0}
        \end{bmatrix} \nonumber \annot{By $f(x) = \sigma(x) - \sigma(-x)$}\\ 
        = & ~ \frac{1}{2n} 
        \sum_{t=1}^{n+1} \sum_{j=1}^{N} 
        \begin{bmatrix}
        \mathbf{0} \\
        \mathbf{I}_{K \times K} \otimes 
        \bar{p}_t(j-1) \cdot \bar{s}_t(j) \\
        \mathbf{0}
        \end{bmatrix} \nonumber \annot{By definition of Kronecker product}\\ 
        = & ~  \frac{1}{2n} 
        \sum_{t=1}^{n}
        \begin{bmatrix}
            \mathbf{0} \\
            \bar{A}_t(1) \\
            \vdots \\
            \bar{A}_t(N) \\
            \mathbf{0}
        \end{bmatrix},
        \annot{By $\bar{s}_{n+1}(j)=\mathbf{0}$ follows \cref{lem:aprox_s_j_m}}
    \end{align}
    where $\bar{A}_t(j) := \mathbf{I}_{K \times K} \otimes 
    \bar{p}_t(j-1) \cdot \bar{s}_t(j)$ denotes the approximation for $A_t(j)$.
    Therefore, by the definition of ReLU attention layer follows \cref{def:attn}, for any 
    $i \in [n+1]$,
    \begin{align*}
        \Tilde{h}_i = & ~
        [{\rm Attn}_{\theta_{2N+3}}(h_i)] \\
        = & ~ h_i + \frac{1}{n+1} \sum_{i=1}^{n+1} \sum_{m \in[2],
        j \in [N], k \in[K]} \sigma(\langle 
        Q_{m,j,k}^{2N+3} h_s, K_{m,j,k}^{2N+3} h_i \rangle) V_{m,j,k}^{2N+3} h_i \\
        = & ~ [x_i; y_i; w; \bar{p}_i; \bar{r}'_i; g_i;\bar{s}_i; 
        \mathbf{0}; 1; t_i] 
        - \frac{\eta}{2n} \sum_{t=1}^{n}
        \begin{bmatrix}
            \mathbf{0} \\
            \bar{A}_t(1) \\
            \vdots \\
            \bar{A}_t(N) \\
            \mathbf{0}
        \end{bmatrix} \\
        = & ~ [x_i; y_i;w - \eta \nabla_w \bar{\mathcal{L}}_n(w) ; \bar{p}_i; \bar{r}'_i; g_i; \mathbf{0}; 1; t_i].
        \annot{By definition of $\nabla_w \bar{\mathcal{L}}_n(w)$}
    \end{align*}
    Since we do not use approximation technique like \cref{def:sum_of_relus_m}, this step do not generate extra error.
    Besides,by \eqref{eqn:tf_norm}, matrices have operator norm bounds 
    \begin{align*}
        \max_{j,m,k} \|Q_{j,m,k}^{2N+3}\|
        _1 \leq  1, \quad
        \max_{j,m,k} \|K_{j,m,k}^{2N+3}\|
        _1 \leq  1, \quad
        \sum_{j,m,k} \|V_{j,m,k}^{2N+3}\|
        _1 \leq  2 \eta NK.
    \end{align*}
    Consequently, $B_{\theta_{2N+3}}
    \leq 1 + 2 \eta NK $.
    Fix any $\epsilon>0$, then we pick appropriate $\epsilon_r, \epsilon_r', \epsilon_l$ such that 
    \begin{align*}
        \|\epsilon^{(l-1)}\|_2 
        = \eta \|\nabla_w \bar{\mathcal{L}}_n(\bar{w}^{(l-1)}) - \nabla_w \mathcal{L}_n(\bar{w}^{(l-1)})\|_2 
        \leq \eta \epsilon.
    \end{align*}
    By \cref{def:s_m} and \cref{lem:error_g_s_m}, for any $j \in [N-1], i \in [n]$, it holds
    \begin{align*}
        & ~ \|\bar{A}_i(j) - A_i(j)\|_2 \\
        \leq & ~ \sum_{k=1}^{K}
        \|\bar{s}_i(j)[k] 
        \bar{p}_i(j-1) - s_i(j)[k] 
        p_i(j-1)\|_2 \annot{By \cref{def:abbv_m} and definition of $\bar{A}_i(j)$}\\
        \leq & ~ \sum_{k=1}^{K}
        \abs{\bar{s}_i(j)[k] - s_i(j)[k]} \cdot \|\bar{p}_i(j-1)\|_2 +
        \abs{s_i(j)[k]} \cdot 
        \|\bar{p}_i(j-1) - p_i(j-1)\|_2 \annot{By triangle inequality}\\
        \leq & ~ P [(E_s^{r} \epsilon_r + E_s^{r'} \epsilon_{r'} + E_s^{l} \epsilon_l) \sqrt{P} B_r + B_s E_r \epsilon_r],
        \annot{By \eqref{eqn:e_r_m} and \cref{lem:error_g_s_m} }
    \end{align*}
    where $B_s$ is the upper bound of $\bar{s}_i(j)[k]$ and $E_s^r, E_s^{r'}, E_s^l$ are the
    coefficients of $\epsilon_r, \epsilon_{r}', \epsilon_l$ in the upper bounds of $\abs{\bar{s}_i(j)[k] - s_i(j)[k]}$ follow \cref{lem:error_g_s_m}, respectively.
    We can drive similar results as $j=N$.
    Actually, by $P=\max\{\sqrt{K},\sqrt{d}\}$ follows \cref{lem:error_g_s_m}, above inequality also holds for $j=N$.
    Therefore, the error in total such that for any $w$,
    \begin{align*}
        & ~ 
        \|\nabla_w \bar{\mathcal{L}}_n(w) - 
        \nabla_w \mathcal{L}_n(w)\|_2 \\
        = & ~ \|\frac{1}{2n} \sum_{t=1}^{n}
        \begin{bmatrix}
            \mathbf{0} \\
            \bar{A}_t(1) \\
            \vdots \\
            \bar{A}_t(N) \\
            \mathbf{0}
        \end{bmatrix} -\frac{1}{2n} \sum_{t=1}^{n}
        \begin{bmatrix}
            \mathbf{0} \\
            A_t(1) \\
            \vdots \\
            A_t(N) \\
            \mathbf{0}
        \end{bmatrix} \|_2 \annot{By definition of $\bar{\mathcal{L}}_n(w)$ and $\mathcal{L}_n(w)$} \\
        \leq & ~ \frac{1}{2}
        \max_{1 \leq t \leq n}
        \{\sum_{j=1}^{N} 
        \|\bar{A}_t(j) - A_t(j)\|_2\} \\
        \leq & ~ \frac{N}{2}
         P [(E_s^{r} \epsilon_r + E_s^{r'} \epsilon_{r'} + E_s^{l} \epsilon_l) \sqrt{P} B_r + B_s E_r \epsilon_r].
        \annot{By the error accumulation results derived before}
    \end{align*}
    Let $C_l,C_r,C_{r'}$ denotes coefficients in front of $\epsilon_l,\epsilon_r,
    \epsilon_{r'}$ respectively.
    Then it holds
    \begin{align*}
        C_l = & ~ N P^{\frac{3}{2}} B_r E_s^l, \\
        C_r = & ~ N P^{\frac{3}{2}} B_r E_s^r + N P B_s E_r, \\
        C_{r'} = & ~
        N P^{\frac{3}{2}} B_r E_s^{r'}.
    \end{align*}
    Thus, to ensure $ \|\nabla_w \bar{\mathcal{L}}_n(w) - \nabla_w \mathcal{L}_n(w)\|_2 \leq \epsilon$, we only need to select $\epsilon_l, \epsilon_r, \epsilon_{r}'$ as
    \begin{align*}
        \epsilon_l = \frac{2\epsilon}
        {3 C_l}, \quad 
        \epsilon_r = \frac{2\epsilon}
        {3 C_r},
        \quad
        \epsilon_{r}' = \frac{2\epsilon}
        {3 C_{r'}}.
    \end{align*}

    Therefore, we only need to pick the last MLP layer ${\rm MLP}_{2N+4}$ such that it maps
    \begin{align*}
        [x_i; y_i; w - \eta \nabla_w \mathcal{L}_n(w) ; \bar{p}_i; \bar{r}'_i; g_i; \bar{s}_i; \mathbf{0}; 1; t_i]
        \xrightarrow{{\rm MLP}_{2N+4}}
        [x_i; y_i; {\rm Proj}_{\mathcal{W}}(w - \eta \nabla_w \mathcal{L}_n(w));  \mathbf{0}; 1; t_i].
    \end{align*}
    By our assumption on the map ${\rm Proj}_{\mathcal{W}}$, this is easy.
    
    Finally, we analyze how many embedding dimensions of Transformers are needed to implement the above ICGD.
    Recall that 
    \begin{align*}
        x_i, y_i \in \R^d, 
        w \in \R^{2dK + (N-2)K^2},
        \bar{p}_i \in  
        \R^{(N-1)K+d},
        \bar{r}'_i \in  \R^{(N-2)K+d},
        g_i \in \R^d,
        \bar{s}_i \in \R^{(N-1)K+d}.
    \end{align*}
    Therefore,  $\max 
    \{\Omega(NK^2),D_w\}$ embedding dimensions of Transformer are required to implement ICGD on deep models.
    
    Combining the above, we complete the proof.
\end{proof}

\begin{remark}[Modest Assumptions]
    Our assumptions remain modest. For example, we require that the loss function $l(\cdot)$, the activation function $r(\cdot)$, and its derivative $r^{'}(\cdot)$ are $C^4$-smoothness. 
    Many settings meet these conditions, including those using the sigmoid activation function as $r(\cdot)$ and the squared loss function.
\end{remark}

\subsection{Proof of \texorpdfstring{\cref{coro:error_ICGD_m}}{}}
\label{proof:coro:error_ICGD_m}

\begin{corollary}[\cref{coro:error_ICGD_m} Restated: Error for implementing ICGD on $N$-layer neural network]
    Fix $L \geq 1$, under the same setting as \cref{thm:icl_gd_m}, $(2N+4)L$-layer neural networks
    ${\rm NN}_{\theta}$ approximates
    the true gradient descent trajectory 
    $\{w^l_{\rm GD}\}_{l\geq0} \in \R^{D_N}$ with the error accumulation
    \begin{align*}
        \|\bar{w}^{l}-w^{l}_{\rm GD}\|_2
        \leq L_f^{-1} (1 + n L_f)^l \epsilon,
    \end{align*}
    where $L_f$ denotes the Lipschitz constant of $\mathcal{L}_N(w)$ within $\mathcal{W}$.
\end{corollary}

First we introduce a helper lemma.
\begin{lemma}[Error for Approximating GD, Lemma G.1 of \cite{bai2023transformers}]
\label{lem:error_general_m}
    Let $\mathcal{W} \subset \R^d$ is a convex bounded domain and ${\rm Proj}_\mathcal{W}$ projects all vectors into $\mathcal{W}$.
    Suppose $f:\mathcal{W} \rightarrow R$ and $\nabla f$ is $L_f$-Lipschitz on $\mathcal{W}$.
    Fix any $\epsilon>0$,
    let sequences
    $\{\bar{w}^l\}_{l\geq0} \in \R^d$ and
    $\{w^l_{\rm GD}\}_{l\geq0} \in \R^d$
    are given by $\bar{w}^0 = w^0_{\rm GD} = \mathbf{0}$, then for all $l \geq 0$,
    \begin{align*}
        \bar{w}^{l} = & ~
        {\rm Proj}_{\mathcal{W}} 
        (\bar{w}^{l-1} - \eta \nabla \mathcal{L}_n(\bar{w}^{l-1}) + \epsilon^{l-1}), \quad \|\epsilon^{l-1}\|_2 \leq \eta \epsilon,\\
        w^{l}_{\rm GD} = & ~
        {\rm Proj}_{\mathcal{W}} 
        (w^{l-1}_{\rm GD} - \eta \nabla \mathcal{L}_n(w^{l-1}_{\rm GD}))
    \end{align*}
    To show the convergence,  we define the gradient mapping at $w$ with step size $\eta$ as,
    \begin{align*}
        {\rm G}^f_{\mathcal{W},\eta}
        := \frac{w - 
        {\rm Proj}_{\mathcal{W}} 
        (w - \eta \nabla \mathcal{L}_n(w))}{\eta}.
    \end{align*} 
    Then if $\eta \leq L_f$, for all $L \geq 1$, convergence holds
    \begin{align*}
        \min_{l \in [L-1]}
        \|{\rm G}^f_ {\mathcal{W},\eta}(\bar{w}^{l})\|^2_2 
        \leq
        \frac{1}{L} \sum_{l=1}^{L-1}
        \|{\rm G}^f_ {\mathcal{W},\eta}(\bar{w}^{l})\|^2_2 
        \leq
        \frac{8(f(\mathbf{0}) - 
        \inf_{w \in \mathcal{W}} f(w))}{\eta L} + 
        10 \epsilon^2.
    \end{align*}
    Moreover, for any $l \geq 0$, the error accumulation is
    \begin{align*}
        \|\bar{w}^{l}-w^{l}_{\rm GD}\|_2
        \leq L_f^{-1} (1 + n L_f)^l \epsilon.
    \end{align*}
\end{lemma}
\cref{lem:error_general_m} shows \cref{thm:icl_gd_m} leads to exponential error accumulation in the general case.
Moreover, \cref{lem:error_general_m} also provides convergence of approximating GD.
Then we proof \cref{coro:error_ICGD_m}.
\begin{proof}
For any small $\epsilon$, by \cref{thm:icl_gd_m}, the neural network ${\rm NN}_{\theta}$ implements each gradient descent step with error bounded by $\epsilon$. 
Then we simply apply \cref{lem:error_general_m} to complete the proof.
\end{proof}

\clearpage
\section{Extension: Different Input and Output Dimensions}
\label{app:sec_diff_io}

In this section, we explore the ICGD on $N$-layer neural networks under the setting where the dimensions of input $x_i$ and label $y_i$ can be different.
Specifically, we consider our prompt datasets $\{(x_i,y_i)\}_{i \in [n]}$ where $x_i \in \R^{d_x}$ and $y_i \in \R^{d_y}$.
We start with our new $N$-layer neural network.

\begin{definition}
[$N$-Layer Neural Network]
\label{def:N_nn_m_io}
An $N$-Layer Neural Network comprises $N-1$ hidden layers and $1$ output layer, all constructed similarly.
Let $r:\R \rightarrow \R$ be the activation function.
For the hidden layers: 
for any $i \in [n+1], j \in 
[N-1]$, and $ k \in [K]$, 
the output for the first $j$ layers w.r.t. input $x_i \in \R^d$, denoted by ${\rm pred}_h(x_i;j)\in \R^K$, is defined as recursive form:
\begin{align*}
    {\rm pred}_h 
    (x_i ; 1)[k]
    := r(v_{1_{k}}^{\top} x_i), \quad\text{and}\quad
    {\rm pred}_h 
    (x_i ; j)[k]
    := r(v_{j_{k}}^{\top} {\rm pred}_h 
    (x_i ; j-1)),  
    \end{align*}  
where  $v_{1_k} \in \R^d$  and $v_{j_k}\in\R^K$ for $j\in 
\{2,\ldots,N-1\}$ are the 
$k$-th parameter vectors in the first layer and the $j$-th layer, respectively.
For the output layer ($N$-th layer), 
the output for the first $N$ layers (i.e the entire neural network) w.r.t. input $x_i \in \R^{d_x}$, denoted by ${\rm pred}_o(x_i ;w, N) \in \R^{d_y}$, is defined for any $k \in [d_y]$ as follows:
\begin{align*}
    {\rm pred}_o(x_i ;w, N)[k]
    := r(v_{N_{k}}^{\top} 
    {\rm pred}_h 
    (x_i ; N-1)),
\end{align*} 
where $v_{N_k}\in\R^K$ are the $k$-th parameter vectors in the $N$-th layer and $w \in \R^{(d_x +d_y)K + (N-2)K^2}$ denotes the vector containing all parameters in the neural network, 
\begin{align*}
    w := 
    \begin{bmatrix}
        v_{1_1}^\top, \ldots, 
        v_{1_K}^\top, \ldots, v_{j_k}^\top, \ldots
        v_{{N-1}_1}^\top, \ldots, 
        v_{{N-1}_K}^\top,  
        v_{{N}_1}^\top, \ldots, 
        v_{{N}_{d_y}}^\top 
    \end{bmatrix}^\top.
\end{align*}
\end{definition}

Notice that our new $N$-layer neural network only modify the output layer compared to \cref{def:N_nn_m}.
Intuitively, this results in minimal change in output, which allows our framework in \cref{sec:ICL_for_LNN_m} to function across varying input/output dimensions.
Theoretically, we derive the explicit form of gradient $\nabla \mathcal{L}_n(w)$.

\begin{lemma}
[Decomposition of One Gradient Descent Step]
\label{lem:decomposition_gd_m_diff_io}
    Fix any $B_v, \eta > 0$.
    Suppose the empirical loss function $\mathcal{L}_n(w)$ on $n$ data points $\{(x_i,y_i)\}_{i \in [n]}$ is defined as 
    \begin{align*}
        \mathcal{L}_n(w) \coloneqq \frac{1}{2n} \sum_{i=1}^n \ell(f(w, x_i), y_i),\quad
        \text{where $\ell: \mathbb{R}^{d_y} \times \mathbb{R}^{d_y} \rightarrow \mathbb{R}$ is a loss function,}
    \end{align*}
    where $f(w, x_i), y_i)$ is the output of $N$-layer neural networks (\cref{def:N_nn_m_io}) with modified output layer. 
    Suppose closed domain $\mathcal{W}$ and projection function
    ${\rm Proj}_{\mathcal{W}}(w)$ follows \eqref{eqn:domain_w_m}. 
    Let $A_i(j), r'_i(j), R_i(j), V_j$ be as defined in \cref{def:abbv_m} (with modified dimensions), then the explicit form of gradient $\nabla \mathcal{L}_n(w)$ becomes
    \begin{align*}
        \nabla \mathcal{L}_n(w) =  
        \frac{1}{2n} \sum_{i=1}^{n}
         \begin{bmatrix}
            A_i(1) \\
            \vdots \\
            A_i(N)
         \end{bmatrix},
    \end{align*}
    where $A_i(j)$ denote the derivative of $\ell(p_i(N), y_i)$ with respect to the parameters in the $j$-th layer, 
    \begin{align*}
        A_i(j) = \begin{cases}
            (R_i(N-1) \cdot V_{N} \cdot \ldots \cdot R_i(j-1) \cdot
            \begin{bmatrix}
            \textbf{I}_{K \times K} \otimes 
            p_i(j-1)^{\top} 
            \end{bmatrix}
            )^{\top} \cdot (\pdv {\ell(p_i(N), y_i)}{p_i(N)})^\top, & j \neq N \\
            (R_i(N-1) \cdot
            \begin{bmatrix}
            \textbf{I}_{d_y \times d_y} \otimes 
            p_i(N-1)^{\top} 
            \end{bmatrix}
            )^{\top} \cdot (\pdv {\ell(p_i(N), y_i)}{p_i(N)})^\top, & j = N.
        \end{cases}
    \end{align*}
\end{lemma}

\begin{proof}
    Simply follow the proof of \cref{lem:decomposition_gd_m}.
    We show the different terms compared to \cref{def:abbv_m}:
    \begin{itemize}
        \item Let $D_j \in \mathbb{R}$ denote the total number of parameters in the first $j$ layers. 
        \begin{align*}
            D_j = \begin{cases}
            0, & j = 0 \\
            d_x K, & j = 1 \\
            (j-1)K^2 + d_x K, & 2 \leq j \leq N-1 \\
            (N-2)K^2 + (d_x + d_y)K, & j = N,
            \end{cases}
        \end{align*}
        \item The intermediate term $R_i(N-1)$,
        \begin{align*}
        R_i(N-1) = \mathrm{diag}\{r'(v_{{j+1}_1}^\top p_i(j)), \ldots, r'(v_{{j+1}_{d_y}}^\top p_i(j))\} \in \R^{d_y \times d_y}.
        \end{align*}
        \item The parameters matrices of the first and the last layers:
        \begin{align*}
        V_j := 
        \begin{cases}
            \begin{bmatrix} v_{1_1}, \ldots, v_{1_K} \end{bmatrix}^\top \in \R^{K \times d_x}, & j=1 \\
            \begin{bmatrix} v_{N_1}, \ldots, v_{N_{d_y}} \end{bmatrix}^\top \in \R^{d_y \times K}, & j=N.
        \end{cases}
    \end{align*}
    \end{itemize}
    Thus we complete the proof.
\end{proof}

\cref{lem:decomposition_gd_m_diff_io} shows that the explicit form of gradient $\nabla \mathcal{L}_n(w)$ holds the same structure as \cref{lem:decomposition_gd_m}.
Therefore, it is simple to follow our framework in \cref{sec:ICL_for_LNN_m} to approximate $\nabla \mathcal{L}_n(w)$ term by term.
Finally, we introduce the generalized version of main result \cref{thm:icl_gd_m}.

\begin{theorem}[In-Context Gradient Descent on $N$-layer NNs]
\label{thm:icl_gd_m_io}
    Fix any $B_v, \eta, \epsilon > 0, L \geq 1$.
    For any input sequences takes from $\eqref{eqn:input}$, where $\{(x_i,y_i)\}_{i \in [n]}$ and $x_i \in \R^{d_x}$ and $y_i \in \R^{d_y}$, their exist upper bounds $B_x,B_y$ such that for any $i \in [n]$, $\|y_i\|_2 \leq B_y$, $\|x_i\|_2 \leq B_x$.
    Assume functions $r(t)$, $r'(t)$ and $u(t,y)[k]$ are $L_r,L_{r'},L_l$-Lipschitz continuous.
    Suppose $\mathcal{W}$ is a closed domain such that for any $j \in [N-1]$ and $k \in [K]$,
    \begin{align*}
        \mathcal{W} \subset \left\{
        w = [v_{j_k}] \in \mathbb{R}^{D_N} : \|v_{j_k}\|_2 \leq B_v
        \right\},      
    \end{align*}
    and ${\rm Proj}_{\mathcal{W}}$ project $w$ into bounded domain $\mathcal{W}$. 
    Assume
    ${\rm Proj}_{\mathcal{W}} = {\rm MLP}_{\theta}$ for some MLP layer with hidden dimension 
    $D_w$ parameters $\|\theta\| \leq C_{w}$.
    If functions $r(t)$, $r'(t)$ and $u(t,y)[k]$ are $C^4$-smoothness, then for any $\epsilon > 0$, there exists a transformer model ${\rm NN}_{\theta}$ with $(2N+4)L$ hidden layers
    consists of $L$ neural network blocks 
    ${\rm TF}_{\theta}^{N+2} \circ
    {\rm EWML}_{\theta}^{N} \circ
    {\rm TF}_{\theta}^{2}$,
    \begin{align*}
        {\rm NN}_{\theta} := 
        {\rm TF}_{\theta}^{N+2} \circ
        {\rm EWML}_{\theta}^{N} \circ
        {\rm TF}_{\theta}^{2}
        \circ \ldots \circ 
        {\rm TF}_{\theta}^{N+2} \circ
        {\rm EWML}_{\theta}^{N} \circ
        {\rm TF}_{\theta}^{2},
    \end{align*}
    such that the heads number $M^l$, parameter dimensions $D^l$, and the parameter norms
    $B_{\theta^l}$ suffice
    \begin{align*}
        \max_{l \in [(2N+4)L]} M^l \leq \Tilde{O}(\epsilon^{-2}), \quad
        \max_{l \in [(2N+4)L]} D^l \leq O(K^2N) + D_w, \quad
        \max_{l \in [(2N+4)L]} B_{\theta^l} \leq 
        O(\eta) + C_w + 1,
    \end{align*}
    where $\Tilde{O}(\cdot)$ hides the constants that depend on $d,K,N$, the radius parameters $B_x,B_y,B_v$ and the smoothness
    of $r$ and $\ell$. 
    And this neural network such that for any input sequences $H^{(0)}$, take from \eqref{eqn:input}, ${\rm NN_{\theta}}(H^{(0)})$ implements $L$ steps in-context gradient descent on risk $\mathcal{L}_n(w)$ follows \cref{lem:decomposition_gd_m_diff_io}: For every $l \in [L]$, the $(2N+4)l$-th layer outputs $h_i^{((2N+4)l)} = [x_i; y_i; \bar{w}^{(l)}; \mathbf{0}; 1; t_i]$ for every $i \in [n+1]$, and approximation gradients
    $\bar{w}^{(l)}$ such that 
    \begin{align*}
        \bar{w}^{(l)} = 
        {\rm Proj}_{\mathcal{W}} 
        (\bar{w}^{(l-1)} - \eta \nabla \mathcal{L}_n(\bar{w}^{(l-1)}) + \epsilon^{(l-1)}), \quad \bar{w}^{(0)} = \mathbf{0},
    \end{align*}
    where $\|\epsilon^{(l-1)}\|_2 \leq \eta \epsilon$ is an error term.
\end{theorem}

\clearpage
\section{Extension: Softmax Transformer}
\label{app:sec_soft}

In this part, we demonstrate the existence of pretrained $\Softmax$ transformers capable of implementing ICGD on an $N$-layer neural network. 
First, we introduce our main technique: the universal approximation property of softmax transformers in \cref{app:subsec_app_soft}. 
Then, we prove the existence of pretrained softmax transformers that implement ICGD on $N$-layer neural networks in \cref{app: subsec_icgd_soft}.

\subsection{Axillary Lemma: Universal Approximation of Softmax Transformer}
\label{app:subsec_app_soft}

\textbf{Softmax-Attention Layer.}
We replace modified normalized ReLU activation $\sigma/n$ in ReLU attention layer (\cref{def:attn}) by standard softmax. 
Thus, for any input sequence $H \in \R^{D \times n}$,  a single head attention layer outputs
\begin{align} \label{eq:softmax_attn}
    {\rm Attn}
    \left( H \right) 
    = H + W^{(O)} (V H) 
    \Softmax
    \left[ (K H)^{\top}
    (Q H)\right],
\end{align} 
where $W^{(O)}, Q, K, V \in \mathbb{R}^{D \times D} \in \mathbb{R}^{d \times d}$ are the weight matrices.
Then we introduce the softmax transformer block, which consists of two feed-forward neural network layers and a single-head
self-attention layer with the softmax function.

\begin{definition}[Transformer Block $\calT_{\Softmax}$]\label{def:Trans_net_class}
For any input sequences $H \in \R^{D \times n}$,
let ${\rm FF}(H) \coloneqq H + W_2 \cdot {\rm ReLU}(W_1 H + b_1 \one_L^\sT) + b_2 \one_L^\sT $ be the Feed-Forward layer, where $d'$ is hidden dimensions, $W_1 \in \R^{d' \times D}$, $W_2 \in \R^{D \times d'}$, $ b_1 \in \RR^{l}$, and $ b_2 \in \RR^{d}$.
We configure a transformer block with Softmax-attention layer as
$\calT_{\Softmax}\coloneqq \{ {\rm FF} \circ {\rm Attn} \circ {\rm FF}:\R^{d\times L}\to \R^{d\times L} \}$.
\end{definition}

\textbf{Universal Approximation of Softmax-Transformer.}
We show the universal approximation theorem for
Transformer blocks (\cref{def:Trans_net_class}). 
Specifically, Transformer blocks $\calT_{\Softmax}$ are universal
approximators for continuous permutation equivariant functions on bounded domain.
\begin{lemma}[Universal Approximation of $\calT_{\Softmax}$]
\label{lemma:univerality_softmax}
Let $f(\cdot) \coloneqq \RR^{d \times n} \to \RR^{d \times n}$ be any $L$-Lipschitz permutation equivariant function supported on $[0, B_x]^{d \times n}$. 
We denote the discrete input domain of $[0,B_x]^{d \times n}$ by a grid $\mathbb{G}_{D}$ with granularity
$D \in \mathbb{N}$ defined as $\mathbb{G}_{D}=\{B_x/D,2B_x/D,\ldots,B_x\}^{d\times n}\subset\mathbb{R}^{d\times n}$.
For any $\kappa > 0$, there exists a transformer network $f_{\Softmax} \in \calT_{\Softmax}$, such that for any $Z \in [0, B_x]^{d \times n}$, it approximate $f(Z)$ as:
\begin{align*}
    \|f_{\Softmax}(Z) - f(Z) \|_2 \leq \kappa.
\end{align*}
\end{lemma}

\begin{proof}[Proof Sketch]
    First, we use a piece-wise constant function to approximate $f$ and derive an upper bound based on its $L$-Lipschitz property. 
    Next, we demonstrate how the feed-forward neural network  $\mathcal{F}_1^{(FF)}$ quantizes the continuous input domain into the discrete domain $\mathbb{G}_D$  through a multiple-step function, using ReLU functions to create a piece-wise linear approximation.  
    Then, we apply the self-attention layer  $\mathcal{F}^{(SA)}$  on  $\mathcal{F}_1^{(FF)}$ , establishing a bounded output region for  $\mathcal{F}_S^{(SA)} \circ \mathcal{F}_1^{(FF)}$. 
    Finally, we employ a second feed-forward network $\mathcal{F}_2^{(FF)}$ to predict $f_{\Softmax}(Z)$ and assess the approximation error relative to the actual output $f(Z)$ .
    See \cref{app:univerality_softmax} for a detailed proof.
\end{proof}

\subsection{In-Context Gradient Descent with Softmax Transformer}
\label{app: subsec_icgd_soft}
\textbf{In-Context Gradient Descent with Softmax Transformer.}
By applying universal approximation theory (\cref{lemma:univerality_softmax}), we now illustrate how to use Transformer block $\calT_{\Softmax}$ (\cref{def:Trans_net_class}) and MLP layers (\cref{def:mlp}) to implement ICGD on general risk function $\mathcal{L}_n(w)$.  
\begin{theorem}[\cref{thm:icgd_soft_main} Restated: In-Context Gradient Descent on General Risk Function]
\label{thm:icgd_soft}
    Fix any $B_w, \eta, \epsilon > 0, L \geq 1$.
    For any input sequences takes from $\eqref{eqn:input}$, their exist upper bounds $B_x,B_y$ such that for any $i \in [n]$, $\|y_i\|_{\max} \leq B_y$, $\|x_i\|_{\max} \leq B_x$.
    Suppose $\mathcal{W}$ is a closed domain such that
    $\|w\|_{\max} \leq B_w$
    and ${\rm Proj}_{\mathcal{W}}$ project $w$ into bounded domain $\mathcal{W}$. 
    Assume
    ${\rm Proj}_{\mathcal{W}} = {\rm MLP}_{\theta}$ for some MLP layer.
    Define $l(w, x_i,y_i)$ as a loss function with $L$-Lipschitz gradient.
    Let $\mathcal{L}_n(w) = \frac{1}{n}\sum_{i=1}^n 
    \ell(w, x_i,y_i)$ denote the empirical loss function, then there exists a $\Softmax$-transformer ${\rm NN}_{\theta}$,  such that for any input sequences $H^{(0)}$, take from \eqref{eqn:input}, ${\rm NN_{\theta}}(H^{(0)})$  implements $L$ steps in-context gradient descent on $\mathcal{L}_n(w)$: 
    For every $l \in [L]$, the $4l$-th layer outputs $h_i^{(4l)} = [x_i; y_i; \bar{w}^{(l)}; \mathbf{0}; 1; t_i]$ for every $i \in [n+1]$, and approximation gradients
    $\bar{w}^{(l)}$ such that 
    \begin{align*}
        \bar{w}^{(l)} = 
        {\rm Proj}_{\mathcal{W}} 
        (\bar{w}^{(l-1)} - \eta \nabla \mathcal{L}_n(\bar{w}^{(l-1)}) + \epsilon^{(l-1)}), \quad \bar{w}^{(0)} = \mathbf{0},
    \end{align*}
    where $\|\epsilon^{(l-1)}\|_2 \leq \eta \epsilon$ is an error term.
\end{theorem}

\subsection{Proof of \texorpdfstring{\cref{thm:icgd_soft_main}}{}}
\label{app:proof_thm_icgd_soft}

\begin{proof}[Proof of \cref{thm:icgd_soft_main}]
    We only need to construct a $4$ layers transformer capable of implementing single step gradient descent.
    With out loss of generality, we assume $w \in \R^{D_w}$.
    Recall that the input sequences $H \in \R^{D \times n}$ takes form
    \begin{align}
    \label{eqn:input_l}
    H \coloneqq \begin{bmatrix}
    x_1 & x_2 & \cdots & x_n & x_{n+1} \\
    y_1 & y_2 & \cdots & y_n & 0 \\
    q_1 & q_2 & \cdots & q_n & q_{n+1}
    \end{bmatrix} \in \mathbb{R}^{D \times (n+1)},\quad
    q_i \coloneqq 
    \begin{bmatrix}
    w \\
    0 \\
    1 \\
    t_i
    \end{bmatrix} 
    \in \mathbb{R}^{D-(d+1)}.
    \end{align}
    Let function $f: \R^{D \times n} \rightarrow \R^{D \times n}$ output 
    \begin{align*}
        f(H) = 
        \begin{bmatrix}
            x_1 & x_2 & \cdots & x_n & x_{n+1} \\
            y_1 & y_2 & \cdots & y_n & 0 \\
            q_1 & q_2 & \cdots & q_n & q_{n+1}
        \end{bmatrix},
        \quad
        q_i \coloneqq 
        \begin{bmatrix}
        w - \eta \nabla \mathcal{L}_n(w) \\
        0 \\
        1 \\
        t_i
        \end{bmatrix} 
        \in \mathbb{R}^{D-(d+1)}.
    \end{align*}
    By \cref{lemma:univerality_softmax}, for any $\kappa>0$, there exists a transformer network $f_{\Softmax} \in \calT_{\Softmax}$, such that for any input $H \in [-B, B]^{d \times L}$, we have $\norm{f_{\Softmax}(H) - f(H)}_{2} \le \kappa$.
    Therefore, by the equivalence of matrix norms, $\norm{f_{\Softmax}(H) - f(H)}_{\max} \le \kappa$ holds without loss of generality.
    Above $B:=\max 
    \{B_x,B_y,B_w,1\}$ denotes the upper bound for every elements in $H$.
    Thus, we obtain 
    $\bar{w}$ from the identical position of $w$ in $f_{\Softmax}(H)$.
    Suppose we choose $\kappa = \frac{\epsilon}{\sqrt{D_w}}$, then it holds
    \begin{align*}
        \|\bar{w}-(w-\eta \nabla \mathcal{L}_n(w)\|_2
        \leq & ~ \sqrt{D_w} 
        \|\bar{w}-(w-\eta \nabla \mathcal{L}_n(w)\|_{\max} \\
        \leq & ~ \|f_{\Softmax} - f(H)\|_{\max} \\
        \leq & ~ \sqrt{D_w} \cdot
        \frac{\epsilon}{\sqrt{D_w}} \\
        \leq  & ~ \epsilon.
    \end{align*}
    Finally, by our assumption, there exists an MLP layer such that for any $i \in [n+1]$, it maps
    \begin{align*}
        [x_i; y_i; w - \eta \nabla \mathcal{L}_n(w); \mathbf{0}; 1; t_i]
        \xrightarrow
        {{\rm MLP}}
        [x_i; y_i; {\rm Proj}_{\mathcal{W}}(w - \eta \nabla_w \mathcal{L}_n(w));  \mathbf{0}; 1; t_i].
    \end{align*}
    Therefore, a four-layer transformer $f_{\Softmax} \circ {\rm MLP}$ is capable of implementing one-step gradient descent through ICL. 
    As a direct corollary, 
    there exist a $4L$-layer transformer consists of $L$ identical blocks $f_{\Softmax} \circ {\rm MLP}$ to approximate $L$ steps gradient descent algorithm.
    Each block approximates a one-step gradient descent algorithm on general risk function $\mathcal{L}_n(w)$.
\end{proof}

\subsection{Proof of \texorpdfstring{\cref{lemma:univerality_softmax}}{}}
\label{app:univerality_softmax}

In this section, we introduce a helper lemma \cref{lem:soft_contextual} to prove \cref{lemma:univerality_softmax}.
At the beginning, we assume all input sequences are separated by a certain distance.
\begin{definition}[Token-wise Separateness, Definition 1 of \cite{kajitsuka2023transformers}]
    Let $N \geq 1$ and $Z^{(1)}, \ldots, Z^{(N)} \in \mathbb{R}^{d \times n}$ be input sequences. 
    Then, $Z^{(1)}, \ldots, Z^{(N)}$ are called token-wise $\left(r_{\min }, r_{\max }, \delta\right)$-separated if the following three conditions hold.
    \begin{itemize}
        \item For any $i \in[N]$ and $k \in[n],\left\|Z_{:, k}^{(i)}\right\|_2 >r_{\min }$ holds.
        \item For any $i \in[N]$ and $k \in[n],\left\|Z_{:, k}^{(i)}\right\|_2 <r_{\max }$ holds.
        \item For any $i, j \in[N]$ and $k, l \in[n]$ with $Z_{:, k}^{(i)} \neq Z_{:, l}^{(j)},\left\|Z_{:, k}^{(i)}-Z_{:, l}^{(j)}\right\|_2 > \delta$ holds.
    \end{itemize} 
Note that we refer to $Z^{(1)}, \ldots, Z^{(N)}$ as token-wise $\left(r_{\max }, \epsilon\right)$-separated instead if the sequences satisfy the last two conditions.
\end{definition}

Then we introduce the definition of contextual mapping. 
Intuitively, a contextual mapping can provide every input sequence with 
a unique id, which enables us to construct approximation for labels. 
\begin{definition}[Contextual mapping, Definition 2 of \cite{kajitsuka2023transformers}]
    Let input sequences $Z^{(1)}, \ldots, Z^{(N)} \in \mathbb{R}^{d \times n}$. Then, a map $q: \mathbb{R}^{d \times n} \rightarrow \mathbb{R}^{d \times n}$ is called an $(r, \delta)$-contextual mapping if the following two conditions hold:
    \begin{itemize}
        \item For any $i \in[N]$ and $k \in[n],\left\|q\left(Z^{(i)}\right)_{:, k}\right\|_2 <r$ holds.
        \item For any $i, j \in[N]$ and $k, l \in[n]$, if $Z_{:, k}^{(i)} \neq Z_{:, l}^{(j)}$, then $\left\|q\left(Z^{(i)}\right)_{:, k}-q\left(Z^{(j)}\right)_{:, l}\right\|_2>\delta$ holds.
    \end{itemize}
    In particular, $q\left(Z^{(i)}\right)$ for $i \in[N]$ is called a context id of $Z^{(i)}$.
\end{definition}

Next, we show that a softmax-based 1-layer attention block with low-rank
weight matrices is a contextual mapping for almost all input sequences.
\begin{lemma}[Softmax attention is contextual mapping, Theorem 2 of \cite{kajitsuka2023transformers}]
\label{lem:soft_contextual}
Let $Z^{(1)}, \ldots, Z^{(N)} \in \mathbb{R}^{d \times n}$ be input sequences with no duplicate word token in each sequence, that is,
\begin{align*}
Z_{:, k}^{(i)} \neq Z_{:, l}^{(i)},
\end{align*}
for any $i \in[N]$ and $k, l \in[n]$. Also assume that $Z^{(1)}, \ldots, Z^{(N)}$ are token-wise $\left(r_{\min }, r_{\max }, \epsilon\right)$ separated. 
Then, there exist weight matrices $W^{(O)} \in \mathbb{R}^{d \times s}$ and $V, K, Q \in$ $\mathbb{R}^{s \times d}$ such that the ranks of $V, K$ and $Q$ are all 1, and 1-layer single head attention with softmax, i.e., $\mathcal{F}_S^{(S A)}$ with $h=1$ is an $(r, \delta)$-contextual mapping for the input sequences $Z^{(1)}, \ldots, Z^{(N)} \in \mathbb{R}^{d \times n}$ with $r$ and $\delta$ defined by
\begin{align*}
r = & ~ r_{\max }+\frac{\epsilon}{4} \\
\delta = & ~ \frac{2(\log n)^2 \epsilon^2 r_{\min }}{r_{\max }^2(|\mathcal{V}|+1)^4(2 \log n+3) \pi d} \exp \left(-(|\mathcal{V}|+1)^4 \frac{(2 \log n+3) \pi d r_{\max }^2}{4 \epsilon r_{\min }}\right).
\end{align*}
\end{lemma}
Applying \cref{lem:soft_contextual}, we extends Proposition 1 of \cite{kajitsuka2023transformers} to our \cref{lemma:univerality_softmax}\footnote{This extension builds on the results of \cite{hu2024fundamental}, which extend the rank-1 requirement to any rank for attention weights. Additionally, \citet{hu2024statistical} apply similar techniques to analyze the statistical rates of diffusion transformers (DiTs).
}.
We provide explicit upper bound of error $\| f_{\Softmax}(Z) - f(Z) \|_2$ and analysis with function $f$ of a broader supported domain.

\begin{lemma}[\cref{lemma:univerality_softmax} Restated: Universal Approximation of $\calT_{\Softmax}$]
Let $f(\cdot) \coloneqq \RR^{d \times n} \to \RR^{d \times n}$ be any $L$-Lipschitz permutation equivariant function supported on $[0, B_x]^{d \times n}$. 
We denote the discrete input domain of $[0,B_x]^{d \times n}$ by a grid $\mathbb{G}_{D}$ with granularity
$D \in \mathbb{N}$ defined as $\mathbb{G}_{D}=\{B_x/D,2B_x/D,\ldots,B_x\}^{d\times n}\subset\mathbb{R}^{d\times n}$.
For any $\kappa > 0$, there exists a transformer network $f_{\Softmax} \in \calT_{\Softmax}$ (\cref{def:Trans_net_class}), such that for any $Z \in [0, B_x]^{d \times n}$, it approximate $f(Z)$ as:
\begin{align*}
    \| f_{\Softmax}(Z) - f(Z) \|_2 \leq \kappa.
\end{align*}
\end{lemma}

\begin{proof}
    We begin our 3-step proof.
    \paragraph{Approximation of $f$ by piece-wise constant function.}
    Since $f$ is a continuous function on a compact set,  $f$ has maximum and minimum values on the domain.  
    By scaling with $\mathcal{F} _{1}^{( FF) }$ and $\mathcal{F} _{2}^{( FF) }$, $f$ is assumed to be normalized: for any $Z\in\mathbb{R}^{d\times n} \setminus [0,B_x]^{d \times n}$
    \begin{align*}
        f(Z) = 0,
    \end{align*}
    and for any $Z\in[0,B_x]^{d\times n}$
    \begin{align*}
        -B_y \leq f(Z)\leq B_y.    
    \end{align*}
    
    Let $D\in\mathbb{N}$ be the granularity of a grid $\mathbb{G}_{D}$:
    \begin{align*}
    \mathbb{G}_{D}=\{\frac{B_x}{D},\frac{2B_x}{D},\ldots,B_x\}^{d\times n}\subset\mathbb{R}^{d\times n},
    \end{align*}
    where each coordinate only take discrete value  $B_x/D, 2B_x/D,...,B_x$.
    Now with a continuous input $Z$, we approximate $f$ by using a piece-wise constant function $\Bar{f}$ evaluating on the nearest grid point $L$ of $Z$ in the following way:
    \begin{align}
    \Bar{f}(Z)=\sum_{L\in \mathbb{G}_{D}}f\left(L\right)1_{Z\in L+[-B_x/D,0)^{d\times n}}.
    \end{align}

    Additionally if $Z \in L+[-1/D,0)^{d\times n}$, denote it as $Q(Z) = L$.

    Now we bound the piece-wise constant approximation error $ \| f-\Bar{f} \|$ as follows.

    Define set $P_D = \{ L+[-B_x/D,0)^{d\times n}| L\in \mathbb{G}_D \}$. 
    It is a set of regions of size $(\frac{B_x}{D})^{d\times n}$, whose vertexes are the points in $\mathbb{G}_D$.

    For any subset $U \in P_D$, the maximal difference of $f$ and $\Bar{f}$ in this region is: 
    \begin{align}
        \max_{Z\in U} \|f(Z)-\Bar{f}(Z)\|_2 
        = & ~
        \max_{Z\in U} \|f(Z)-f(Q(Z))\|_2 \nonumber \\
        \leq & ~
        \max_{Z, Z'\in U} \|f(Z) - f(Z')\|_2 \nonumber \\
        \leq & ~
        L \cdot \max_{Z, Z'\in U} \|Z - Z' \|_2 \annot{By $f$ is a $L$-Lipschitz function} \nonumber \\
        = & ~
        L \cdot \sqrt{dn \cdot (\frac{B_x}{D})^2} \annot{$Z$, $Z'$ are in the same $\frac{B_x}{D}$-wide ($d\cdot n$)-dimension $U$.} \\
        = & ~
        \frac{L\sqrt{dn}B_x}{D}.
        \label{eqn:f_f_bar}
    \end{align}
    
    \paragraph{Quantization of input using $\mathcal{F}_1^{(FF)}$.}
    In the second step, we use $\mathcal{F}_1^{(FF)}$ to quantize the continuous input domain into $\mathbb{G}_D.$
    This process is achieved by a multiple-step function, and we use ReLU functions to approximate this multiple-step functions.
    This ReLU function can be easily implemented by a one-layer feed-forward network.

    First for any small $\delta>0$ and $z\in\mathbb{R}$, we construct a $\delta$-approximated step function using ReLU functions:
    \begin{align}
    \label{eqn:first_relu}
        \frac{\sigma_R\left[\frac{z}{\delta}\right]-\sigma_R\left[\frac{z}{\delta}-B_x\right]}D
        =
        \begin{cases}0&z<0\\\frac{z}{\delta D}&0\leq z<\delta B_x\\\frac{B_x}{D}&\delta B_x\leq z\end{cases},  
    \end{align}
    where a one-hidden-layer feed-forward neural network is able to implement this.
    By shifting \eqref{eqn:first_relu} by $B_x$, for any $t \in [D-1]$, we have:
    \begin{align}
    \label{eqn:relu_appro_t}
        \frac{\sigma_R\left[\frac{z}{\delta}-\frac{tB_x}{\delta D}\right]-\sigma_R\left[\frac{z}{\delta}-B_x-\frac{tB_x}{\delta D}\right]}D
        =
        \begin{cases}
        0&z<\frac{tB_x}{D}\\
        \frac{z}{\delta D}&\frac{tB_x}{D}\leq z<\delta B_x+\frac{tB_x}{D}\\
        \frac{B_x}{D}&\delta B_x+\frac{tB_x}{D}\leq z
        \end{cases},
    \end{align}

    when $\delta$ is small the above function approximates to a step function:
    \begin{align*}
        {\rm quant}_D^{(t)}(z)
        =
        \begin{cases}
        0&z \leq \frac{tB_x}{D}\\
        \frac{B_x}{D}&\frac{tB_x}{D}\leq z
        \end{cases}.
    \end{align*}
    
    By adding up \eqref{eqn:relu_appro_t} at every $t \in [D-1]$, we have an approximated multiple-step function
    \begin{align}
        & \sum_{t=0}^{D-1}\frac{\sigma_R\left[\frac{z}{\delta}-\frac{tB_x}{\delta D}\right]-\sigma_R\left[\frac{z}{\delta}-B_x-\frac{tB_x}{\delta D}\right]}D \label{eq:2.112} \\
        \approx ~
        & \sum_{t=0}^{D-1}{\rm quant}_D^{(t)}(z) \annot{when $\delta$ is small.} \\
        = ~ & {\rm quant}_D(z) \nonumber \\
        = ~
        & \begin{cases}0&z<0\\ \frac{B_x}{D}&0\leq z<\frac{B_x}{D}\\\vdots&\vdots\\B_x&B_x-\frac{B_x}{D}\leq z\end{cases} \label{eqn:quant}.
    \end{align}

    Note that the error of approximation at $z$ here estimated as:
    \begin{align}
        \abs{\sum_{t=0}^{D-1}\frac{\sigma_R\left[\frac{z}{\delta}-\frac{tB_x}{\delta D}\right]-\sigma_R\left[\frac{z}{\delta}-B_x-\frac{tB_x}{\delta D}\right]} D- {\rm quant}_D(z)} 
        \leq \frac{B_x}{D}
        \label{differ_quant_and_linear},
    \end{align}

    and for matrix $Z \in \R^{d \times n}$:
    \begin{align*}
        & ~ \| \sum_{t=0}^{D-1}\frac{\sigma_R\left[\frac{Z}{\delta}-\frac{Q(Z)}{\delta D}\right]-\sigma_R\left[\frac{Z}{\delta}-B_xE-\frac{Q(Z)}{\delta D}\right]}D
        - {\rm quant}_D(Z) \|_2 \\
        & ~ \leq 
        \sqrt{d \times n \times (\frac{B_x}{D})^2}  \annot{$Z \in \R^{d \times n}$} \\
        & ~ =
        \frac{B_x\sqrt{dn}}{D}.
    \end{align*}

    Subtract the last step function from \eqref{eq:2.112} we get the desired result: 
    \begin{align}
    \label{eqn:relu_real_quant}
        \sum_{t=0}^{D-1}\frac{\sigma_R\left[\frac{z}{\delta}-
        \frac{tB_x}{\delta D} \right]-\sigma_R\left[\frac{z}{\delta}-B_x-\frac{tB_x}{\delta D}\right]}D-(\sigma_R\left[\frac{z}{\delta}-\frac{B_x}{\delta}\right]-\sigma_R\left[\frac{z}{\delta} - 1 -\frac{B_x}{\delta}\right]). 
    \end{align}

    This equation approximate the quantization of input domain $[0,B_x]$ into $\{B_x/D,\ldots,B_x\}$ and making $\R \setminus[0,B_x]$ to $0$.
    In addition to the quantization of input domain $[0,B_x]$, we add a penalty term for input out of $[0,B_x]$ in the following way:
    \begin{align}
    \label{eqn:penalty}
        & ~ -B_x\sigma_R\left[\frac{(z-B_x)}{\delta}\right]+B_x\sigma_R\left[\frac{(z-B_x)}{\delta}-1\right]-B_x\sigma_R\left[\frac{-z}{\delta}\right]+B_x\sigma_R\left[\frac{-z}{\delta}-1\right]\\
        \approx & ~
        {\rm penalty}(z)=
        \begin{cases} 
            -B_x, & z\leq0\\
            0, & 0<z\leq B_x\\
            -B_x, & B_x<z .
        \end{cases}. \nonumber
    \end{align}

    Both \eqref{eqn:relu_real_quant} and \eqref{eqn:penalty} can be realized by the one-layer feed-forward neural network.
    Also, it is straightforward to show that generate both of them to input $Z \in \R^{d \times n}$.
    
    Combining both components together, the fırst feed-forward neural network layer $\mathcal{F}_{1}^{(FF)}$ approximates the following function $\overline{\mathcal{F}}_1^{(FF)}(Z)$:
    \begin{align}
    \label{eqn:ffn_1}
        \mathcal{F}_{1}^{(FF)} \approx \overline{\mathcal{F}}_1^{(FF)}(Z)={\rm quant}_D^{d\times n}(Z)+\sum_{t=1}^d\sum_{k=1}^n {\rm penalty}(Z_{t,k}).
    \end{align}
    
    Note how we generalize ${\rm penalty (\cdot)}$ to multi-dimensional occasions in the above equation.
    Whenever an input sequence $Z$ has one entry $Z_{t,k}$ out of $[0,B_x]^{d\times n}$, we penalize the whole input sequence by adding a $-B_x$ to all entries. 
    This makes all entries of this quantization lower bounded by $-dnB_x$.
    
     \eqref{eqn:ffn_1} quantizes inputs in $[0,B_x]^{d\times n}$ with granularity $D$, while every element of the output is non-positive for inputs outside $[0,B_x]^{d\times n}.$ 
     In particular, the norm of the output is upper-bounded when every entry in $Z$ is out of $[0,B_x]$, this adds $-dnB_x$ penalties to all entries:
    \begin{align}
        \max_{Z\in\mathbb{R}^{d\times n}}\left\|\mathcal{F}_1^{(FF)}(Z)_{:,k}\right\|_2
        = & ~
        \sqrt{d \cdot (-dnB_x)^2}  \annot{One column is $d-$dimension.} \\
        \leq & ~
        dn\cdot\sqrt{d}B_x,
        \label{max_F^FF}
    \end{align}
    for any $k\in[n]$.

    \paragraph{Estimating the Influence of Self-Attention $\mathcal{F}^{(SA)}$.}
    Define $\tilde{\mathbb{G}}_D\subset\mathbb{G}_D$ as:
    \begin{align}
        \tilde{\mathbb{G}}_D=\{L\in\mathbb{G}_D\mid\forall k,l\in[n],L_{:,k}\neq L_{:,l}\}\:.
        \label{tilde_G_D}
    \end{align}

    It is a set of all the input sequences that don't have have identical tokens after quantization.
    
    Within this set, the elements are at least $\frac{B_x}{D}$ separated by the quantization. 
    Thus \cref{lem:soft_contextual} allows us to construct a self-attention $\mathcal{F}^{(SA)}$ to be a contextual mapping for such input sequences. 
    
    Since when $D$ is sufficiently large, originally different tokens will still be different after quantization. In this context, we omit $\mathbb{G}_D / \tilde{\mathbb{G}}_D$ for simplicity.

    From the proof of \cref{lem:soft_contextual} in \cite{kajitsuka2023transformers}, we follow their way to construct self-attention and have following equation:
    \begin{align}
      \left\|\mathcal{F}_S^{(SA)}(Z)_{:,k}-Z_{:,k}\right\|_2
      < \frac{1}{4\sqrt{d}D} \max_{k'\in[n]}
      \|Z_{:,k'}\|_2  
      \label{max_F^SA},
    \end{align}
    
    for any $k\in[n]$ and $Z\in\mathbb{R}^{d\times n}$. 
    
    Combining this upper-bound with \eqref{max_F^FF} we have
    
    \begin{align}   
    \left\|\mathcal{F}_S^{(SA)}\circ\mathcal{F}_1^{(FF)}\left(Z\right)_{:,k} -\mathcal{F}^{(FF)}\left(Z\right)_{:,k}\right\|_2
    < & ~
    \frac{1}{4\sqrt{d}D}\max_{k'\in[n]}\|\mathcal{F}^{(FF)}(Z_{:,k})\|_2 \nonumber \\
    < & ~
    \frac{1}{4\sqrt{d}D}\times dn\sqrt{d}B_x \annot{By $\eqref{max_F^FF}$} \\
    = & ~
    \frac{dnB_x}{4D}
    \label{bound_of_out_ranged}.
    \end{align}

    We show that if we take large enough $D$, every element of the output for $Z\in\mathbb{R}^{d\times n}\backslash[0,B_x]^{d\times n}$
    is upper-bounded by
    
    \begin{align}
        \mathcal{F}_S^{(SA)}\circ\mathcal{F}_1^{(FF)}\left(Z\right)_{t,k}<\frac{B_x}{4D}\quad(\forall t\in[d],\:k\in[n]).
        \label{out_zone_bound}
    \end{align}
    
    To show \eqref{out_zone_bound} holds, we consider the opposite occasion that there exists a $ \mathcal{F}_S^{(SA)}\circ\mathcal{F}_1^{(FF)}\left(Z\right)_{t_0,k_0} \geq B_x/4D$.
    Then we divide the case into two sub cases:
    \begin{enumerate}
        \item The whole $\mathcal{F}_1^{(FF)}\left(Z\right)$ receives no less than $2$ penalties.
    In this occasion, since every entry consists of two counterparts in \eqref{eqn:ffn_1}:
    the quantization part ${\rm quant_D^{d \times n}}(Z) \in [0,B_x]$ and aggregated with a penalty part $\sum_{t=1}^d\sum_{k=1}^n {\rm penalty}(Z_{t,k}) \leq -2B_x$, 
    for every entry we have $\mathcal{F}^{(FF)}\left(Z\right)_{t,k} \leq -B_x$. 
    
    This yields that:
    \begin{align}
        \|\mathcal{F}_S^{(SA)}\circ\mathcal{F}_1^{(FF)}\left(Z\right)_{:,k_0}-\mathcal{F}^{(FF)}\left(Z\right)_{:,k_0}\|_2
        \geq & ~
        \| \mathcal{F}_S^{(SA)}\circ\mathcal{F}_1^{(FF)}\left(Z\right)_{t_0,k_0}-\mathcal{F}^{(FF)}\left(Z\right)_{t_0,k_0} \|_2 \nonumber\\
        \geq & ~
        | \frac{B_x}{4D}-(-B_x) | \nonumber\\
        \geq & ~
        \frac{dn}{4D}B_x \annot{for a large enough D},
    \end{align}

    thus we derive a contradiction towards $\eqref{bound_of_out_ranged}$ from the assumption, proving it to be incorrect.

        \item 
        The whole $\mathcal{F}_1^{(FF)}\left(Z\right)$ receives only one penalty. 
    In this case all entries in $Z$ is penalized by $-B_x$ and satisfies:
    \begin{align}
    \label{eqn:case_2}    \mathcal{F}_1^{(FF)}\left(Z\right)_{t,k} \in [-B_x,0]^{d\times n}.
    \end{align}
    By \eqref{max_F^SA}, this further denotes:
    \begin{align}
        \left\|\mathcal{F}_S^{(SA)}\circ \mathcal{F}_1^{(FF)}\left(Z\right)_{:,k}-\mathcal{F}_1^{(FF)}\left(Z\right)_{:,k}\right\|_2
        < & ~
        \frac{1}{4\sqrt{d}D}\max_{k'\in[n]}\|\mathcal{F}_1^{(FF)}\left(Z\right)_{:,k'}\|_2\annot{By \eqref{max_F^SA}} \\
        \leq & ~
        \frac{1}{4\sqrt{d}D}\sqrt{d\times B_x^2} \annot{By \eqref{eqn:case_2}} \\
        = & ~
        \frac{B_x}{4D}.
        \label{eqn:bound_2}
    \end{align}   
    \end{enumerate}

    Yet by our assumption, there exists such an entry $\mathcal{F}_S^{(SA)} \circ \mathcal{F}^{(FF)}\left(Z\right)_{t_0,k_0} \geq B_x/4D$, which since $\mathcal{F}_1^{(FF)}\left(Z\right)_{t_0,k_0} \leq 0$, yields:
    \begin{align*}
        \left\|\mathcal{F}_S^{(SA)} \circ \mathcal{F}_1^{(FF)}\left(Z\right)_{:,k_0}-\mathcal{F}_1^{(FF)}\left(Z\right)_{:,k_0}\right\|_2
        \geq & ~
        \left\|\mathcal{F}_S^{(SA)} \circ \mathcal{F}_1^{(FF)}\left(Z\right)_{t_0,k_0}-\mathcal{F}_1^{(FF)}\left(Z\right)_{t_0,k_0}\right\|_2 \\
        \geq & ~ 
        | \frac{B_x}{4D} - 0 | \\
        = & ~
        \frac{B_x}{4D}
    \end{align*}
    
    The final conclusion contradict the former result, suggesting the prerequisite to be fallacious.

    Joining the incorrectness of the two sub-cases of the opposite occasion, we confirm the upper bound when input $Z$ is outside $[0, B_x]^{d \times n}$ in \eqref{out_zone_bound}.

    For the input $Z$ inside $[0,B_x]^{d\times n}$, we now show it is lower-bounded by
    \begin{align}
        \mathcal{F}_S^{(SA)}\circ\mathcal{F}_1^{(FF)}\left(Z\right)_{t,k}>\frac{3B_x}{4D}\quad(\forall t\in[d],\:k\in[n]).
        \label{in_zone_bound}
    \end{align}

    By our construction, every entry $Z$ in $[0,B_x]^{d\times n}$ satisfies:
    \begin{align}
        \mathcal{F}_1^{(FF)}\left(Z\right)_{t,k} \in [\frac{B_x}{D},B_x].
        \label{eqn:z_inside_bound}
    \end{align}

    By $\eqref{max_F^SA}$:
    \begin{align}
      & ~ \left\|\mathcal{F}_S^{(SA)} \circ \mathcal{F}_1^{(FF)}\left(Z\right)_{:,k}-\mathcal{F}_1^{(FF)}\left(Z\right)_{:,k}\right\|_2 \nonumber\\
      < & ~
      \frac{1}{4\sqrt{d}D}\max_{k'\in[n]}\|\mathcal{F}_1^{(FF)}\left(Z\right)_{:k'}\|_2  \annot{By \eqref{max_F^SA}} \nonumber \\
      \leq & ~ 
      \frac{1}{4\sqrt{d}D}\sqrt{d\times B_x^2} \annot{$d$-dimensional vector with each entry has maximum value $B_x$.} \\
      = & ~
      \frac{B_x}{4D} \label{eqn:saff_ff_column_error}.
    \end{align}

    This yields:
    \begin{align}
        |\mathcal{F}_S^{(SA)} \circ \mathcal{F}_1^{(FF)}\left(Z\right)_{t,k}-\mathcal{F}_1^{(FF)}\left(Z\right)_{t,k}|
        \leq & ~
        \left\|\mathcal{F}_S^{(SA)} \circ \mathcal{F}_1^{(FF)}\left(Z\right)_{:,k}-\mathcal{F}_1^{(FF)}\left(Z\right)_{:,k}\right\|_2 \nonumber \\
        < & ~ 
        \frac{B_x}{4D}.
    \label{eqn:saff_ff_entry_error}
    \end{align}
    Finally, we have:
    \begin{align}
        & ~ \mathcal{F}_S^{(SA)}\circ \mathcal{F}_1^{(FF)}\left(Z\right)_{t,k} \nonumber\\
        > & ~
        \mathcal{F}_1^{(FF)}\left(Z\right)_{t,k} - \left\|\mathcal{F}_S^{(SA)} \circ \mathcal{F}_1^{(FF)}\left(Z\right)_{t,k}-\mathcal{F}_1^{(FF)}\left(Z\right)_{t,k}\right\|_2 \nonumber \\
        > & ~
        \frac{B_x}{D} - \|\mathcal{F}_S^{(SA)} \circ \mathcal{F}_1^{(FF)}\left(Z\right)_{t,k}-\mathcal{F}_1^{(FF)}\left(Z\right)_{t,k}\|_2 \annot{By \eqref{eqn:z_inside_bound}.} \\
        > & ~
        \frac{B_x}{D} - \frac{B_x}{4D} \annot{By \eqref{eqn:saff_ff_entry_error}} \\
        = & ~
        \frac{3B_x}{4D} \nonumber.
    \end{align}
    Hence we finally finish the proof for the upper bound of $\mathcal{F}_S^{(SA)}\circ\mathcal{F}_1^{(FF)}\left(Z\right)_{t,k}$ for $Z$ outside $[0, B_x]$ in \eqref{out_zone_bound} and lower bound for $Z$ inside $[0, B_x]$ in \eqref{in_zone_bound}.

    \paragraph{Approximation Error.}
    Now, we can conclude our work by constructing the final feed-forward network $\mathcal{F}_2^{(FF)}$. 
    It receives the output of the self-attention layer and maps the ones in $\tilde{\mathbb{G}}_D\subset(3B_x/4D,\infty)^{d\times n}$ to the corresponding value of the target function, and the rest in $(-\infty,B_x/4D)^{d\times n}$ to $0$.

    In order to adapt to the $L_2$ norm, we use a continuous and Lipschitz function to map the input $Z$ to its targeted corresponding output $f(Q(Z))$.
    
    According to piece-wise linear approximation, function ${\mathcal{F}}_2^{(FF)}$ exists such that for any input $L \in G_D$, it maps it to corresponding $f(L)$, and for an arbitrary input $Z$, its output suffices:
    \begin{align}
        \mathcal{F}_2^{(FF)}(Z) \in [\min_{\| L-Z \|_{\max} \leq \frac{B_x}{2D}}  f(L),\max_{\| L-Z \|_{\max} \leq \frac{B_x}{2D}}  f(L)].
        \label{tilde_f_attribute}
    \end{align}

    Next we estimate the difference between $\mathcal{F}_2^{(FF)}\circ\mathcal{F}_S^{(SA)}\circ\mathcal{F}_1^{(FF)}$ and $\mathcal{F}_2^{(FF)}\circ\mathcal{F}_S^{(SA)}\circ\overline{\mathcal{F}}_1^{(FF)}$.
    
    The difference is caused by the difference between  $\overline{\mathcal{F}}_1^{(FF)}$ and $\mathcal{F}_1^{(FF)}$. By \eqref{differ_quant_and_linear}, this difference is bounded by $\frac1D$ in every dimension, for any input $Z \in \R^{d \times n}$:
    \begin{align*}
        \|\overline{\mathcal{F}}_1^{(FF)}(Z)
        - \mathcal{F}_1^{(FF)}(Z)\|_2
        < & ~
        \frac{\sqrt{dn}B_x}{D}.
    \end{align*}

    By \eqref{eqn:bound_2}:
    \begin{align*}
        & ~ \| \mathcal{F}_S^{(SA)}\circ\overline{\mathcal{F}}_1^{(FF)}(Z)-\mathcal{F}_S^{(SA)}\circ\mathcal{F}_1^{(FF)}(Z) \|_2 \\
        \leq & ~ \|\mathcal{F}_S^{(SA)}\circ\overline{\mathcal{F}}_1^{(FF)}(Z) - \overline{\mathcal{F}}_1^{(FF)}(Z)\|_2 + \|\overline{\mathcal{F}}_1^{(FF)}(Z) - \mathcal{F}_1^{(FF)}(Z)\|_2 \\ & ~ + \|\mathcal{F}_1^{(FF)}(Z) - \mathcal{F}_S^{(SA)}\circ\mathcal{F}_1^{(FF)}(Z)\|_2 \annot{By triangle inequality}\\
        \leq & ~
        \frac{\sqrt{dn }B_x}{D} + 2 \cdot \frac{\sqrt{n} B_x}{4D}.
        \annot{By $\|A\|_2 \leq \|A\|_F$ and \eqref{eqn:bound_2}}
    \end{align*}
    
    In the section on quantization of the input, we used piece-wise linear functions \eqref{eq:2.112} to approximate piece-wise-constant functions \eqref{eqn:quant}, this creates a deviation for the inputs on the boundaries of the constant regions. Consider $Z$ as one of these inputs whose value deviated from
    $\mathcal{F}_2^{(FF)}\circ\mathcal{F}_S^{(SA)}\circ\overline{\mathcal{F}}_1^{(FF)}(Q(Z))$. Let $f(L_1)$ denote the value given to $\mathcal{F}_2^{(FF)}\circ\mathcal{F}_S^{(SA)}\circ\mathcal{F}_1^{(FF)}(Z)$. Because the deviation take the output to a grid at most $\sqrt{dn}B_x/D+\sqrt{n}B_x/2D$ away from its original grid, under the quantization of the output, $f(L_1)$ at most deviate from its original output $ \mathcal{F}_2^{(FF)}\circ\mathcal{F}_S^{(SA)}\circ\overline{\mathcal{F}}_1^{(FF)} (Z)$ by the distance of $\sqrt{dn}B_x/D+\sqrt{n}B_x/2D$ aggregated with $2$ times of the maximal distance within a grid. They sum up to be:
    \begin{align*}
        \| \mathcal{F}_2^{(FF)}\circ\mathcal{F}_S^{(SA)}\circ\mathcal{F}_1^{(FF)}
        -\mathcal{F}_2^{(FF)}\circ\mathcal{F}_S^{(SA)}\circ\overline{\mathcal{F}}_1^{(FF)}
        \|_2
        \leq & ~
        L \cdot (\frac{2\sqrt{dn}B_x+\sqrt{n}B_x}{2D}+2\frac{\sqrt{dn}B_x}{D}) \\
        < & ~
        L\frac{6\sqrt{dn}B_x+\sqrt{n}B_x}{2D}.
    \end{align*}
    
    Lastly, by condition we neglect the $\mathbb{G}_D\setminus\tilde{\mathbb{G}}_D$ part. This yields:
    \begin{align*}
        \mathcal{F}_2^{(FF)}\circ\mathcal{F}_S^{(SA)}\circ\overline{\mathcal{F}}_1^{(FF)}
        =
        \overline{f}.
    \end{align*}

    Thus, adding up the errors yields:
    \begin{align*}
        & ~ \| f - \mathcal{F}_2^{(FF)}\circ\mathcal{F}_S^{(SA)}\circ{\mathcal{F}}_1^{(FF)} 
        \|_2 \\
        \leq & ~
        \| f - \Bar{f} \|_2 + \|\Bar{f} - \mathcal{F}_2^{(FF)}\circ\mathcal{F}_S^{(SA)}\circ{\mathcal{F}}_1^{(FF)} \|_2 \annot{By triangle inequality}\\
        = & ~
        L\frac{6\sqrt{dn}B_x+\sqrt{n}B_x}{2D} + L\frac{\sqrt{dn}B_x}{D} \annot{By \eqref{eqn:f_f_bar}}\\
        = & ~
        \frac{L(8\sqrt{dn}+\sqrt{n})B_x}{2D}.
    \end{align*}

    For any $\kappa >0$, we select large enough $D$, such that 
    \begin{align*}
        \frac{LB_x}{2D}(8\sqrt{dn}+\sqrt{n}) \leq \kappa.
    \end{align*}
    This completes the proof.
\end{proof}

\clearpage
\section{Experimental Details}
\label{app:sec_exp}

In this section, we conduct experiments to verify the capability of ICL to learn deep feed-forward neural networks.
We conduct the experiments based on 3-layer NN, 4-layer NN and 6-layer NN using both ReLU-Transformer and $\Softmax$-Transformer based on the GPT-2 backbone.

\paragraph{Experimental Objectives.}
Our objectives include the following three parts:
\begin{itemize}
    \item \textbf{Objective 1.}
    Validating the performance of ICL matches that of training $N$-layer networks, i.e., the results in \cref{thm:icl_gd_m}, \cref{thm:icl_gd_m_io}, and \cref{thm:icgd_soft}.
    \item \textbf{Objective 2.}
    Validating the ICL performance in scenarios where the testing distribution diverges from the pretraining one or where prompt lengths exceed those used in pretraining.
    \item \textbf{Objective 3.}
    Validating the ICL performance in scenarios where the distribution of parameters in the $N$-layer network diverges from that of the pretraining phase.
    \item \textbf{Objective 4.}
    Validating that a deeper transformer achieves better ICL performance, supporting the idea that scaling up the transformer enables it to perform more ICGD steps.
\end{itemize}

\textbf{Computational Resource.}
We conduct all experiments using 1 NVIDIA A100 GPU with 80GB of memory. 
Our code is based on the PyTorch implementation of the in-context learning for the transformer \cite{garg2022can} at \url{https://github.com/dtsip/in-context-learning}.

\subsection{Experiments for Objectives 1 and 2}
\label{subsec:obj_1_2}

In this section, we conduct experiments to validate Objectives 1 and 2.
We sample the input of feed-forward network $x \in \RR^{d}$ from the Gaussian mixture distribution: $w_1 N(-2, I_{d}) + w_2 N(2, I_{d})$, where $w_1, w_2 \in \RR$.
We consider three kinds of network $f : \RR^d \rightarrow \RR$, (i) 3-layer NN, (ii) 4-layer NN, and (iii) 6-layer NN.
We generate the true output by $y=f(x)$.
In our setting, we use $d=20$.

\textbf{Model Architecture.}
The sole difference between ReLU-Transformer and $\Softmax$-Transformer is the activation function in the attention layer.
Both models comprise 12 transformer blocks, each with 8 attention heads, and share the same hidden and MLP dimensions of 256.

\textbf{Transformer Pretraining.}
We pretrain the ReLU-Transformer and $\Softmax$-Transformer based on the GPT-2 backbone.
In our setting, we sample the pertaining data from $N(-2, I_{d})$, i.e., $w_1=1$ and $w_2=0$.
Following the pre-training method in  \cite{garg2022can}, we use the batch size as 64.
To construct each sample in a batch, we use the following steps (take the generation for the $i$-th sample as an example):
\begin{enumerate}
    \item Initialize the parameters in $f_i$ with a standard Gaussian distribution, i.e., $N(0, I)$.
    \item Generate $n$ queries $\left\{x_{i, j}\right\}_{j=1}^n$ (i.e., input of $f_i$) from the Gaussian mixture model $\omega_1 N(-2, I_d) + \omega_2 N(2, I_d)$.
    Here we take $n=51$. 
    \item For each query $x_{i, j}$, use $y_{i, j}=f_i(x_{i, j})$ to calculate the true output. 
\end{enumerate}
This generates a training sample for the transformer model with inputs
\begin{align*}
    \left[x_{i, 1}, y_{i, 1}, \cdots, x_{i, 50}, y_{i, 50}, x_{i, 51}\right],
\end{align*}
and training target 
\begin{align*}
    o_i = \left[y_{i, 1}, \cdots, y_{i, 50}, y_{i, 51}\right].
\end{align*} 
We use the MSE loss between prediction and true value of $o_i$. 
The pretraining process iterates for $500 \mathrm{k}$ steps.

\textbf{Testing Method.}
We generate samples similar to the pretraining process.
The batch size is 64, and the number of batch is 100, i.e., we have 6400 samples totally. 
For each sample, we extend the value $n$ from 51 to 76 to learn the performance of in-context learning when the prompt length is longer than we used in pretraining.
The input to the model becomes
\begin{align*}
    \left[x_{i, 1}, y_{i, 1}, \cdots, x_{i, 75}, y_{i, 75}, x_{i, 76}\right].
\end{align*}
We assess performance using the mean R-squared value for all 6400 samples.

\textbf{Baseline.} 
We use the 3-layer, 4-layer, and 6-layer feed-forward neural networks with 200 hidden dimensions as baselines by training them with in-context examples.
Specially, given a testing sample (take the $i$-th sample as an example), which includes prompts $\left\{x_{i, j}, y_{i, j}\right\}_{j=1}^{k-1}$ and a test query $x_{i, k}$.
We use $\left\{x_{i, j}, y_{i, j}\right\}_{j=1}^{k-1}$ to train the network with MSE loss for 100 epochs.
We select the highest R-squared value from each epoch as the testing measure and calculate the average across all 6400 samples.

\subsubsection{Performance of ReLU Transformer.}
\label{subsec:exp_relu}
\begin{figure}[ht]
    \centering
    \begin{subfigure}[b]{0.32\textwidth}
        \includegraphics[width=\textwidth]{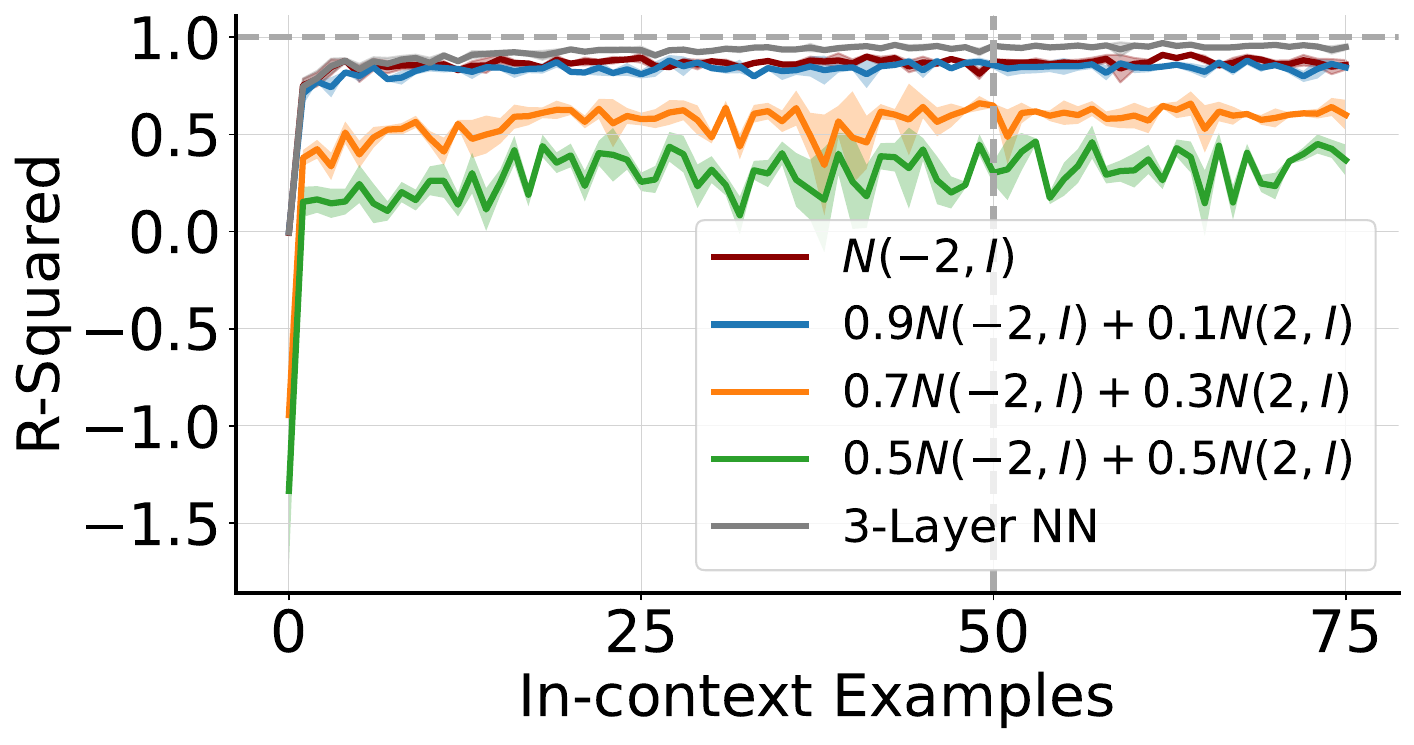}
        \caption{3-Layer NN}
        \label{fig:3_l_relu}
    \end{subfigure}
    \hfill
    \begin{subfigure}[b]{0.32\textwidth}
         \includegraphics[width=\textwidth]{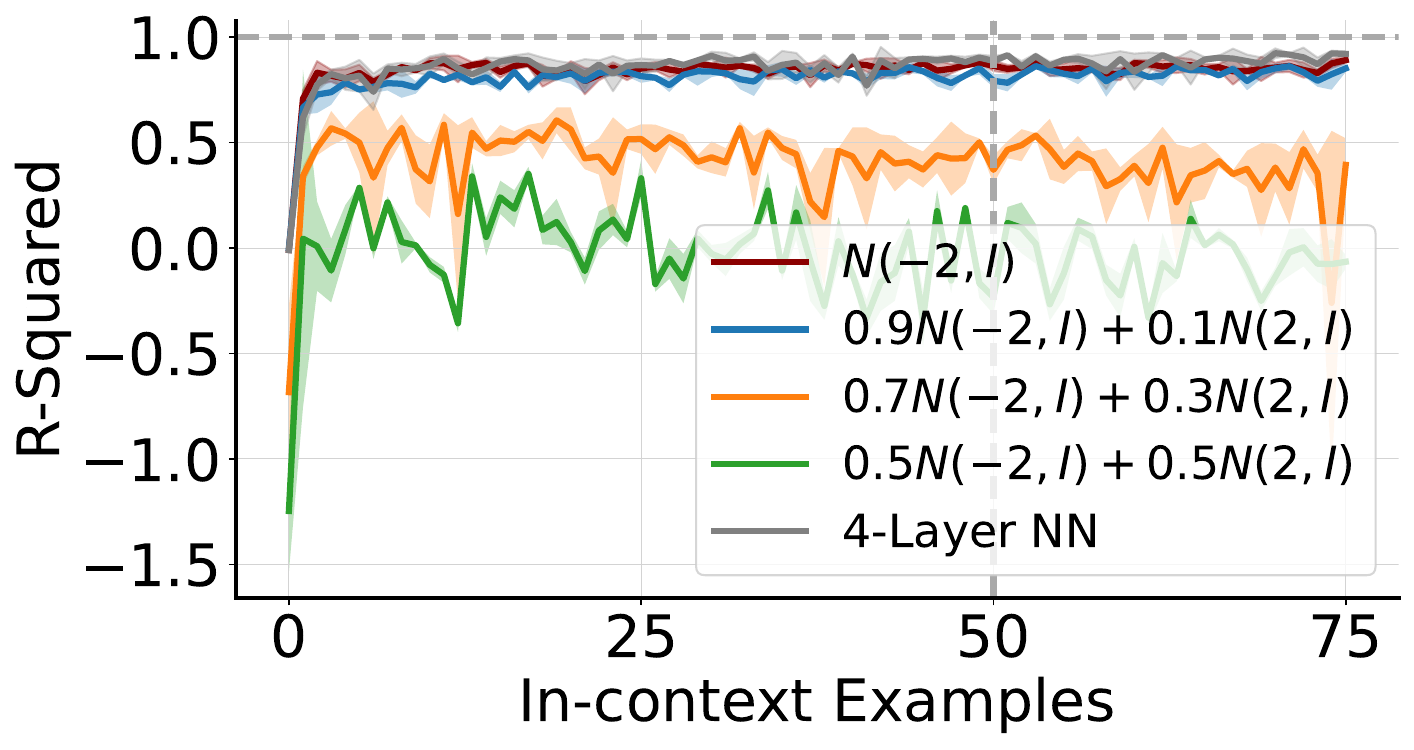}
         \caption{4-Layer NN}
         \label{fig:4_l_relu}
    \end{subfigure}
    \hfill
    \begin{subfigure}[b]{0.32\textwidth}
        \includegraphics[width=\textwidth]{figure/6NN_R.pdf}
        \caption{6-Layer NN}
        \label{fig:6_l_relu}
    \end{subfigure}
    \caption{\textbf{Performance of ICL in ReLU-Transformer:} 
    ICL learns 3-layer, 4-layer, and 6-layer NN and achieves R-squared values comparable to those from training with prompt samples.
    The results also show the ICL performance declines as the testing distribution diverges from the pretraining one.}
    \label{fig:relu}
\end{figure}

We use four different Gaussian mixture distributions $\omega_1 N(-2, I_d) + \omega_2 N(2, I_d)$ for the testing data: (i) $\omega_1 = 1,  \omega_2 = 0$, (ii) $\omega_1 = 0.9,  \omega_2 = 0.1$, (iii) $\omega_1 = 0.7,  \omega_2 = 0.3$, (iv) $\omega_1 = 0.5,  \omega_2 = 0.5$.
Here the distribution in the first setting matches the distribution in pretraining.
We show the results in \cref{fig:relu}.

\subsubsection{Performance of Softmax Transformer.}
\label{subsec:exp_soft}
\begin{figure}[ht]
    \centering
    \begin{subfigure}[b]{0.32\textwidth}
        \includegraphics[width=\textwidth]{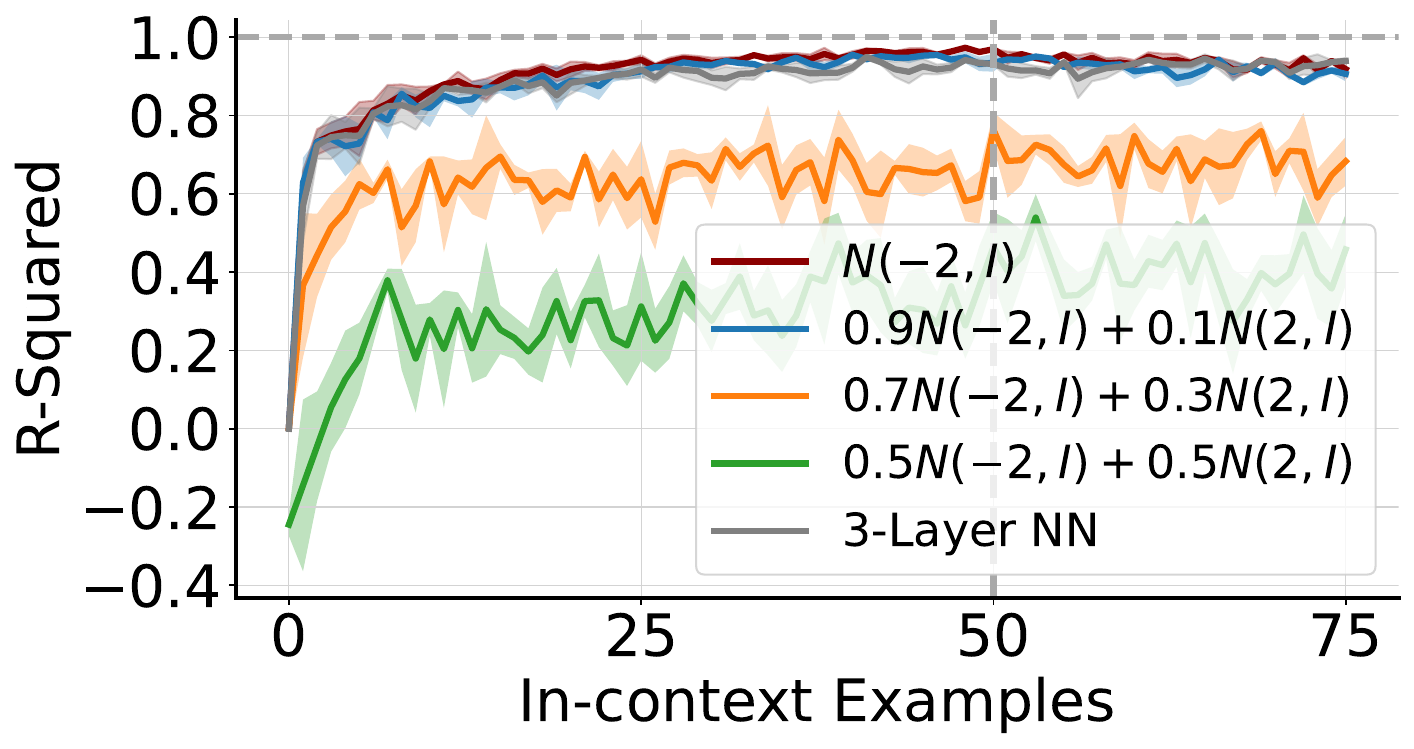}
        \caption{3-Layer NN}
        \label{fig:3_l_softmax}
    \end{subfigure}
    \hfill
    \begin{subfigure}[b]{0.32\textwidth}
         \includegraphics[width=\textwidth]{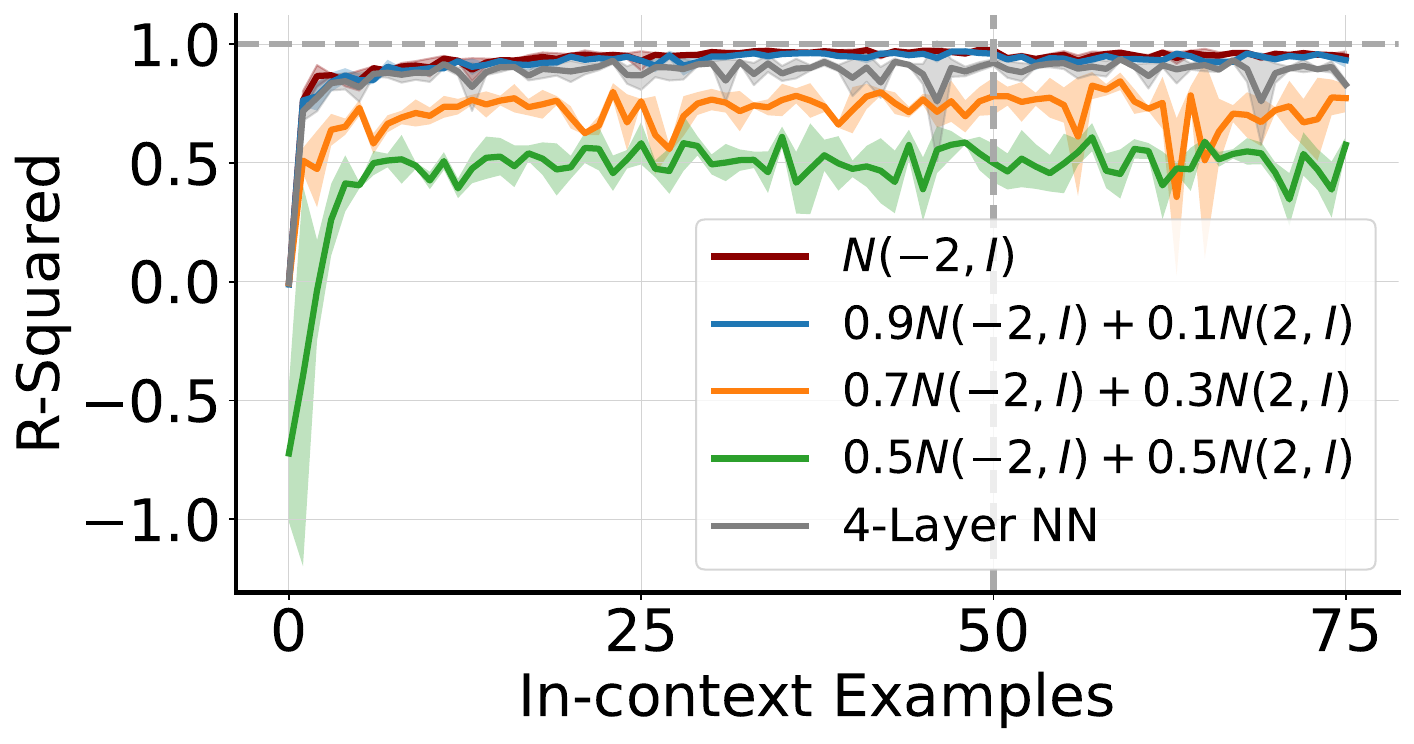}
         \caption{4-Layer NN}
         \label{fig:4_l_softmax}
    \end{subfigure}
    \hfill
    \begin{subfigure}[b]{0.32\textwidth}
        \includegraphics[width=\textwidth]{figure/6NN.pdf}
        \caption{6-Layer NN}
        \label{fig:6_l_softmax}
    \end{subfigure}
    \caption{\textbf{Performance of ICL in $\Softmax$-Transformer:} 
    ICL learns 3-layer, 4-layer, and 6-layer NN and achieves R-squared values comparable to those from training with prompt samples.
    The results also show the ICL performance declines as the testing distribution diverges from the pretraining one.
    Note that performance decreases when the prompt length exceeds the pretraining length (i.e., 50), a well-known issue \cite{dai2019transformer, anil2022exploring}. 
    We believe this is due to the absolute positional encodings in GPT-2, as noted in \cite{zhang2023trained}}
    \label{fig:softmax}
\end{figure}

We use four different Gaussian mixture distribution $\omega_1 N(-2, I_d) + \omega_2 N(2, I_d)$ for the testing data: (i) $\omega_1 = 1,  \omega_2 = 0$, (ii) $\omega_1 = 0.9,  \omega_2 = 0.1$, (iii) $\omega_1 = 0.7,  \omega_2 = 0.3$, (iv) $\omega_1 = 0.5,  \omega_2 = 0.5$.
Here the distribution in the first setting matches the distribution in pretraining.
We show the results in \cref{fig:softmax}.

The results in \cref{subsec:exp_relu} and \cref{subsec:exp_soft} show that the performance of ICL in the transformer matches that of training $N$-layer networks, regardless of whether the prompt lengths are within or exceed those used in pretraining.
Furthermore, the ICL performance declines as the testing distribution diverges from the pretraining one.

\subsection{Experiments for Objective 3}
\label{subsec:obj_3}
\begin{figure}[ht]
    \centering
    \begin{subfigure}[b]{0.32\textwidth}
        \includegraphics[width=\textwidth]{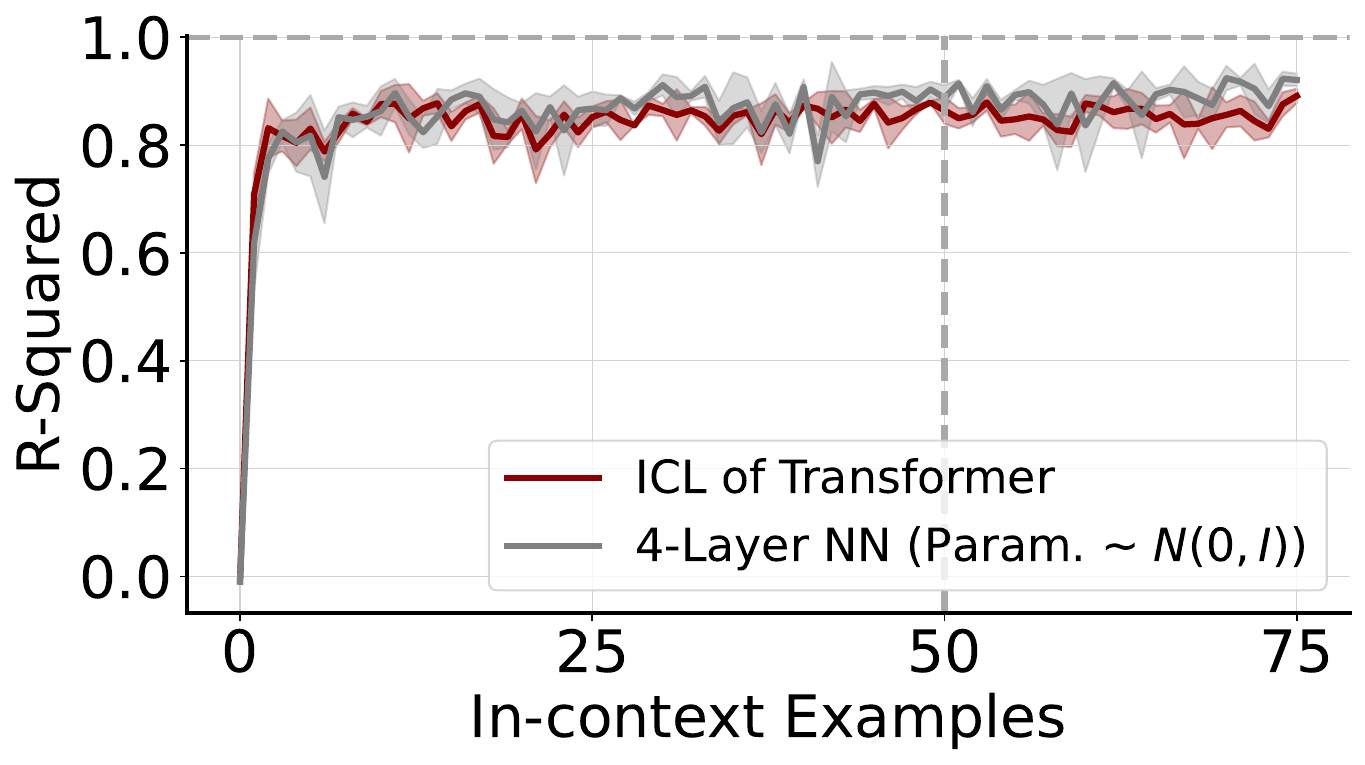}
        \caption{Parameters $\sim N(0, I)$}
    \end{subfigure}
    \hfill
    \begin{subfigure}[b]{0.32\textwidth}
         \includegraphics[width=\textwidth]{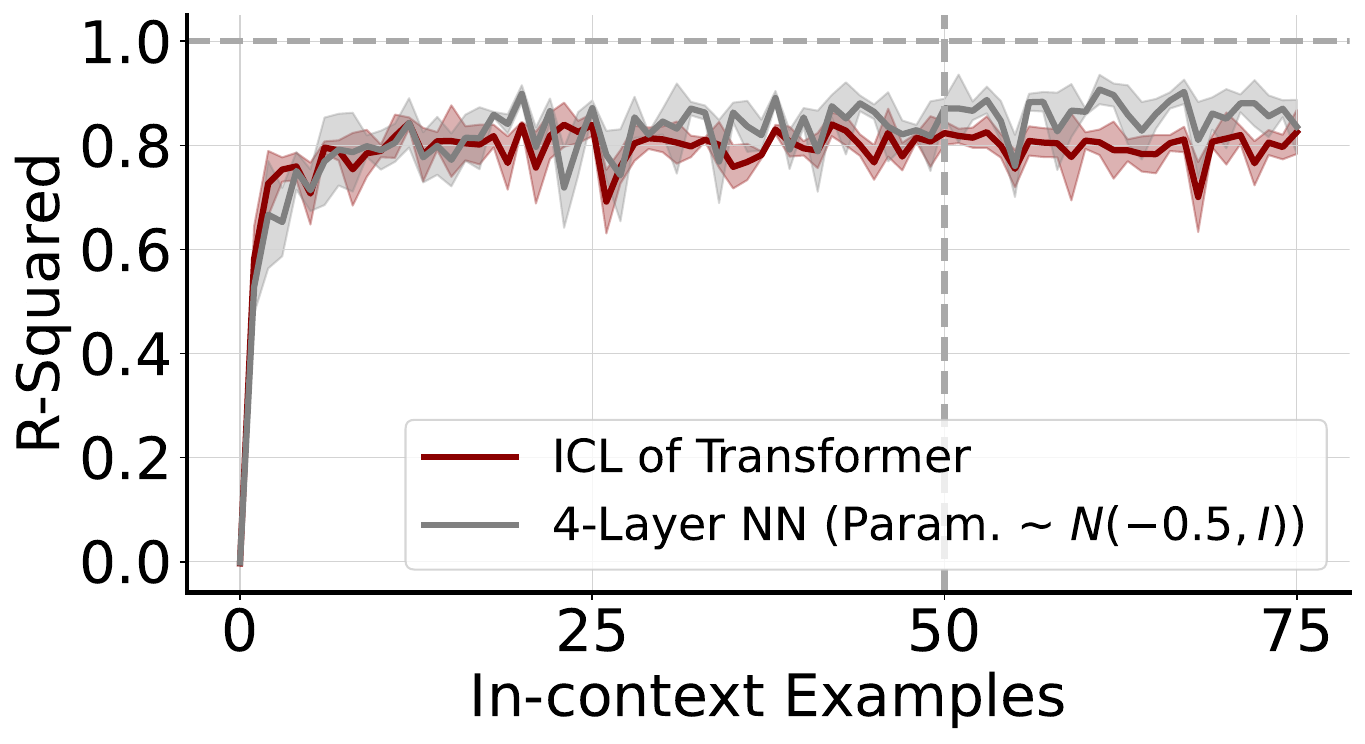}
         \caption{Parameters $\sim N(-0.5, I)$}
    \end{subfigure}
    \hfill
    \begin{subfigure}[b]{0.32\textwidth}
        \includegraphics[width=\textwidth]{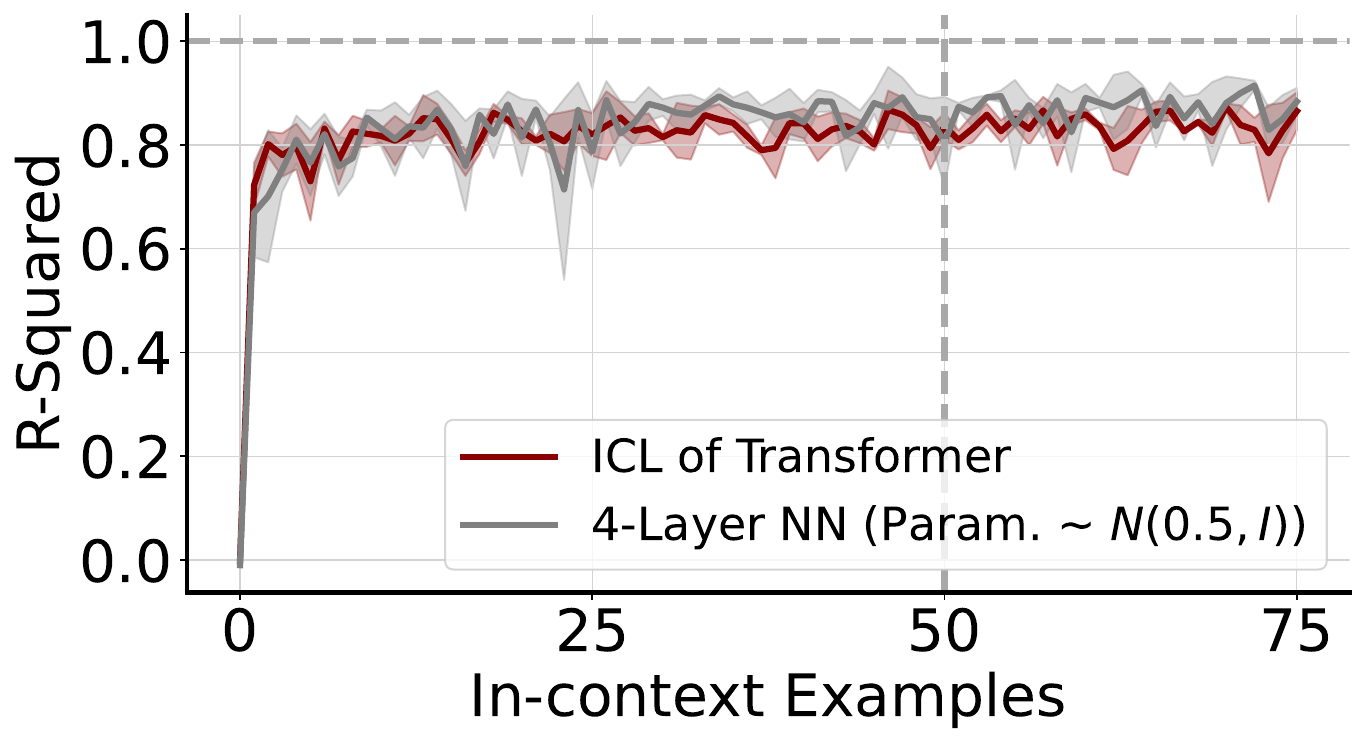}
        \caption{Parameters $\sim N(0.5, I)$}
    \end{subfigure}
    \caption{\textbf{Performance of ICL Across Various $N$-layer Network Parameter Distributions for the ReLU-Transformer:}
    ICL learns 4-layer NN and achieves R-squared values comparable to those from training with prompt samples, even when the parameter distribution in the $N$-layer network during testing diverges from that in the pretraining phase ($N(0, I)$).}
    \label{fig:relu_diff_para}
\end{figure}

\begin{figure}[ht]
    \centering
    \begin{subfigure}[b]{0.32\textwidth}
        \includegraphics[width=\textwidth]{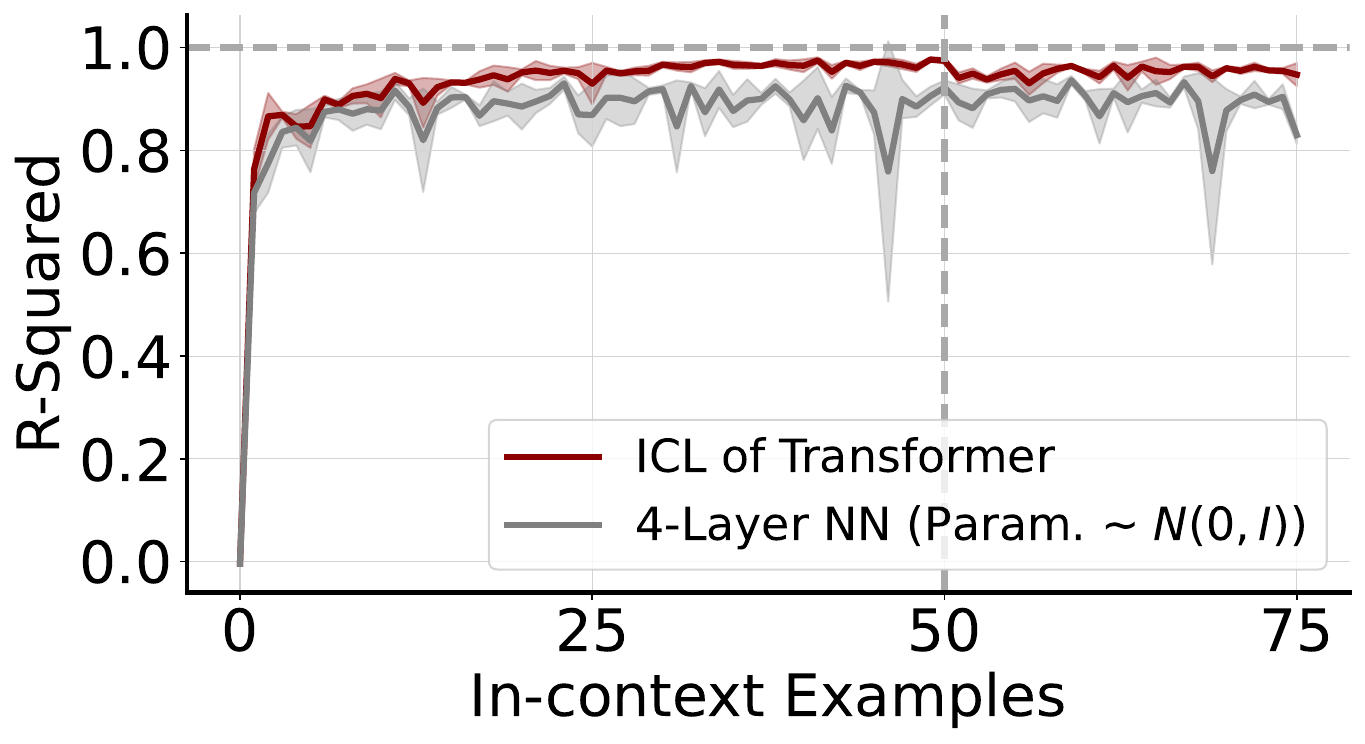}
        \caption{Parameters $\sim N(0, I)$}
    \end{subfigure}
    \hfill
    \begin{subfigure}[b]{0.32\textwidth}
         \includegraphics[width=\textwidth]{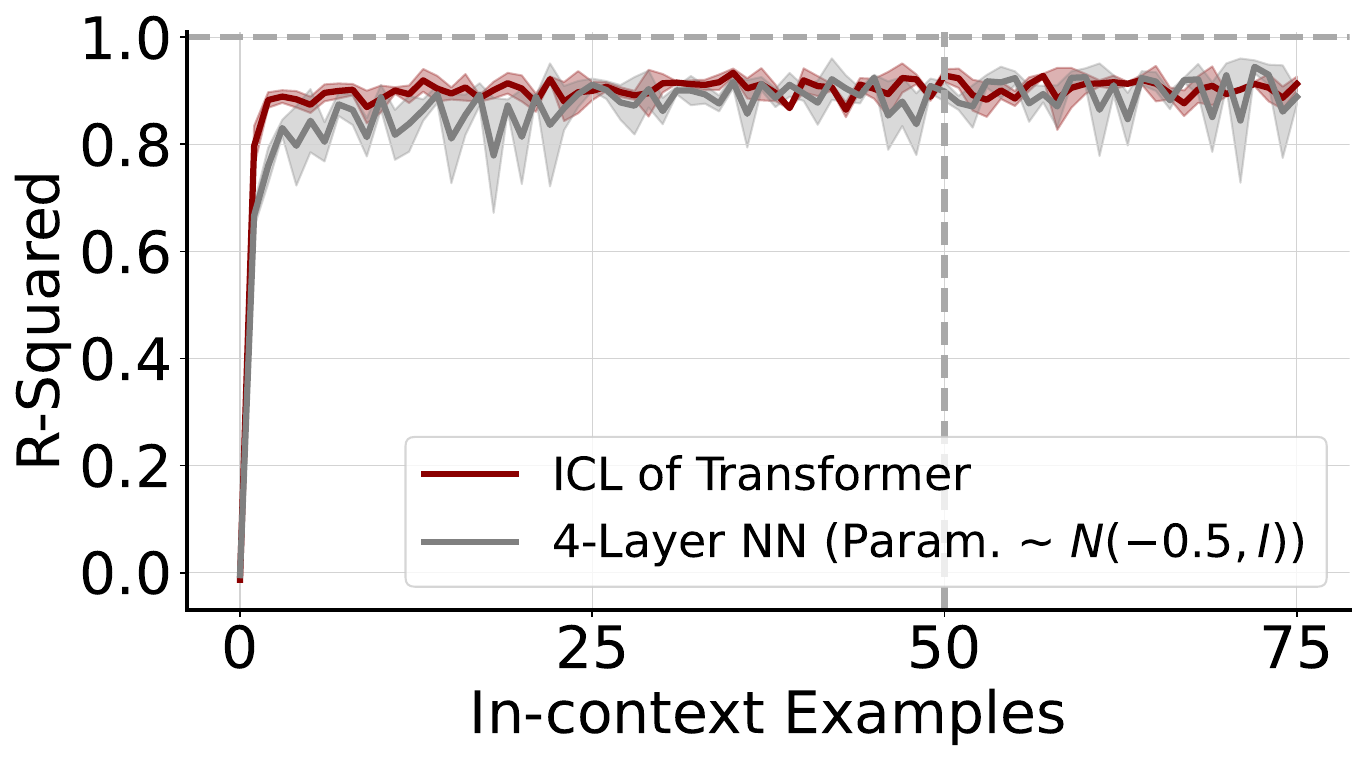}
         \caption{Parameters $\sim N(-0.5, I)$}
    \end{subfigure}
    \hfill
    \begin{subfigure}[b]{0.32\textwidth}
        \includegraphics[width=\textwidth]{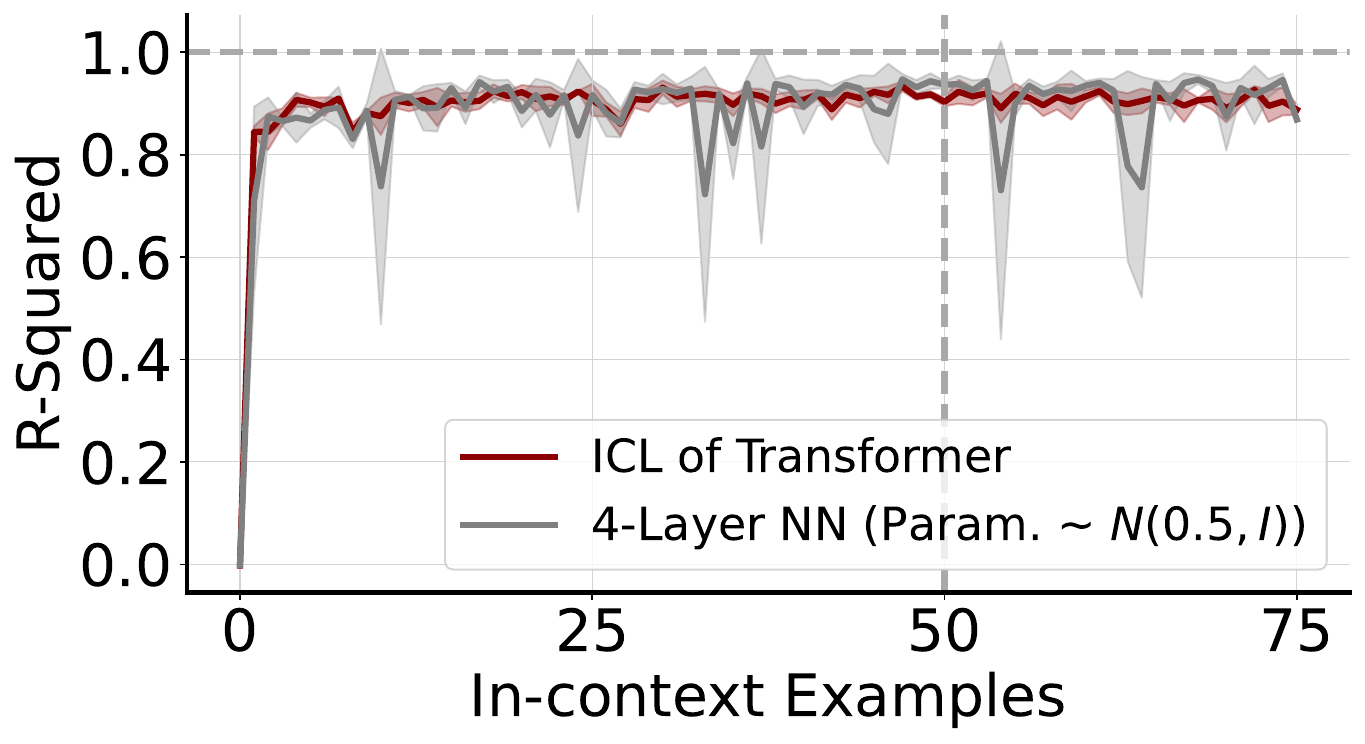}
        \caption{Parameters $\sim N(0.5, I)$}
    \end{subfigure}
    \caption{\textbf{Performance of ICL Across Various $N$-layer Network Parameter Distributions for the $\Softmax$-Transformer:}
    ICL learns 4-layer NN and achieves R-squared values comparable to those from training with prompt samples, even when the parameter distribution in the $N$-layer network during testing diverges from that in the pretraining phase ($N(0, I)$).}
    \label{fig:soft_diff_para}
\end{figure}

In this section, we conduct experiments to validate Objective 3. 
For these experiments, we use testing data that is identical to the training data, which follows a distribution of $N(-2, I_d)$.
We vary the distribution of parameters in the $N$-layer network.
During the training process, we set the distribution as $N(0, I)$.
In the testing process, we examine different distributions, including $N(0, I)$, $N(-0.5, I)$, and $N(0.5, I)$. 
All other model hyperparameters and experimental details remain consistent with those described in \cref{subsec:obj_1_2}. 
We evaluate the ICL performance of both the ReLU-Transformer and the Softmax-Transformer for $4$-layer networks, as shown in \cref{fig:relu_diff_para} and \cref{fig:soft_diff_para}. 
The results demonstrate that the ICL performance in the transformer matches that of training $N$-layer networks, regardless of whether the parameter distribution in the $N$-layer network diverges from that of the pretraining phase.

\subsection{Experiments for Objective 4}

\begin{figure}[ht]
    \centering
    \begin{subfigure}[b]{0.48\textwidth}
        \includegraphics[width=\textwidth]{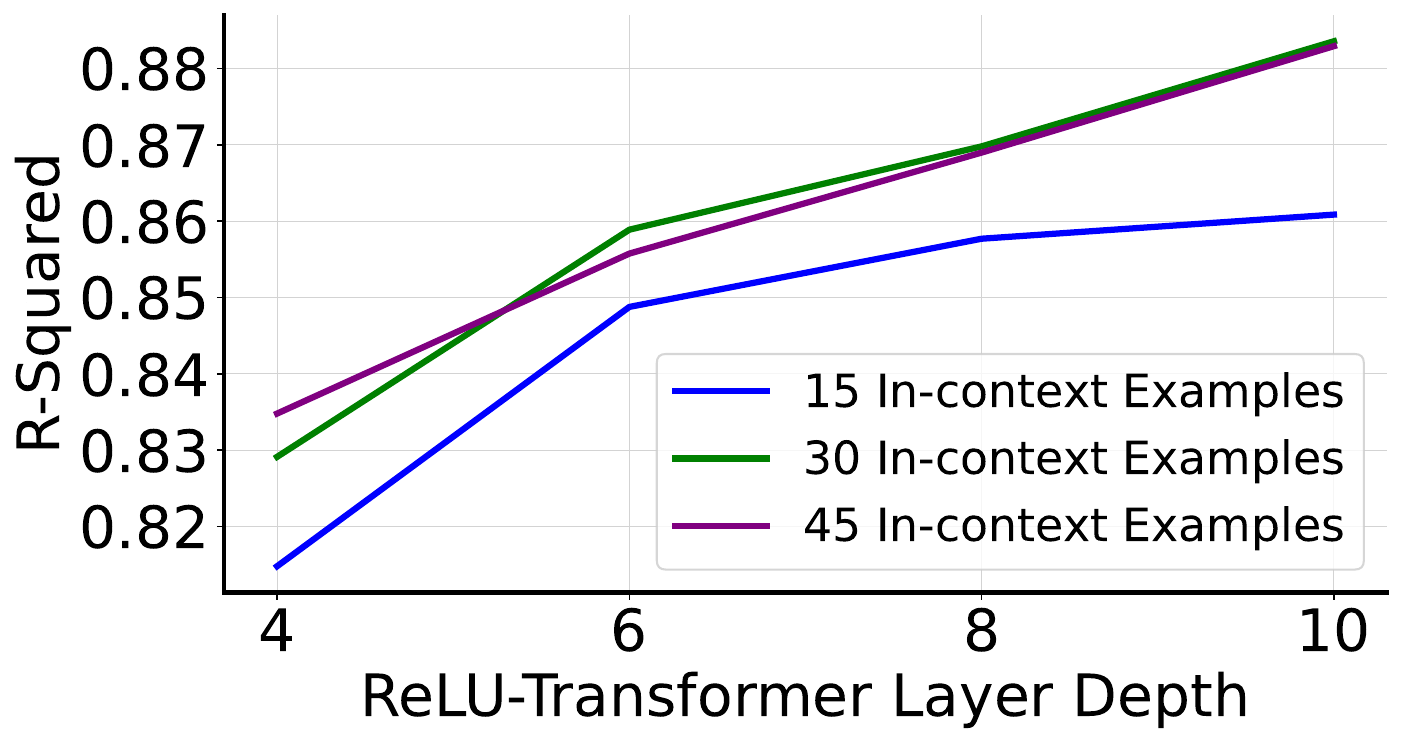}
        \caption{ReLU-Transformer}
    \end{subfigure}
    \hfill
    \begin{subfigure}[b]{0.48\textwidth}
         \includegraphics[width=\textwidth]{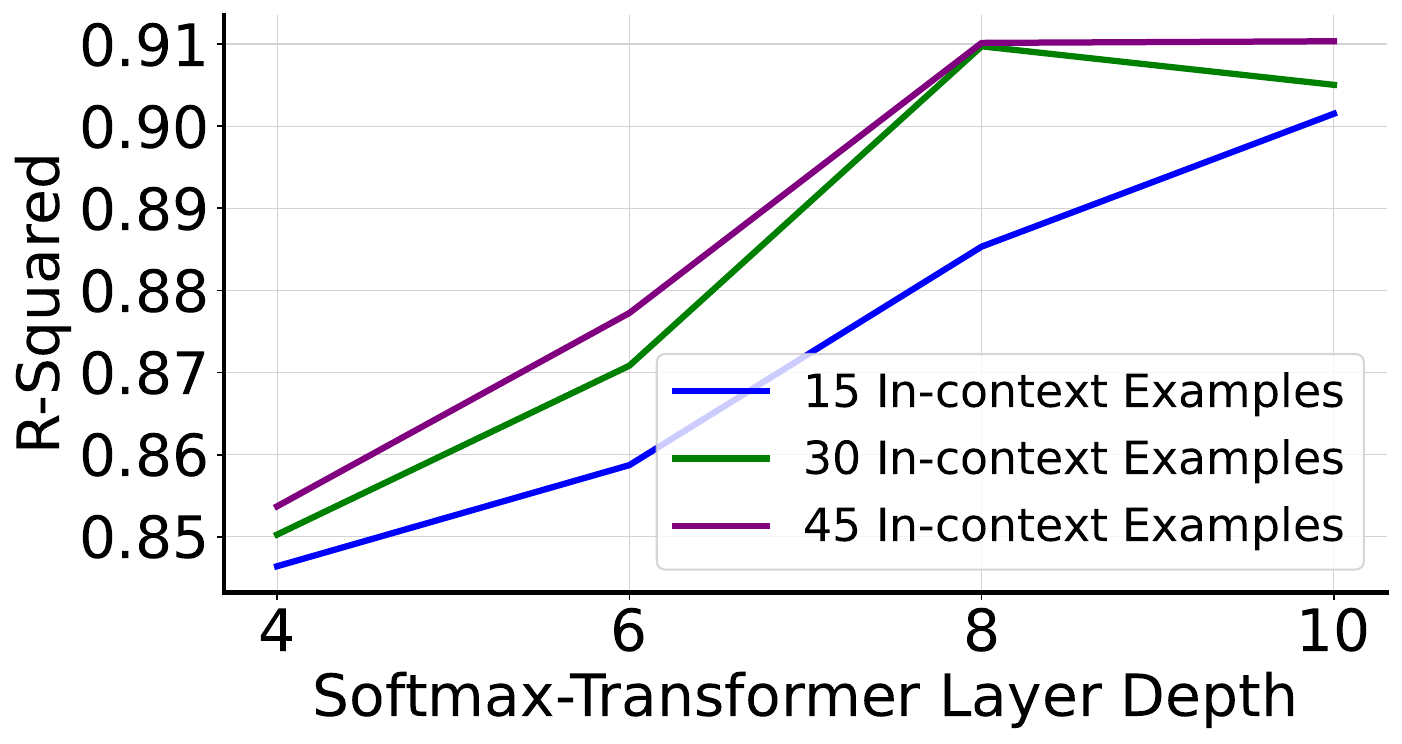}
         \caption{$\Softmax$-Transformer}
    \end{subfigure}
    \caption{\textbf{Performance of ICL Across Varying Transformer Depths}: 
    We use the number of in-context examples as 15, 30, or 45 for both the ReLU-Transformer and the $\Softmax$-Transformer. 
    The results show that a deeper transformer achieves better ICL performance, supporting the idea that scaling up the transformer enables it to perform more ICGD steps.}
    \label{fig:layer_depth}
\end{figure}

In this section, we conduct experiments to validate Objective 4. For these experiments, we use testing data identical to the pertaining data from $N(-2, I_d)$.
We vary the number of layers in the transformer architecture, testing configurations with 4, 6, 8 and 10 layers. 
All other model hyperparameters and experimental details remain consistent with those described in  \cref{subsec:obj_1_2}.
We evaluate the ICL performance of both the ReLU-Transformer and the $\Softmax$-Transformer with 15, 30, and 45 in-context examples, as shown in \cref{fig:layer_depth}.
The results show that a deeper transformer achieves better ICL performance, supporting the idea that scaling up the transformer enables it to perform more ICGD steps.

\clearpage
\section{Application: ICL for Diffusion Score Approximation}
\label{app:sec_icl_diff}

In this part, we give an important application of our work, i.e., learn the score function of diffusion models by the in-context learning of transformer models.
We give the preliminaries about score matching generative diffusion models in \cref{app:subsec_score_matching}.
Then, we give the analysis for ICL to approximate the diffusion score function in \cref{app:subsec_icl_score_appro}.

\subsection{Score Matching Generative Diffusion Models}
\label{app:subsec_score_matching}

\textbf{Diffusion Model.}
Let $x_0\in\R^d$ be initial data following target data distribution $x_0\sim P_0$.
In essence, a diffusion generative model consists of two stochastic process in $\R^{d}$:
\begin{itemize}
    \item A forward process gradually add noise to the initial data (e.g., images): $x_0\to x_1\to \cdots \to x_T$.

    \item A backward process gradually remove noise from pure noise:
    $y_T\to y_{T-1}\to \cdots \to y_0$.
\end{itemize}
Importantly, the backward process is the reversed forward process,
i.e., $y_t \stackrel{\mathrm{d}}{\approx} x_{T-t}$ for $i\in 0,\ldots,T$.\footnote{$\stackrel{\mathrm{d}}{\approx}$ denotes distributional equivalence.}
This allows the backward process to  reconstruct the initial data from noise, and hence generative.
To achieve this time-reversal, a diffusion model learns the reverse process by ensuring the backward conditional distributions mirror the forward ones.
The most prevalent technique for aligning these conditional dynamics is through ``score matching'' — a strategy training a model to match score function, i.e., the gradients of the log marginal density of the forward process \cite{songscore,song2020sliced,vincent2011connection}.
To be precise, let $P_t, p_t(\cdot)$  denote the distribution function and destiny function of $x_t$.  
The score function is given by $\nabla \log p_t(\cdot)$.
In this work, we focus on leveraging the in-context learning (ICL) capability of transformers to emulate the score-matching training process. 

\textbf{Score Matching Loss.}
We introduce the basic setting of score-matching as follows\footnote{Please also see \cref{sec:related_works} and  \cite{chen2024overview,chan2024tutorial,yang2023diffusion} for overviews.}.
To estimate the score function, we use the following loss to train a score network $s_W(\cdot, t)$ with parameters $W$:
\begin{align}
\label{eq:score_match_loss}
\min_{W} \int_{T_0}^T \gamma(t) \EE_{x_t \sim P_t} \left[\norm{s_W(x_t, t) - \nabla \log p_t(x_t)}_2^2 \right] \dd  t,\quad
\text{where $\gamma(t)$ is a weight function,}
\end{align}
and $T_0$ is a small value for stabilizing training and preventing the score function from diverging. 
In practice, as $\nabla \log p_t(\cdot)$ is unknown, we minimize the following equivalent loss \cite{vincent2011connection}.
\begin{align}
\label{eq:score_match_loss_pract}
\min_{W} \int_{T_0}^T \gamma(t) \EE_{x_0 \sim P_0} \left[ \EE_{x_t|x_0} \left[ \norm{s_W(x_t, t) - \nabla \log p(x_t|x_0)}_2^2 \right] \right] \dd  t,
\end{align}
where $p(x_t|x_0)$ is distribution of $x_t$ conditioned on $x_0$.

\subsection{ICL for Score Approximation}
\label{app:subsec_icl_score_appro}

We first give the problem setup about the ICL for score approximation as the following:

\begin{problem}[In-Context Learning (ICL) for Score Function $\nabla \log p_t(\cdot)$]
\label{prob:icl_score}
Consider the score function $\nabla \log p_t(\cdot)$ for any $t \geq 0$. Given a dataset $\mathcal{D}_n \coloneqq \left\{ (x_i, y_i) \right\}_{i \in [n]}$, where $\left\{ x_i \right\}_{i \in [n]} \subseteq \R^d$ and $y_i = \nabla \log p_{t_i}(x_i) \subseteq \R^d$ ($t_i \geq 0$), and a test input $x_{n+1}$, the goal of ``ICL for Score Function'' is to find a transformer $\calT$ to predict $y_{n+1}$ based on $x_{n+1}$ and the in-context dataset $\mathcal{D}_n$.
In essence, the desired transformer $\calT$ serves as the trained score network $s_W(\cdot,t)$.
\end{problem}
To solve \cref{prob:icl_score}, we follow two steps:
(i) Approximate the diffusion score function $\nabla \log p_t(\cdot)$ with a multi-layer feed-forward network with ReLU activation functions under the given training dataset $\mathcal{D}_n$.
(ii) Approximate the gradient descent used to train this network by the in-context learning of the Transformer until convergence, using the same training set $\mathcal{D}_n$ as the prompts of ICL. 

For the first step, we follow the score approximation results based on a multi-layer feed-forward network with ReLU activation in \cite{chen2023score}, stated as next lemma.

\begin{lemma}[Score Approximation by Feed-Forward Networks, Theorem 1 of \cite{chen2023score}]
\label{lm:score_appro_ff}
    Given an approximation error $\epsilon >0$, for any initial data distribution $P_0$, there exist a multi-layer feed-forward network with ReLU activation, $f(w, x, t): \R^{D_w} \times \R^d \times \R \rightarrow \R^d$. 
    Then for any $t \in [T_0, T]$, we have $\norm{f(w, \cdot, t) - \nabla \log p_t(\cdot)}_{L^2(P_t)} \leq \calO(\epsilon)$.
\end{lemma}

With the approximation result, we reduce the \cref{prob:icl_score} to \cref{prob:icl_n_m}, where the loss function is \eqref{eq:score_match_loss}.
Following \cref{thm:icl_gd_m}, we show that the in-context learning of transformer models can approximate the score function of diffusion model.